%% file: main_wacv.tex
\documentclass[10pt,twocolumn,letterpaper]{article}

\usepackage[pagenumbers]{wacv}  %

\input{preamble}

\usepackage[pagebackref,breaklinks,colorlinks]{hyperref}

\usepackage[capitalize]{cleveref}
\crefname{section}{Sec.}{Secs.}
\Crefname{section}{Section}{Sections}
\Crefname{table}{Table}{Tables}
\crefname{table}{Tab.}{Tabs.}

\def\wacvPaperID{1384} %
\def\confName{WACV}
\def\confYear{2025}

\title{Neural Graph Map: \\ Dense Mapping with Efficient Loop Closure Integration}

\author{Leonard Bruns$^{1}$
\qquad Jun Zhang$^{2}$ \qquad Patric Jensfelt$^1$\\
$^{1}$KTH Royal Institute of Technology, Stockholm, Sweden\\
$^{2}$TU Graz, Graz, Austria\\
{\tt\small \{leonardb, patric\}@kth.se}, {\tt\small jun.zhang@tugraz.at}
}

\begin{document}
\maketitle

\begin{abstract}
    Neural field-based SLAM methods typically employ a single, monolithic field as their scene representation. This prevents efficient incorporation of loop closure constraints and limits scalability. To address these shortcomings, we propose a novel RGB-D neural mapping framework in which the scene is represented by a collection of lightweight neural fields which are dynamically anchored to the pose graph of a sparse visual SLAM system. Our approach shows the ability to integrate large-scale loop closures, while requiring only minimal reintegration. Furthermore, we verify the scalability of our approach by demonstrating successful building-scale mapping taking multiple loop closures into account during the optimization, and show that our method outperforms existing state-of-the-art approaches on large scenes in terms of quality and runtime. Our code is available open-source at \url{https://github.com/KTH-RPL/neural_graph_mapping}.
\end{abstract}

\section{Introduction}\label{sec:introduction}

Simultaneous localization and mapping (SLAM) using cameras, often referred to as visual SLAM, has been a long standing problem in computer vision \cite{harris19883d,klein2007parallel,davison2007monoslam}. In particular, dense visual SLAM aims to construct a detailed geometric representation of the environment enabling various applications, such as, occlusion handling in augmented reality \cite{newcombe2011dtam}, planning and collision checking in robotics \cite{adamkiewicz2022vision}, or camera localization \cite{moreau2022lens}. Often volumetric scene representations are employed as they are well-suited for online data integration. Traditional volumetric representations use grid-based structures \cite{newcombe2011ismar}, often accelerated through the use of octrees \cite{meagher1982geometric,hornung2013octomap} or voxel hashing \cite{niessner2013real}. However, incorporating loop closure constraints in volumetric maps is difficult \cite{whelan2014real,thomas2017modeling,dai2017bundlefusion,reijgwart2019voxgraph} compared to sparse keypoint-based maps, which can easily be deformed. 

\begin{figure}[t]
    \centering
    \resizebox{\linewidth}{!}{\input{figures/fig1.tikz}}
    \caption{We propose to represent a scene by a set of neural fields (centers indicated by \textcolor{blue!70}{blue spheres}) anchored to keyframes in a pose graph with each field capturing the scene within a ball surrounding it. This allows to dynamically extend the scene while also incorporating loop closure deformations into the volumetric scene representation without requiring full reintegration.}
    \label{fig:firstfig}
    \vspace*{-\baselineskip}
\end{figure}

Recently, neural fields have emerged as a promising volumetric scene representation due to their amenability to differentiable rendering-based optimization \cite{tancik2020fourier}. Building on the first neural field-based SLAM method iMAP \cite{sucar2021imap}, various adaptations of this method have been proposed. These adaptations primarily focused on enhancing optimization speed by altering network architecture, sampling scheme, and rendering formulation. Nevertheless, the majority of existing neural field-based SLAM methods \cite{sucar2021imap,zhu2022nice,johari2023eslam,wang2023co,zhang2023go} remain constrained by their monolithic data structure (i.e., a single fixed-size architecture). In that regard, neural fields share the same limitation as other volumetric scene representations: after data has been integrated, the volumetric scene cannot be easily deformed to take loop closure constraints into account. 

To address this limitation, submap-based approaches have been proposed that maintain a small number of neural field-based submaps \cite{matsuki2024newton,tang2023mips,liso2024loopy}. However, submap-based methods come with their own set of difficulties such as inaccurate submap registration, transition artifacts at submap boundaries, and higher computational cost.

Instead of few, larger submaps, we propose to represent the scene by an extendable set of smaller, lightweight neural fields (see \cref{fig:firstfig}). These fields are dynamically anchored to keyframes in a pose graph, collectively forming the volumetric map while maintaining the ability to deform as the keyframe poses change upon loop closure. Our design allows to combine the benefits of sparse pose graph-based SLAM methods while maintaining a volumetric scene representation that remains consistent with the pose graph without requiring costly reintegration of previous sensor data. Furthermore, our method eliminates the necessity for fixed scene boundaries common among existing neural field-based SLAM methods. It achieves this by allocating additional fields dynamically as new parts of the scene come into view.

To summarize, our contributions are
\begin{itemize}\setlength\itemsep{0em}
    \item a novel RGB-D neural mapping framework that combines the robust, accurate tracking and efficient loop closure handling of sparse visual SLAM with the differentiable-rendering-based dense mapping of neural scene representations,
    \item a multi-field scene representation in which neural fields are anchored relative to a pose graph allowing for large-scale map deformations while limiting necessary reintegration, and
    \item a thorough comparison to multiple state-of-the-art methods on scenes of varying scales, including a novel set of sequences for the larger Replica scenes that allow to benchmark robustness and scalability.
\end{itemize}

\section{Related Work}\label{sec:relatedwork}

\paragraph{Traditional Volumetric Loop Closure} The difficulty of incorporating loop closure constraints in volumetric scene representations is not limited to neural representations. Traditional volumetric representations such as sparse grids and octrees also require solutions to efficiently adapt to pose graph changes. Notable examples include: Kintinuous \cite{whelan2014real} in which the volumetric map is only used for local fusion and globally a mesh is deformed based on a deformation graph; BundleFusion \cite{dai2017bundlefusion} in which frames whose pose has changed are removed and reintegrated; %
VoxGraph \cite{reijgwart2019voxgraph} in which submaps are aligned and resulting constraints are included in the pose graph optimization; and recently LivePose \cite{stier2023livepose} which extends the idea of BundleFusion to an RGB setting. Our work aims to enable efficient loop closure integration specifically for neural field-based representations for which de- and reintegration is difficult.

\paragraph{Neural Field-Based Scene Representations} Neural fields are parametrized differentiable functions that map coordinates to quantities at that point in space. Initially, neural fields were trained using 3D supervision to represent shapes as occupancy \cite{mescheder2019occupancy,chen2019learning} or signed distance fields (SDFs) \cite{park2019deepsdf}. Shortly after, neural radiance fields (NeRFs) were introduced optimized using only 2D images through differentiable volume rendering \cite{mildenhall2020nerf}. Replacing NeRF's density with an SDF-based representation higher quality surface reconstructions can be achieved \cite{wang2021neus,azinovic2022neural}.

Few works have investigated the use of multiple neural fields. In Block-NeRF~\cite{tancik2022block} multiple fields are combined to represent whole neighborhoods. However, the field positions are predefined. In Nerflets~\cite{zhang2023nerflets}, fields of varying size are composed to create an editable scene representation. NeRFuser~\cite{fang2023nerfuser} proposes methods for registering and blending multiple NeRFs. Finally, vMAP~\cite{kong2023vmap} uses multiple fields to represent individual objects highlighting the potential of vectorizing neural field evaluations. Our work follows similar motivations: by splitting the scene into multiple independent fields, it becomes both, efficiently deformable upon loop closures, and dynamically extendable as the camera moves.

\begin{figure*}[!ht]
    \centering
    \scriptsize
    \resizebox{\linewidth}{!}{\input{figures/framework.tex}}
    \caption{Neural graph map overview. Our framework can be used with any pose graph-based SLAM system and maintains a set of neural fields anchored relative to the keyframes of the pose graph. Each field captures the scene in a sphere surrounding it in local coordinates. This set of fields represents a deformable, volumetric scene representation that can easily adapt to and stay in sync with the pose graph. The fields are optimized independently always using the latest pose graph information for supervision. This design reduces transition artifacts, enables efficient optimization, and allows to average the output of overlapping fields in 3D leading to well-defined queries.}
    \label{fig:overview}
    \vspace*{-\baselineskip}
\end{figure*}

\paragraph{Neural Field-Based SLAM} iMAP ~\cite{sucar2021imap} was the first SLAM method using a multilayer perceptron (MLP) as the underlying scene representation. During optimization previous keyframes are continuously reintegrated to avoid forgetting. Subsequently, various modifications of this approach have been suggested aiming to alleviate this forgetting issue and improve run-time.

NICE-SLAM~\cite{zhu2022nice} addresses the forgetting issue by employing a hierarchical feature grid combined with a fixed, pretrained decoder MLP. By adopting an SDF-based field combined with efficient learnable positional encodings, ESLAM \cite{johari2023eslam} and Co-SLAM~\cite{wang2023co} demonstrate significantly improved optimization times. Specifically, ESLAM uses a tri-plane \cite{chan2022efficient} encoding, whereas Co-SLAM uses the aforementioned multi-resolution voxel hash encoding \cite{mueller2022instant}. Co-SLAM further complements the hash encoding with one-blob encodings \cite{muller2019neural} to improve scene completion. Point-SLAM \cite{sandstrom2023point} replaces NICE-SLAM's regular feature grid with with an irregular neural point cloud allowing to allocate more capacity in complex areas.
However, none of the aforementioned approaches support integration of loop closure constraints, preventing their use for larger scenes, in which accumulation of drift is unavoidable.

\cite{matsuki2024newton} follows a similar idea to ours, also using multiple fields posed relative to a pose graph. However, they employ space warping in which each field in principle covers the entire scene. Synthesized views from multiple fields are merged through alpha compositing. This approach is well-suited for novel view synthesis; however, it generalizes poorly to mesh extraction and other geometric queries. Instead, in our work, each field covers a Euclidean ball around it, which allows general geometric queries. In MIPS-Fusion \cite{tang2023mips} global bundle adjustment at the level of fields is performed. Loopy-SLAM \cite{liso2024loopy} also describes a submap-based system built on top of Point-SLAM \cite{sandstrom2023point}. Instead of larger keyframe-centric submaps, our approach represents the scene by many small fields, each covering a sphere surrounding it. By adjusting the field poses relative to the pose graph, while avoiding hard assignments of keyframes to fields during the optimization, our method allows to adapt to local and global pose graph deformations while avoiding transition artifacts common for submap-based methods.

Instead of using submaps, GO-SLAM \cite{zhang2023go} adapts to loop closures by biasing the optimization to keyframes whose pose changed the most. However, it still uses a single neural field, which requires reoptimization upon loop closure. Instead, we achieve instant map deformation upon loop closure by moving fields relative to the pose graph.

\section{Method}\label{sec:method}
We consider the problem of online RGB-D mapping without ground-truth poses. That is, given a stream of RGB-D frames $(\mathbf{C}_t,\mathbf{D}_t)$ composed of color images $\mathbf{C}_t\in \mathbb{R}^{H\times W\times 3}$ and depth maps $\mathbf{D}_t\in \mathbb{R}^{H\times W}$, our goal is to build a dense scene representation at each time step $t$.

\cref{fig:overview} gives an overview of the proposed framework. Similar to other recent neural field-based mapping approaches \cite{kong2023vmap,matsuki2024newton,zhang2023go}, our approach uses an off-the-shelf keyframe-based SLAM system to provide a set of posed keyframes as well as the pose of the current frame.

\subsection{Multi-Field Scene Representation}\label{sec:mfsr}

Let $\mathcal{K}_t=\{(\mathbf{C}_k, \mathbf{D}_k) \mid k= 1,\dots,K_t \}$ and $\mathcal{T}_t=\{\tensor*[^{\mathrm{w}}]{\mathbf{T}}{_{k}}\in\mathrm{SE}(3) \mid k= 1,\dots,K_t\}$ denote the set of keyframes and keyframe poses at time step $t$, respectively. %

We propose to represent the scene by an extendable set of posed neural fields
\begin{equation}
    \mathcal{F}_t=\{(f_i, \tensor*[^{\mathrm{w}}]{\mathbf{T}}{_{i}}) \mid i=1,...,F_t\},
\end{equation}
where each field
\begin{equation}
    \begin{aligned}
        f_i: \mathbb{R}^{3} &\to \mathbb{R}^3\times \mathbb{R} \\
        \boldsymbol{x} &\mapsto (\boldsymbol{c},s)
    \end{aligned}
\end{equation}
maps a 3D point $\boldsymbol{x}$ in its local reference frame to a color $\boldsymbol{c}$ and truncated signed distance $s$ at that point. Each field's reference frame is defined by its pose $\tensor*[^{\mathrm{w}}]{\mathbf{T}}{_{i}}\in \mathrm{SE}(3)$, which can be decomposed into a position $\tensor*[^{\mathrm{w}}]{\boldsymbol{t}}{_{i}}\in \mathbb{R}^3$ and orientation $\tensor*[^{\mathrm{w}}]{\mathbf{R}}{_{i}}\in \mathrm{SO}(3)$. Each field only captures the scene within a sphere of fixed radius $r$. 

We further define a function $p(i)$ that anchors a field $i$ to a parent keyframe $k$. At each iteration, given a field $i$ and its parent keyframe $k_i=p(i)$, we compute the field's pose according to $\tensor*[^{\mathrm{w}}]{\mathbf{T}}{_{i}}=\tensor*[^{\mathrm{w}}]{\mathbf{T}}{_{k_i}}\tensor*[^{k_i}]{\widetilde{\mathbf{T}}}{_{\mathrm{w}}}\tensor*[^{\mathrm{w}}]{\widetilde{\mathbf{T}}}{_{i}}$, where $\widetilde{\mathbf{T}}$ denotes transforms from the previous time step. That is, fields move as if they are rigidly connected to their parent keyframe. The parent keyframe can however change over time depending on the anchoring strategy defined by $p(i)$.

In the following sections, the proposed strategies to instantiate and dynamically anchor fields to the pose graph (\cref{sec:ins}), optimize the fields (\cref{sec:tsdfoptimization}), and query (e.g., for mesh extraction and view synthesis) the full scene representation (\cref{sec:queries}) are described. Finally, further architecture details are provided (\cref{sec:archdetails}).

\subsection{Field Instantiation and Anchoring}\label{sec:ins}

Whenever a new keyframe $k$ is added to the pose graph, new fields are instantiated such that all observed 3D points $\tensor*[^{\mathrm{w}}]{\mathcal{X}}{_k}$  in the keyframe are covered by at least one field of radius $r$. An approximate two-stage algorithm is used (see \cref{fig:gridinit}). First, the uncovered 3D points $\mathcal{X}_{\mathrm{unc}}=\{ \boldsymbol{x} \in \tensor*[^{\mathrm{w}}]{\mathcal{X}}{_k} \mid \lVert \boldsymbol{x} - \tensor*[^{\mathrm{w}}]{\boldsymbol{t}}{_{i}} \rVert > r\,\forall\,i\in\mathcal{N}_f \}$ are found. Second, the space is divided into voxel cells with a side length of $g=2r/\sqrt{3}$, such that a cell is fully covered by a field of radius $r$ when the center of the field is at the center of the cell. New fields are instantiated in the center of all cells that contain a point from $\mathcal{X}_{\mathrm{unc}}$ and no field center. This scheme ensures that a minimum distance of $g/2$ to existing fields is maintained; however, points in $\mathcal{X}_\mathrm{unc}$ might remain uncovered when fields move away from the cell centers through pose graph deformations. To alleviate the chance of uncovered points over time, the voxel grid is randomly shifted for every added keyframe. %

New fields added this way are initially anchored to the keyframe that triggered the creation. Over time fields are supervised by all keyframes that observe it (see \cref{sec:tsdfoptimization}), some of which might be significantly closer than the initial keyframe. Therefore, we define the parenting function $p(i)$ as the closest keyframe (Euclidean distance) with valid depth observation used to supervise the field. 

\begin{figure}[htb]
    \centering
    \scriptsize
    \input{figures/instantiation.tikz}
    \caption{Grid-based instantiation scheme. A new keyframe (top-left) observes world points $\mathcal{X}$ (black \& \textcolor{red!80!black}{red} dots). The points \textcolor{red!80!black}{$\mathcal{X}_\mathrm{unc}$ (red dots)} that are not covered by any of the existing fields \textcolor{blue!70}{$\mathcal{F}$ (blue circles)} are determined. Cells that contain uncovered points and no existing field center \textcolor{orange}{(hatched orange)} are used to instantiate new fields \textcolor{orange}{$\mathcal{F}_\mathrm{new}$ (orange circles)}. The new fields are positioned in the center of the uncovered cells and anchored to the keyframe.}.
    \label{fig:gridinit}
    \vspace*{-\baselineskip}
\end{figure}

\subsection{TSDF-Based Optimization}\label{sec:tsdfoptimization}

At each time step $t$, a fixed number of optimization steps $N_\mathrm{it}$ are performed on up to $N_\mathrm{f}$ fields. During the optimization the fields $f_i$ are trained independently, allowing for efficient parallelization \cite{kong2023vmap}. Only the fields' parameters are used as optimization variables. The fields' poses are updated solely according to the pose graph as described in \cref{sec:mfsr}.

Due to its well-defined isosurface for mesh extraction, we model the geometry as a truncated signed distance field (TSDF) adopting the TSDF-based rendering model from \cite{azinovic2022neural} and modify it to be occlusion-aware. Specifically, instead of directly converting signed distances to sample weights as done in \cite{azinovic2022neural,wang2023co}, we first convert them to occupancy probabilities. That is, given $N_\mathrm{s}$ samples along a ray $\boldsymbol{x}_i=\boldsymbol{o}+l_i\boldsymbol{d}, i=1,...,N_\mathrm{s}$ with origin $\boldsymbol{o}$ and direction $\boldsymbol{d}$, the colors and signed distances $(\boldsymbol{c}_i, s_i)=f(\boldsymbol{x}_i)$ along the ray are computed. The signed distances are converted to occupancy probabilities according to
\begin{equation}
    o_i =4\sigma\left(\frac{\eta s_i}{\tau}\right)\sigma\left(-\frac{\eta s_i}{\tau}\right),
\end{equation}
where $\sigma$ denotes the sigmoid function, $\tau$ is the truncation distance, and $\eta$ is a parameter that determines how sharply occupancy probability decays around the surface. Note that this definition ensures $o_i=1.0$ for $s_i=0.0$. The ray's rendered color and depth are then computed via the weight $w_i=o_i\prod_{j=1}^{i-1}(1-o_j)$ as 
\begin{equation}
        \boldsymbol{c}=\sum_{i=1}^{N_\mathrm{s}} w_i \boldsymbol{c}_i\quad
        d=\sum_{i=1}^{N_\mathrm{s}} w_i l_i
\end{equation}
assuming the ray's direction $\boldsymbol{d}$ was scaled appropriately. 

\paragraph{Sampling Strategy} Each optimization iteration follows a three-stage sampling procedure. Specifically, we sample: (1) $N_\mathrm{f}$ fields that will be optimized in the next iteration; (2) $N_\mathrm{r}$ ray segments (with associated observed color $\mathbf{c}$ and distance $d$) for each sampled field; (3) $N_\mathrm{s}$ query points along each sampled ray segment. Importantly, no assignments of keyframes to fields are made. Instead, which keyframes observe which field are recomputed in each iteration based on the latest pose graph information. This avoids transition effects stemming from different optimization targets of neighboring fields. The sampling strategy is visualized in \cref{fig:sampling} and details are provided in the supplementary material.

\paragraph{Loss}

Our loss is a weighted sum of four terms: color, depth, TSDF, and free-space. The weighing has been found experimentally to achieve a good trade-off between appearance and geometry quality.

The color loss is computed as the mean L1-norm of the difference between observed and estimated color
\begin{equation}
    l_\mathrm{color}=\frac{1}{|\mathcal{C}|}\sum_{(\boldsymbol{c},\boldsymbol{\tilde{c}})\in\mathcal{C}}\lVert\boldsymbol{c} - \boldsymbol{\tilde{c}}\rVert_1
\end{equation}
and the depth loss is computed as the mean Huber loss \cite{huber1964robust} between observed and estimated depth with $\delta=5\,\mathrm{cm}$
\begin{equation}
    l_\mathrm{depth}=\frac{1}{|\mathcal{D}|}\sum_{(d, \tilde{d})\in\mathcal{D}} \mathrm{Huber}_\delta(d, \tilde{d}),
\end{equation}
where $\mathcal{C}$ and $\mathcal{D}$ are sets containing tuples of the observed and estimated color and depth, respectively.

The query points $\boldsymbol{x}_i$ are filtered into two sets $\mathcal{X}_\mathrm{tsdf}$ and $\mathcal{X}_\mathrm{fs}$ based on whether a query point's projected depth $l_i$ is within the truncation distance $\tau$ of the corresponding observed depth $l_{\mathrm{obs},i}$ or is more than $\tau$ in front of it, respectively. A query point associated estimated signed distance $\tilde{s}_i$ and its signed distance to the observed depth $s_i=l_{\mathrm{obs},i}-l_i$ are then used to compute the TSDF loss as
\begin{equation}
    l_\mathrm{tsdf}=\frac{1}{|\mathcal{X}_\mathrm{tsdf}|}\sum_{\boldsymbol{x}_i\in \mathcal{X}_\mathrm{tsdf}}(\tilde{s}_i - s_i)^2
\end{equation}
and the free-space loss as
\begin{equation}
    l_\mathrm{fs}=\frac{1}{|\mathcal{X}_\mathrm{fs}|}\sum_{\boldsymbol{x}_i\in\mathcal{X}_\mathrm{fs}}(\tilde{s}_i - \tau)^2.
\end{equation}

\begin{figure}[htb]
    \centering
    \scriptsize%
    \input{figures/sampling.tikz}
    \caption{Multi-view sampling procedure. (1) A subset of fields (\textcolor{orange}{orange circles}) to optimize in the next iteration is sampled biased to the currently observed fields (\textcolor{orange}{orange outline}). (2) For each \textcolor{orange}{field}, the observing keyframes (opaque frustums) are approximated based on whether samples on the field boundary (\textcolor{red!80!black}{red dots}) fall into the observed keyframe region. Ray segments to supervise the field are then sampled from all observing keyframes. (3) Each sampled ray segment $[l_\mathrm{min},l_\mathrm{max}]$ is approximated by \textcolor{blue!70}{$N_\mathrm{up}$} uniformly sampled query points that cover the whole field and \textcolor{orange}{$N_\mathrm{dp}$} depth-guided query points distributed within the truncation distance $\tau$ around the observed distance $l_\mathrm{obs}$.}
    \label{fig:sampling}
    \vspace*{-\baselineskip}
\end{figure}

\subsection{Novel View Synthesis and Mesh Extraction}\label{sec:queries}

While the independent optimization ensures that overlapping regions will be supervised with the same data, fields might be at different stages of optimization and unobserved regions might converge to different solutions due to epistemic uncertainty. To reduce the resulting transition artifacts at the boundary of fields, it is possible to query the $k$ nearest fields and average the outputs (see supplementary material for details). For query points outside of all fields' spheres empty space is assumed, that is, $s=1$. For novel view synthesis, since no depth observation is available at inference time, a fixed far distance $l_\mathrm{far}$ is used. Query points are uniformly distributed within the ray interval $[0, l_\mathrm{far}]$ at the same density as the depth-guided points, that is, $N_\mathrm{p,inf}=l_\mathrm{far} N_\mathrm{dp} / (2\tau)$ samples are used. To extract a mesh the marching cubes algorithm \cite{lorensen1987marching} with an unbiased isosurface level of 0 is used. 

\subsection{Architecture Details}\label{sec:archdetails}

ORB-SLAM2 is used as the keyframe-based SLAM system \cite{mur2017orb}. However, our proposed method is agnostic to the exact SLAM framework as long as a pose graph as described in \cref{sec:mfsr} is available.

Each field is parameterized by a permutohedral hash encoding \cite{rosu2023permutosdf} followed by a linear layer with ReLU activation and another linear layer mapping to the color and distance. We also experimented with other positional encodings, such as Fourier encoding \cite{tancik2020fourier}, triplane encoding \cite{chan2022efficient,johari2023eslam}, and voxel hash encoding \cite{mueller2022instant}; however, the best results given the same time budget were achieved with the permutohedral hash encoding closely followed by the voxel hash encoding.

To efficiently evaluate and optimize multiple neural fields in parallel, optimization and evaluation is vectorized following \cite{kong2023vmap}. One limitation imposed by PyTorch's \cite{paszke2019pytorch} vectorization framework is that the batch size has to be the same for all evaluated networks. The design choices described in \cref{sec:tsdfoptimization} are tailored to this limitation sampling the same number of rays per supervised field.

\section{Experiments}\label{sec:experiments}

\subsection{Experimental Setup}\label{sec:implementationdetails}

\paragraph{Datasets} We evaluate our method quantitatively on three synthetic datasets of varying difficulty: the \emph{Replica} \cite{replica19arxiv} sequences from iMAP \cite{sucar2021imap}, the \emph{NRGBD} dataset by \cite{azinovic2022neural}, and three novel sequences on the larger Replica scenes (\emph{Replica-Big}). The latter contains challenging trajectories with larger loops spanning multiple rooms. We also include results on two real-world datasets: ScanNet \cite{dai2017scannet} and Kintinuous \cite{whelan2014real}. The latter contains a large-scale loop and the tested ScanNet sequences typically include one or two smaller loops. Quantitative evaluation of the map is not possible for the real-world datasets, because no accurate ground-truth is available in these datasets: ScanNet provides meshes from BundleFusion \cite{dai2017bundlefusion}, Kintinuous provides none.

\paragraph{Metrics} To assess the reconstruction quality of the methods we report the accuracy ratio, completion ratio, and F1-score \cite{tatarchenko2019single} at a $5\,\mathrm{cm}$ threshold. In the supplementary material we further report the non-robust metrics, accuracy and completion. We do not evaluate tracking metrics, since we use ORB-SLAM2 as the tracking backend and our focus is on dense mapping, while efficiently supporting map deformations due to loop closures.

\paragraph{Baselines} We compare our method to five state-of-the-art neural SLAM methods: NICE-SLAM \cite{zhu2022nice} and Co-SLAM \cite{wang2023co}, which do not support loop closure; GO-SLAM \cite{zhang2023go}, MIPS-Fusion \cite{tang2023mips}, and Loopy-SLAM \cite{liso2024loopy}, which support loop closure. Of these, GO-SLAM uses a monolithic scene representation; MIPS-Fusion few, large submaps; and Loopy-SLAM uses neural point-based submaps. In addition we compare to a monolithic, single-field ablation of our method (Ours-SF), which is discussed in \cref{sec:ablation}.

Loopy-SLAM is the only non-volumetric method representing only the volume within centimeters of the surfaces. When querying (for rendering, which in their case is used for subsequent mesh extraction), Loopy-SLAM relies on ground-truth depth maps. Because none of the other methods do so, it leads to an unfair advantage especially on the noise-free synthetic scenes. Therefore, we exclude Loopy-SLAM from the ranking in the tables.

\paragraph{Implementation Details} For all experiments we use the same parameters, most notably, $r=1\,\mathrm{m}$,  $N_\mathrm{it}=5$, $N_\mathrm{f}=32$, $N_\mathrm{r}=512$, $N_\mathrm{up}=8$, $N_\mathrm{dp}=16$ and $k=2$. We adjust the truncation distance $\tau$ from $0.1\,\mathrm{m}$ for synthetic data to $0.2\,\mathrm{m}$ for real data to account for the increased depth noise. For ORB-SLAM2 we use the default parameters in the RGB-D setup increasing the number of extracted features in challenging scenes with low texture. To facilitate reproducibility of our results, the full evolution of the pose graph was precomputed and played back as if ORB-SLAM is running in parallel. We will publish these recordings in conjunction with our code. All remaining parameters are provided in the supplementary material and as part of the released code.

\subsection{Reconstruction Quality Evaluation}\label{sec:evaluation}

\begin{table}[tb]
    \centering
    \begin{threeparttable}
        \caption{Comparison of mesh reconstruction quality (\textbf{best} \protect\gold, second best \protect\silver, third best \protect\bronze).}\label{tab:meshresults}
        
        \renewcommand{\arraystretch}{1.1}
        \setlength{\tabcolsep}{2pt}
        \scriptsize
        \begin{tabular}{
          @{}
          ll
          S[table-format=2.2]@{\hspace{0.3\tabcolsep}}c
          S[table-format=2.2]@{\hspace{0.3\tabcolsep}}c
          S[table-format=2.2]@{\hspace{0.3\tabcolsep}}c
          S[table-format=2.2]@{\hspace{0.3\tabcolsep}}c
          S[table-format=2.2]@{\hspace{0.3\tabcolsep}}c
          @{}
          }
            \toprule
            & & \multicolumn{2}{c}{Replica} & \multicolumn{2}{c}{NRGBD} & \multicolumn{6}{c}{Replica-Big} \\
            \cmidrule(lr){3-4} \cmidrule(lr){5-6} \cmidrule(lr){7-12} 
            & & \text{Avg.$^\ast$} & & \text{Avg.$^\ast$} & & \texttt{apt0} & & \texttt{apt1} & & \texttt{apt2} \\
            \midrule
            \multirow{3}{*}{NICE-S. \cite{zhu2022nice}} 
            & Acc. R. (\%) & \bfseries 93.17 & \gold & 78.10 &  & 30.67 &  & 41.48 &  & 28.20 &  \\
            & Comp. R. (\%) & 90.11 &  & 70.58 &  & 70.58 &  & 42.21 &  & 26.80 &  \\
            & F1-Score (\%) & 91.60 &  & 74.07 &  & 74.07 &  & 41.84 &  & 27.49 &  \\
            \midrule
            \multirow{3}{*}{Co-S. \cite{wang2023co}} 
            & Acc. R. (\%) & 87.02 &  & 87.23 &  & 18.85 &  & 62.11 &  & 34.71 &  \\
            & Comp. R. (\%) & 86.39 &  & 82.86 & \bronze & 21.93 &  & 66.37 &  & 36.75 &  \\
            & F1-Score (\%) & 86.70 &  & 84.94 & \bronze & 20.27 &  & 64.17 &  & 35.70 &  \\
            \midrule
            \multirow{3}{*}{GO-S. \cite{zhang2023go}} 
            & Acc. R. (\%) & 89.31 &  & 84.10 &  & \bfseries 57.93 & \gold & 74.38 &  & \bfseries 80.22 & \gold \\
            & Comp. R. (\%) & 74.20 &  & 66.19 &  & \bfseries 59.21 & \gold & 73.90 & \bronze & 81.06 & \silver \\
            & F1-Score (\%) & 81.03 &  & 73.98 &  & \bfseries 58.56 & \gold & 74.14 & \bronze & \bfseries 80.64 & \gold \\
            \midrule
            \multirow{3}{*}{MIPS-F. \cite{tang2023mips}} 
            & Acc. R. (\%) & 92.63 &  & 88.29 & \bronze & 28.00 &  & 75.97 & \bronze & 33.16 &  \\
            & Comp. R. (\%) & \bfseries 91.38 & \gold & 75.85 &  & 28.11 &  & 72.14 &  & 34.76 &  \\
            & F1-Score (\%) & 92.00 & \silver & 81.40 &  & 28.06 &  & 74.01 &  & 33.94 &  \\
            \midrule
            \multirow{3}{*}{Loopy-S.$^\dagger$ \cite{liso2024loopy}} 
            & Acc. R. (\%) & 99.99 & & 80.52 &  & \text{\protect\xmark} &  & 31.34 &  & 59.87 &  \\
            & Comp. R. (\%) & 89.28 &  & 72.66 &  & \text{\protect\xmark} &  & 21.71 &  & 79.04 &  \\
            & F1-Score (\%) & 94.32 & & 76.29 &  & \text{\protect\xmark} &  & 25.65 &  & 68.14 &  \\
            \midrule
            \multirow{3}{*}{Ours-SF}
            & Acc. R. (\%) & 93.11 & \silver & \bfseries 93.11 & \gold & 54.69 & \bronze & 93.14 & \silver & 76.86 & \bronze \\
            & Comp. R. (\%) & 91.28 & \silver & \bfseries 85.45 & \gold & 56.07 & \bronze & \bfseries 95.42 & \gold & 79.57 & \bronze \\
            & F1-Score (\%) & \bfseries 92.19 & \gold & \bfseries 89.02 & \gold & 55.37 & \bronze & 94.27 & \silver & 78.19 & \bronze \\
            \midrule
            \multirow{3}{*}{Ours} 
            & Acc. R. (\%) & 92.76 & \bronze & 93.09 & \silver & 56.25 & \silver & \bfseries 94.14 & \gold & 79.28 & \silver \\
            & Comp. R. (\%) & 90.87 & \bronze & 84.70 & \silver & 56.89 & \silver & 95.25 & \silver & \bfseries 81.81 & \gold \\
            & F1-Score (\%) & 91.80 & \bronze & 88.58 & \silver & 56.57 & \silver & \bfseries 94.69 & \gold & 80.52 & \silver \\
             \bottomrule
        \end{tabular}
    \begin{tablenotes}
        \item[$\ast$] Full results are available in the supplementary material.
        \item[$\dagger$] Excluded from ranking because it uses ground-truth depth during evaluation.
    \end{tablenotes}
    \end{threeparttable}
    \vspace*{-\baselineskip}
\end{table}

\begin{figure*}[!htbp]
    \centering
    \scriptsize
    \setlength{\fboxsep}{0pt}
    \setlength{\tabcolsep}{1pt}
    \renewcommand{\arraystretch}{0.8}
    \begin{tabular}{C{1em}C{0.16\linewidth}C{0.16\linewidth}C{0.16\linewidth}C{0.16\linewidth}C{0.16\linewidth}C{0.16\linewidth}}
        & Ground-truth & Co-SLAM \cite{wang2023co} & GO-SLAM \cite{wang2023co} & MIPS-Fusion \cite{tang2023mips} & Loopy-SLAM \cite{liso2024loopy} & Ours \\
        \rotatebox{90}{\texttt{apt0}} &
        \includegraphics[width=\linewidth]{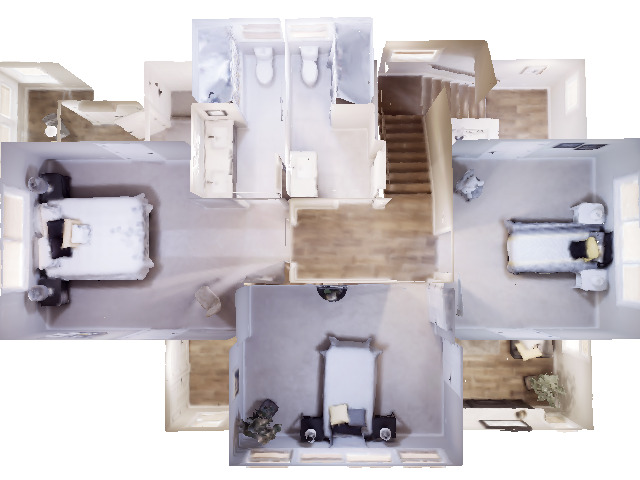} &
        \includegraphics[width=\linewidth]{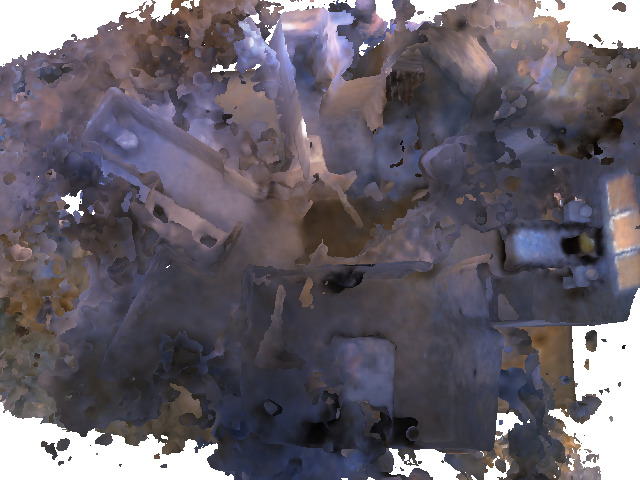} &
        \includegraphics[width=\linewidth]{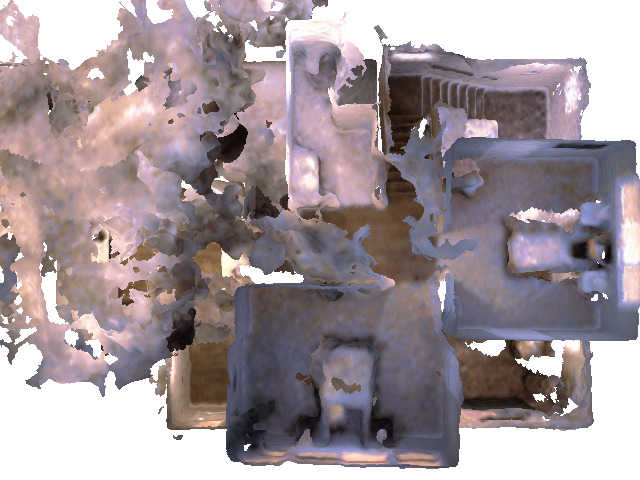} &
        \includegraphics[width=\linewidth]{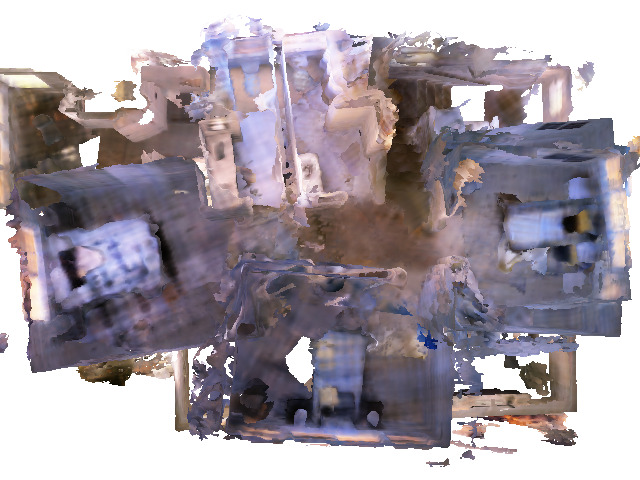} &

        \begin{tabular}{c}{\resizebox{1.5cm}{!}{\xmark}}\end{tabular}

        &
        \includegraphics[width=\linewidth]{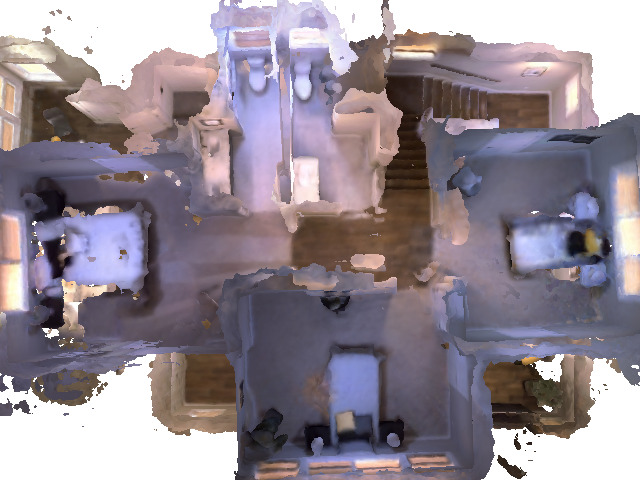} \\
        \rotatebox{90}{\texttt{apt1}} &
        \includegraphics[width=\linewidth]{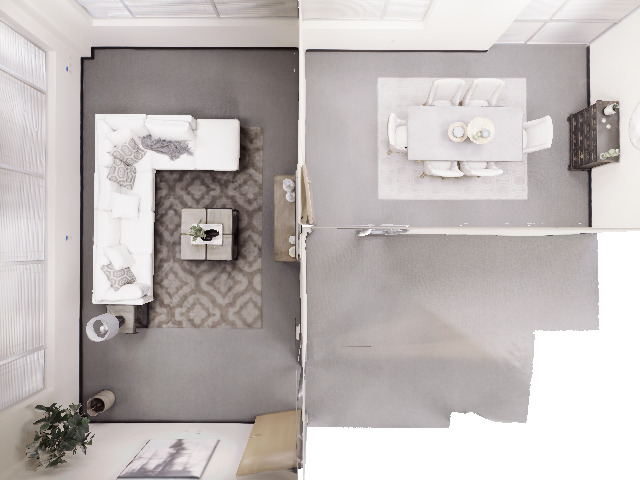} &
        \includegraphics[width=\linewidth]{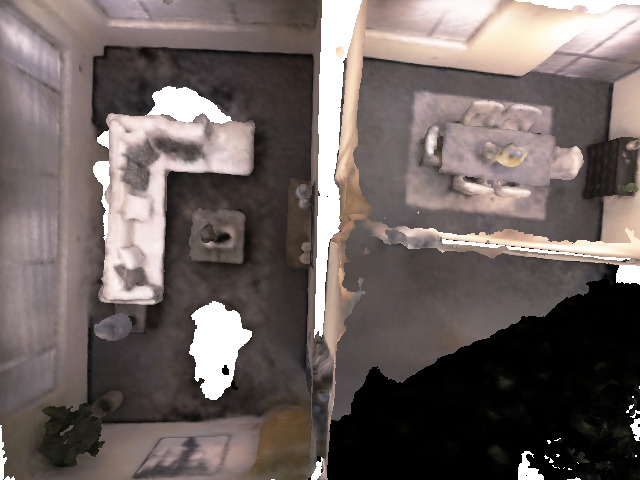} &
        \includegraphics[width=\linewidth]{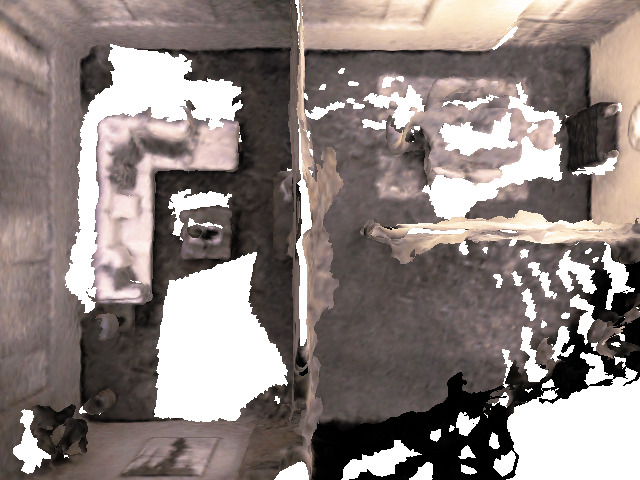} &
        \includegraphics[width=\linewidth]{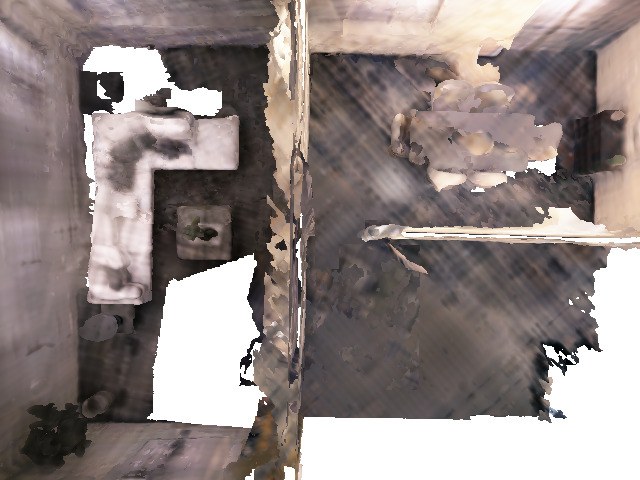} &
        \includegraphics[width=\linewidth]{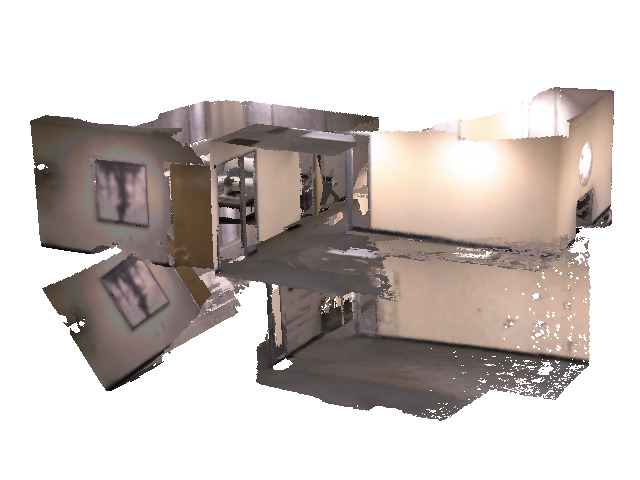} &
        \includegraphics[width=\linewidth]{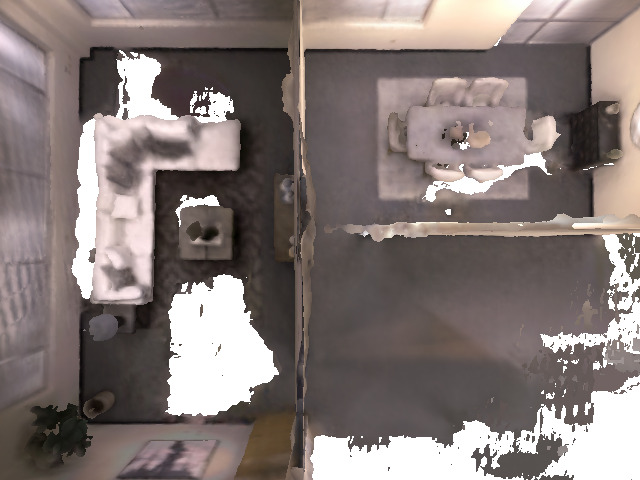} \\
        \rotatebox{90}{\texttt{apt2}} &
        \includegraphics[width=\linewidth]{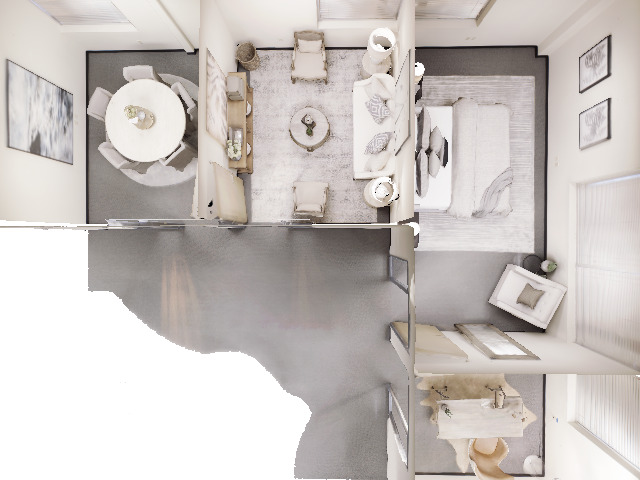} &
        \includegraphics[width=\linewidth]{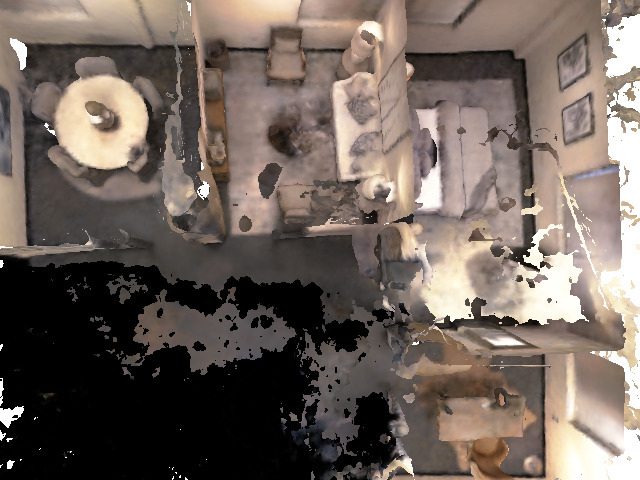} &
        \includegraphics[width=\linewidth]{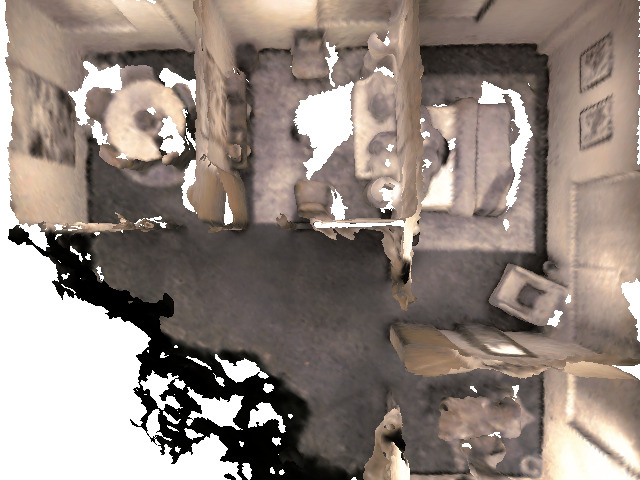} &
        \includegraphics[width=\linewidth]{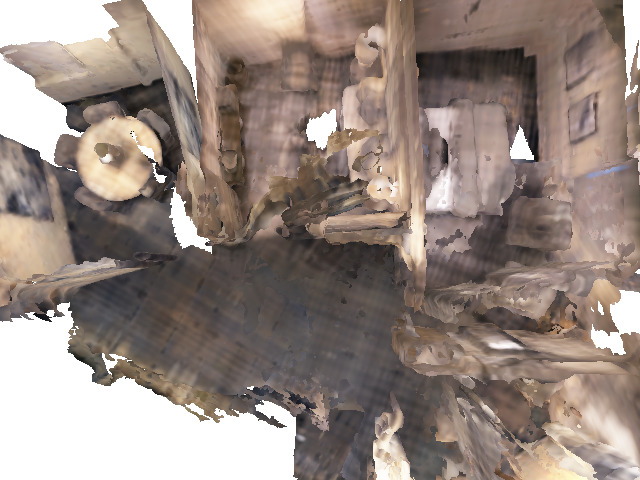} &
        \includegraphics[width=\linewidth]{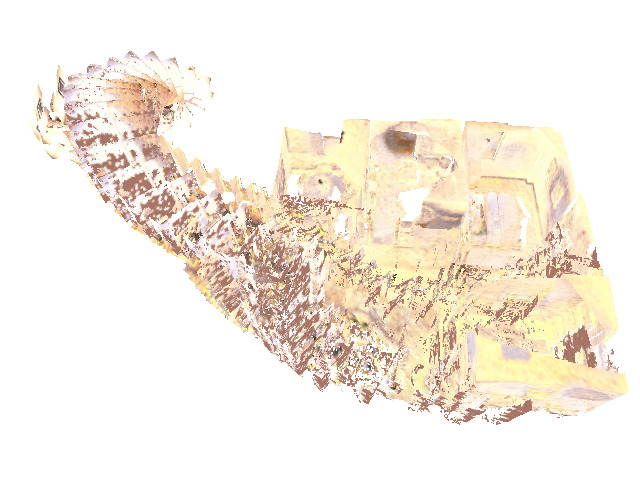} &
        \includegraphics[width=\linewidth]{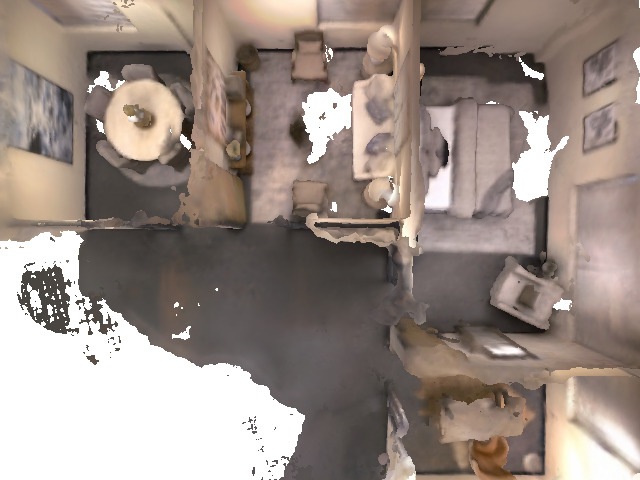}\\
        \rotatebox{90}{\texttt{scene0000\_00}} &
        \includegraphics[width=\linewidth,trim={3cm 0 3cm 0},clip]{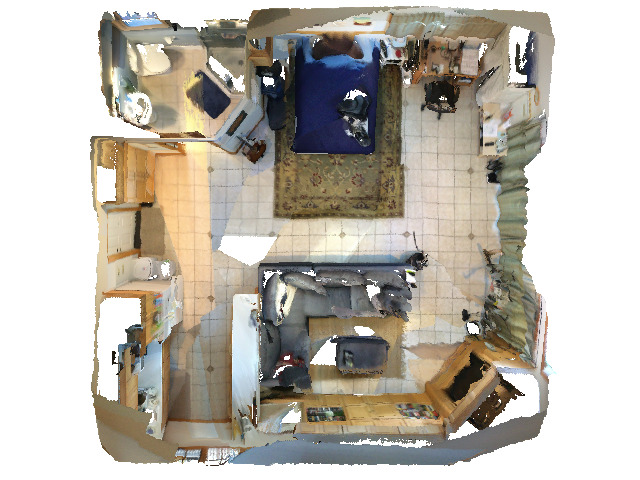} &
        \includegraphics[width=\linewidth,trim={3cm 0 3cm 0},clip]{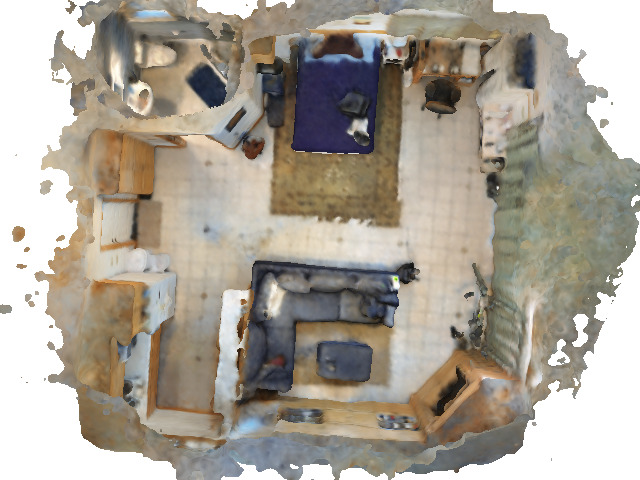} &
        \includegraphics[width=\linewidth,trim={3cm 0 3cm 0},clip]{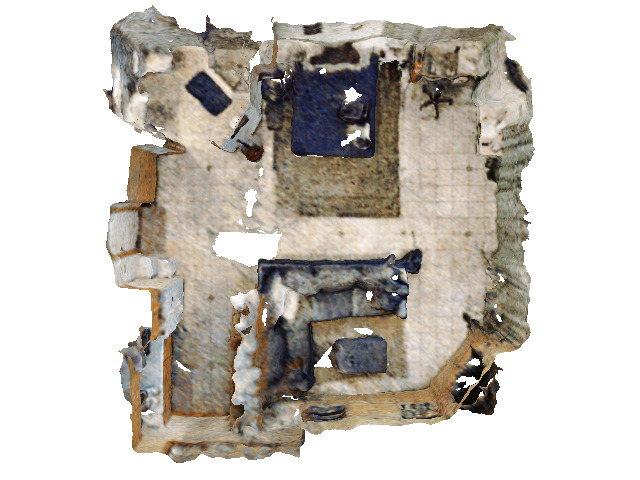} &
        \includegraphics[width=\linewidth,trim={3cm 0 3cm 0},clip]{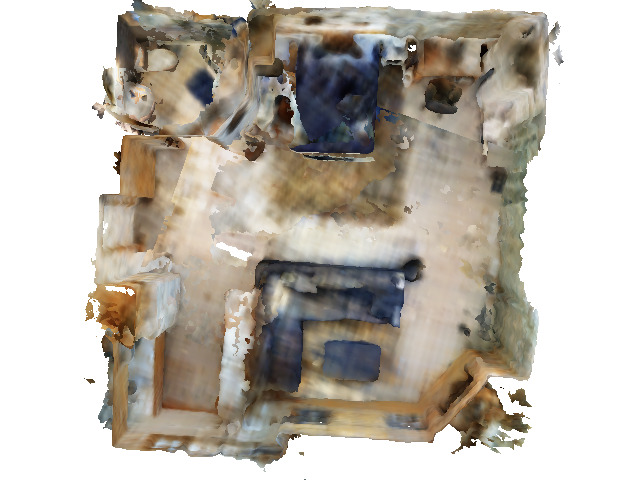} &
        \includegraphics[width=\linewidth,trim={3cm 0 3cm 0},clip]{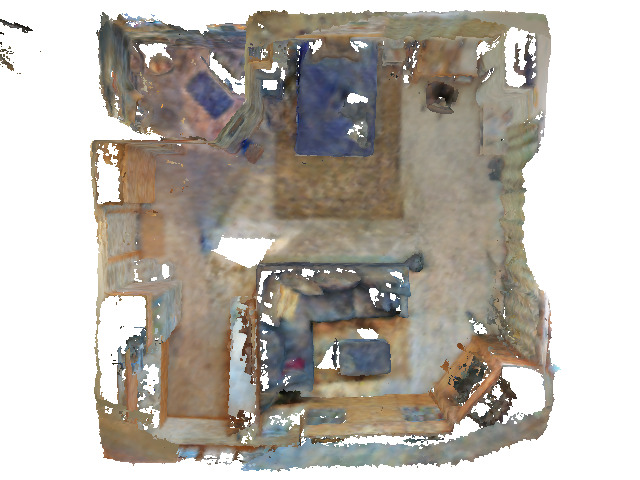} &
        \includegraphics[width=\linewidth,trim={3cm 0 3cm 0},clip]{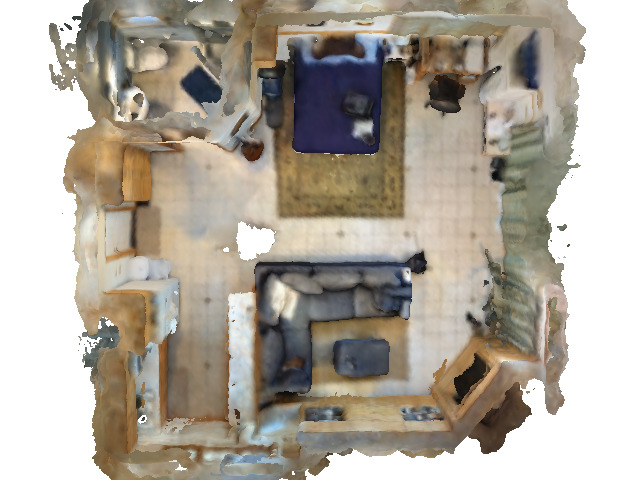} \\
        \rotatebox{90}{\texttt{scene0106\_00}} &
        \includegraphics[width=\linewidth,trim={0 2cm 0 1.5cm},clip]{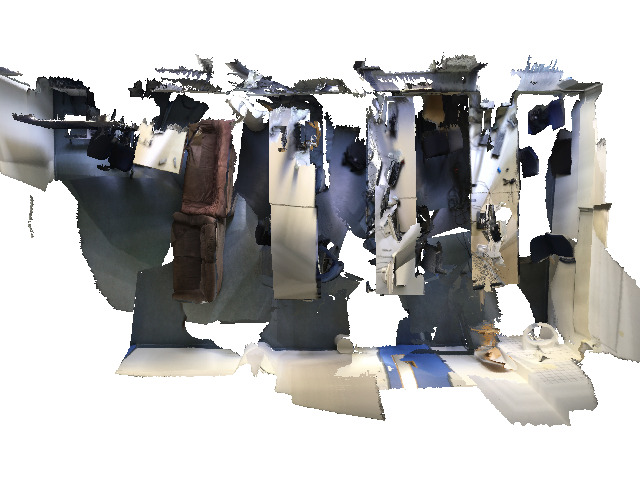} &
        \includegraphics[width=\linewidth,trim={0 2cm 0 1.5cm},clip]{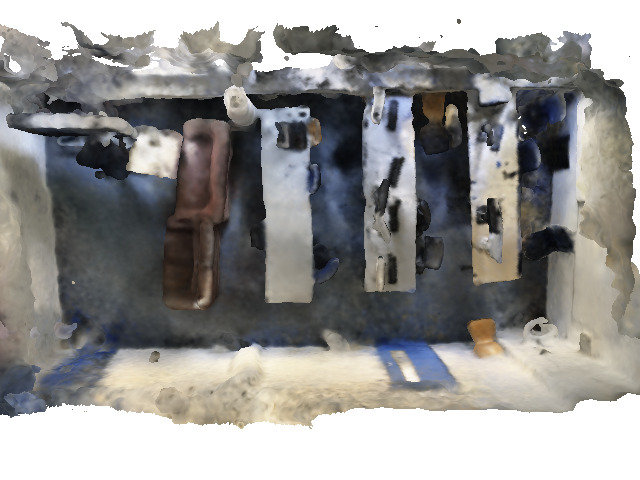} &
        \includegraphics[width=\linewidth,trim={0 2cm 0 1.5cm},clip]{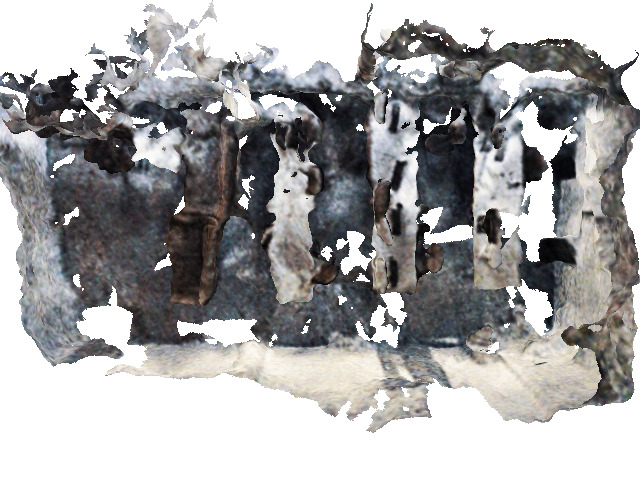} &
        \includegraphics[width=\linewidth,trim={0 2cm 0 1.5cm},clip]{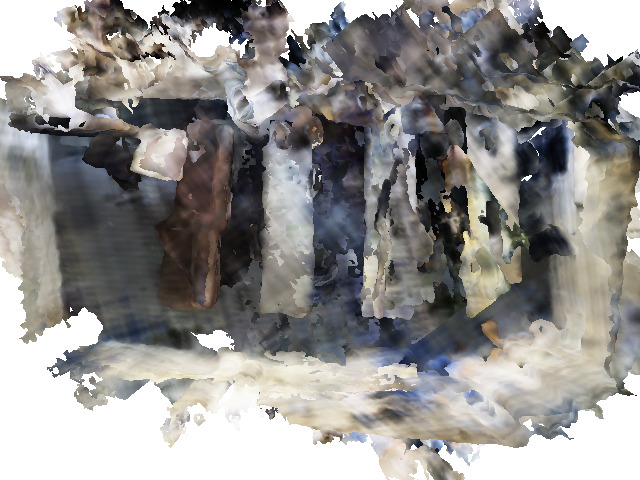} &
        \includegraphics[width=\linewidth,trim={0 2cm 0 1.5cm},clip]{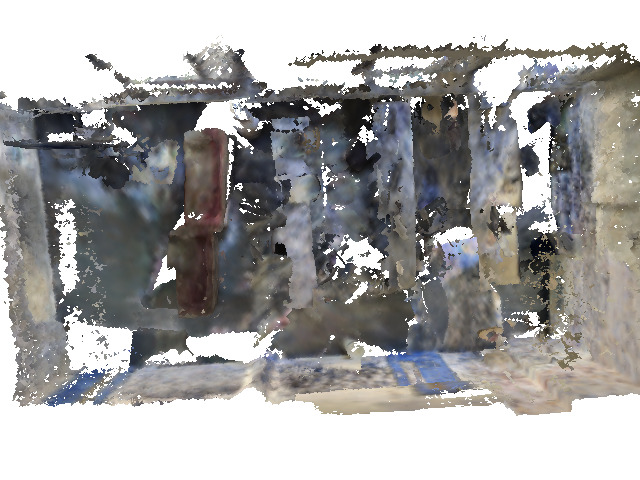} &
        \includegraphics[width=\linewidth,trim={0 2cm 0 1.5cm},clip]{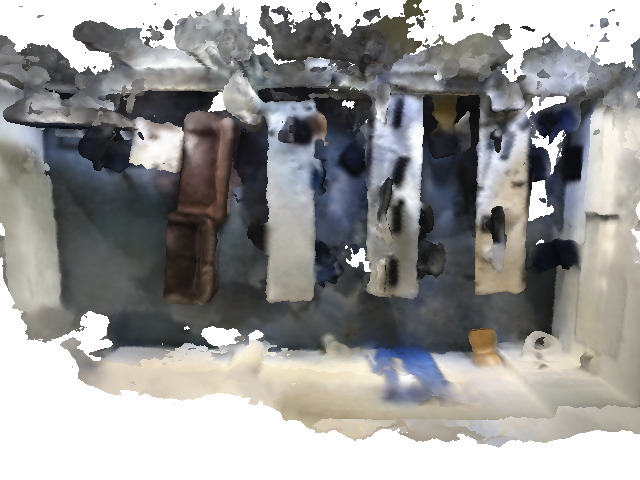} \\
        \rotatebox{90}{\texttt{scene0169\_00}} &
        \includegraphics[width=\linewidth,trim={0 1cm 0 1cm},clip]{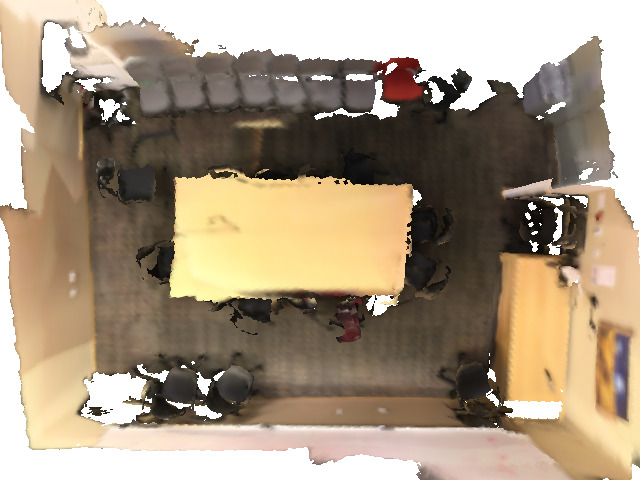} &
        \includegraphics[width=\linewidth,trim={0 1cm 0 1cm},clip]{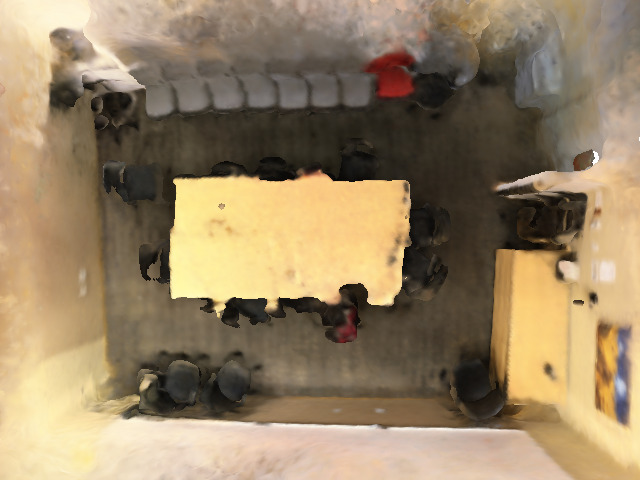} &
        \includegraphics[width=\linewidth,trim={0 1cm 0 1cm},clip]{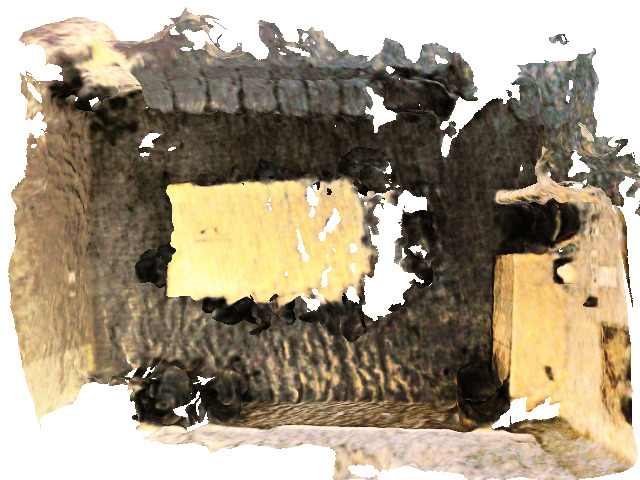} &
        \includegraphics[width=\linewidth,trim={0 1cm 0 1cm},clip]{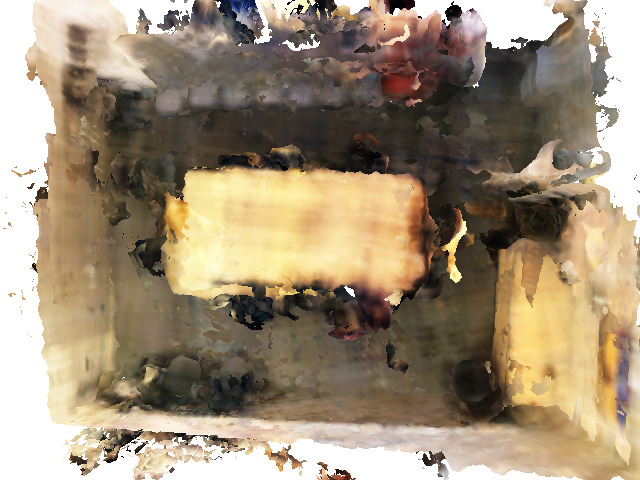} &
        \includegraphics[width=\linewidth,trim={0 1cm 0 1cm},clip]{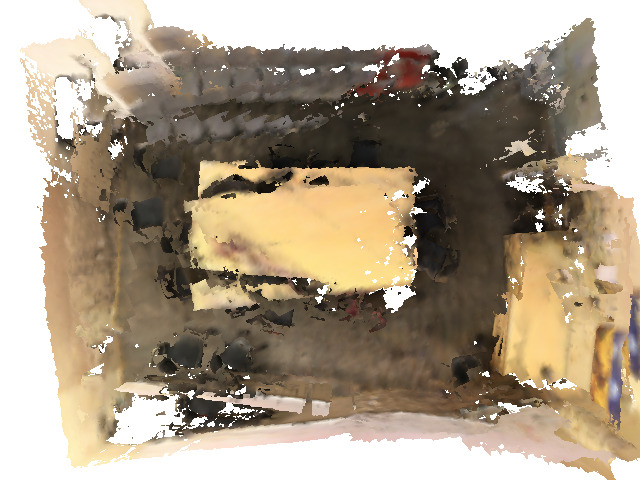} &
        \includegraphics[width=\linewidth,trim={0 1cm 0 1cm},clip]{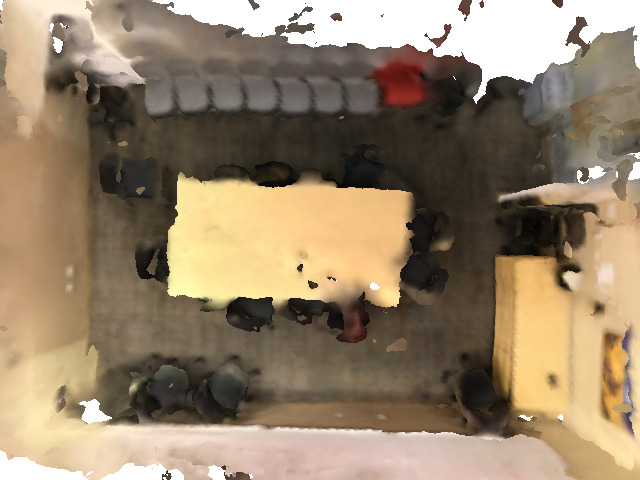} \\
    \end{tabular}
    \caption{Qualitative comparison of final reconstruction on Replica-Big and ScanNet. Alternative view is shown for significant failure cases. The submap-based methods MIPS-Fusion and Loopy-SLAM exhibit artifacts at submap borders and erroneous submap registrations. Our method allows submap-like map deformation at much finer granularity, while significantly reducing transition artifacts.}
    \label{fig:qualitative}
\end{figure*}

\cref{tab:meshresults} reports quantitative results on all datasets. Our method achieves state-of-the-art results on the smaller scenes (Replica, NRGBD) and outperforms existing methods on larger scenes (Replica-Big). The qualitative results on Replica-Big shown in \cref{fig:qualitative} further illustrate the differences. NICE-SLAM and Co-SLAM drift off and cannot correct for the drift upon returning to previous frames leading to poor results. GO-SLAM also benefiting from loop closures successfully maps large parts; however, fails in the last section of \texttt{apt0} and generally exhibits worse reconstruction quality. While GO-SLAM achieves quantitatively slightly better results as our method on \texttt{apt0}, we note that qualitatively our method exhibits better scene completion and higher fidelity overall\footnote{The upper floor has a slight consistent orientation error even after loop closure, which makes many walls fall outside the $5\,\mathrm{cm}$ threshold explaining the mismatch between perceived quality and evaluation result.}. The submap-based methods MIPS-Fusion and Loopy-SLAM fail to achieve globally consistent result on most scenes. Particularly Loopy-SLAM failed on all scenes of Replica-Big\footnote{We could not achieve a successful run on \texttt{apt0} without exceeding 24 GB of GPU memory due to too many created submaps.}. Our method and GO-SLAM benefit from their underlying SLAM systems ORB-SLAM2 and DROID-SLAM, respectively.

\cref{fig:qualitative} shows results on three scenes of the ScanNet dataset (the ``ground-truth'' was obtained using BundleFusion in this case \cite{dai2017bundlefusion}). The benefit of neural scene representations over classic representations in terms of scene completion can easily be observed. While less drift accumulates on these scenes, our method still achieves on par results to the best performing method Co-SLAM. The point-based Loopy-SLAM appears less robust to noisy measurements and creates more noisy, less complete reconstructions.

\cref{fig:kintinuous} shows additional results on the Kintinuous sequences before and after loop closure highlighting the efficient loop closure integration of our method. Because none of the other methods successfully closed the loop, we compare with the monolithic ablation of our approach (see \cref{sec:ablation} for further discussion).

\begin{figure*}[!htbp]%
    \centering%
    \subfloat[Multi-field representation]{%
    \resizebox{!}{5.1cm}{\input{figures/kintinuous.tikz}}%
    }\hfill
    \subfloat[Single-field representation]{
    \resizebox{!}{5.1cm}{\input{figures/kintinuous_mono.tikz}}%
    }%
    \caption{Qualitative results on the Kintinuous sequence for our multi-field representation and the ablated, monolithic variant. We show a top-down and two additional views of the mesh right before and after specific durations. Our multi-field representation adapts within just one frame with minimal loss in quality, whereas the monolithic variant does not reach the same quality even after $7\,\mathrm{s}$ of optimization.}
    \label{fig:kintinuous}
\end{figure*}

\subsection{Run-Time and Model Size Analysis}

\cref{tab:runtime_n_modelsize} reports run-time\footnote{We report time per frame by benchmarking the execution time for the whole sequence and dividing it by the number of frames. Mesh extraction has been deactivated for all methods for fair comparison. All evaluations were run with an Intel Core i9-13900KF and NVIDIA GeForce RTX 4090.} and model size on three representative scenes of different sizes. Our approach compares favorably in terms of run-time despite the overhead of parallelized training of multiple fields. GO-SLAM and Loopy-SLAM slow down significantly for longer sequences, whereas our method slows down the least of all evaluated methods (5\% difference between the smallest and largest scene). Model size is competitive with the submap-based baselines, but worse in comparison to monolithic methods. This might be due to the more constrained hash tables, which can be overparameterized for simple areas. %

Run-time results such as these are difficult to interpret due to varying implementation quality and large number of parameters. Especially NICE-SLAM and Loopy-SLAM are an order-of-magnitude slower than the other methods. Whether this is due to inefficient implementation or inherent complexity is unclear. Also, the baselines perform tracking and mapping, whereas our method uses precomputed ORB-SLAM results. However, ORB-SLAM is a CPU-only method and could therefore run in parallel without significantly affecting the run-time of our approach.

\begin{table}[!tb]
    \centering
    \renewcommand{\arraystretch}{1.1}
    \setlength{\tabcolsep}{5pt}
    \scriptsize
    \caption{Run-time and model size comparison of all methods. Our method compares favorably in terms of processing time compared to all baselines. The model size is larger than for the monolithic baselines, but competitive with the submap-based methods. Loopy-SLAM failed after 16 hours ($0.1\,\mathrm{Hz}$) on \texttt{apt0}.}
    \begin{tabular}{@{}rS[table-format=2.2]S[table-format=2.2]S[table-format=2.2]S[table-format=2.2]S[table-format=2.1]S[table-format=2.1]S[table-format=2.1]@{}}
        \toprule
            & \multicolumn{3}{c}{FPS ($\mathrm{Hz})$} & \multicolumn{3}{c}{Model Size ($\mathrm{MB}$)} \\
            \cmidrule(lr){2-4}
            \cmidrule(lr){5-7}
            & \texttt{br} & \texttt{ck} & \texttt{apt0}
           & \texttt{br} & \texttt{ck} & \texttt{apt0} \\
        \midrule
        NICE-SLAM \cite{zhu2022nice} & 2.74 & 2.70 & 2.25 & 5.3 & 21.6 & 57.1 \\
        Co-SLAM \cite{wang2023co} & 12.70 & 12.84 & 10.75 & 7.0 & 7.0 & 7.0 \\
        GO-SLAM \cite{zhang2023go} & 17.35 & 12.06 & 6.36 & 66.5 & 66.5 & 66.5 \\
        MIPS-Fusion \cite{tang2023mips} & 4.02 & 3.64 & 3.31 & 36.2 & 72.4 & 398.3 \\
        Loopy-SLAM \cite{liso2024loopy} & 0.38 & 0.36 & \text{\protect\xmark} & 21.9 & 48.1 & \text{\protect\xmark} \\
        Ours-SF & 19.78 & 18.92 & 18.84 & 8.4 & 8.4 & 8.4 \\
        Ours & 16.67 & 15.93 & 15.68 & 28.1 & 57.7 & 229.2 \\
        \bottomrule
    \end{tabular}
    \vspace*{-1\baselineskip}
    \label{tab:runtime_n_modelsize}
\end{table}

\paragraph{Field Radius and Hash Map Size} In \cref{fig:effect_of_params} (left) the memory requirement and F1-score for different field radii $r$ and hash table sizes $T$ is reported. In general, larger fields require larger hash maps to maintain the same quality. Our approach works across a wide range of $r$, but the ability to adapt to loop closures will be limited for larger $r$.

\begin{figure}[tb]%
    \centering%
    \scriptsize%
    \includegraphics{figures/radius_n_hm.tikz}%
    \hspace{-0.2cm}%
    \includegraphics{figures/iters_n_bs.tikz}%
    \caption{Left: reconstruction quality and model size for varying field radii $r$ and hash table sizes $T$. Right: reconstruction quality and run-time for varying number of iterations per frame $N_\mathrm{it}$ and number of rays per field $N_\mathrm{r}$.}%
    \label{fig:effect_of_params}
    \vspace*{-2\baselineskip}
\end{figure}

\paragraph{Batch Size and Iterations} In \cref{fig:effect_of_params} (right) the processing time per frame and F1-score for varying batch sizes and number of iterations is reported. Increasing the number of iterations per frame or the batch size leads to improved reconstruction quality. Notably, our approach maintains high reconstruction quality even at frame rates of more than $50\,\mathrm{Hz}$.

\subsection{Ablation Study}\label{sec:ablation}

\paragraph{Multi-Field Representation} In \cref{tab:meshresults} and \cref{tab:runtime_n_modelsize} a comparison to a single-field variant of our method to isolate the effect of the multi-field representation. This variant uses the same rendering, poses, and optimization scheme as the multi-field variant, but represents the scene by a global monolithic field with $T=2^{16}$. On the scenes with significant loop closures (\texttt{apt0} and \texttt{apt2}) the multi-field representation outperforms the monolithic variant. %

These results are at the end of the sequence and hence the monolithic variant has time to reintegrate with updated poses. \cref{fig:kintinuous} shows the map quality right after a large-scale loop closures. Our proposed multi-field representation adapts immediately to the pose adjustments keeping the map intact, while the single-field variant degrades significantly and only slowly reoptimizes without reaching the previous quality after more than $7\,\mathrm{s}$.

\paragraph{Multi-View Supervision} \cref{fig:multiview} compares optimization with targets sampled from all keyframes as described in \cref{sec:tsdfoptimization} to targets sampled from a single view. In the latter case, optimization alternates between a random previous keyframe and the latest keyframe. Single-view supervision shows local forgetting effects, particularly notable when a previously seen wall is observed from the other side. Multi-view optimization avoids this by combining previous and current observations in each optimization step.

\begin{figure}[tb!]
    \centering
    \input{figures/svmv.tikz}
    \caption{Single-view supervision leads to artifacts when the opposite side of a wall is observed. Multi-view supervision significantly reduces this local forgetting effect.}\label{fig:multiview}
    \vspace*{-1.2\baselineskip}
\end{figure}

\section{Conclusion}\label{sec:conclusion}

We have presented a neural mapping framework that anchors lightweight neural fields to the pose graph of a sparse visual SLAM system. This allows to incorporate loop closures into the volumetric map at near-zero cost achieving global map consistency over time. Our approach deforms the map reducing the need for reintegration while allowing geometric queries without requiring any space warping or image-space fusion. Compared to submap-based approaches, it deforms the map at a more granular level and reduces commonly observed artifacts at submap boundaries by avoiding hard assignments of frames to submaps.

\paragraph{Acknowledgments}This work was partially supported by the Wallenberg AI, Autonomous Systems and Software Program (WASP) funded by the Knut and Alice Wallenberg Foundation.

\FloatBarrier

{
    \small
    \bibliographystyle{ieee_fullname}
    \bibliography{main}
}

\input{supplementary_content}

\end{document}

%% file: preamble.tex
\usepackage[dvipsnames]{xcolor}

\DeclareFontShape{OMX}{cmex}{m}{b}{<-> cmexb10}{}
\SetSymbolFont{largesymbols}{bold}{OMX}{cmex}{m}{b}

\usepackage{times}
\usepackage{epsfig}
\usepackage{graphicx}
\usepackage{amsmath}
\usepackage{amssymb}
\usepackage{tensor}
\usepackage{tikz}
\usepackage{mathtools}
\usepackage{threeparttable}
\usepackage{siunitx} 
\usepackage{booktabs}
\usepackage{multirow}
\usepackage{placeins}
\usepackage{pgffor}
\usepackage{pgfplots}
\usepackage{tikzscale}
\usepackage{standalone}
\usepackage{bm}
\usepackage{colortbl}
\usepackage{pifont}
\usepackage{tabularray}

\pgfplotsset{compat=1.18}

\makeatletter
\renewcommand{\paragraph}{%
  \@startsection{paragraph}{4}%
  {\z@}{1.5ex \@plus 1ex \@minus .2ex}{-1em}%
  {\normalfont\normalsize\bfseries}%
}
\makeatother

\definecolor{gold}{RGB}{255,215,0}
\definecolor{silver}{RGB}{192,192,192}
\definecolor{bronze}{RGB}{204,153,102}

\newcommand{\xmark}{%
\tikz[scale=0.16,opacity=1] {
    \node[opacity=0, outer sep=0, inner sep=0, minimum width=0pt, minimum height=0pt] (text) at (0.5, 0.0) {\clap{\smash{\ding{55}}}};%
    \draw[line width=0.4,line cap=round] (0,0) to [bend left=6] (1,1);
    \draw[line width=0.4,line cap=round] (0.2,0.95) to [bend right=3] (0.8,0.05);
}}

\newcommand{\gold}{%
\tikz[scale=0.17, opacity=1] {
    \node[opacity=0, outer sep=0, inner sep=0, minimum width=0pt, minimum height=0pt] (text) at (0.5, 0.0) {\smash{\clap{O}}};%
    \draw[gold,rotate=-10, line width=0.4,line cap=round,fill] (0.5, 0.63) circle(0.39 and 0.45);
}}
\newcommand{\silver}{%
\tikz[scale=0.17, opacity=1] {
    \node[opacity=0, outer sep=0, inner sep=0, minimum width=0pt, minimum height=0pt] (text) at (0.5, 0.0) {\smash{\clap{O}}};%
    \draw[silver,rotate=-10, line width=0.4,line cap=round,fill] (0.5, 0.63) circle(0.39 and 0.45);
}}
\newcommand{\bronze}{%
\tikz[scale=0.17, opacity=1] {
    \node[opacity=0, outer sep=0, inner sep=0, minimum width=0pt, minimum height=0pt] (text) at (0.5, 0.0) {\smash{\clap{O}}};%
    \draw[bronze,rotate=-10, line width=0.4,line cap=round,fill] (0.5, 0.63) circle(0.39 and 0.45);
}}

\newcolumntype{L}[1]{>{\raggedright\let\newline\\\arraybackslash\hspace{0pt}}m{#1}}
\newcolumntype{C}[1]{>{\centering\let\newline\\\arraybackslash\hspace{0pt}}m{#1}}
\newcolumntype{R}[1]{>{\raggedleft\let\newline\\\arraybackslash\hspace{0pt}}m{#1}}

\usepgfplotslibrary{units}

\usetikzlibrary{patterns}
\usetikzlibrary{shadows}    
\usetikzlibrary{decorations.pathreplacing,calligraphy}
\usetikzlibrary{intersections}
\usetikzlibrary{calc,positioning,backgrounds}

\definecolor{mpltab20cb3}{RGB}{158,202,225}
\definecolor{mpltab20co3}{RGB}{253,174,107}
\definecolor{mpltab20cg3}{RGB}{161,217,155}
\definecolor{mpltab20cv3}{RGB}{188,189,220}

\definecolor{LightGray}{rgb}{0.96,0.96,0.96}

%% file: figures/fig1.tikz
\scriptsize
\begin{tikzpicture}
\node[inner sep=0pt, outer sep=0pt] (overview) at (0,0) {\includegraphics[width=4.3cm]{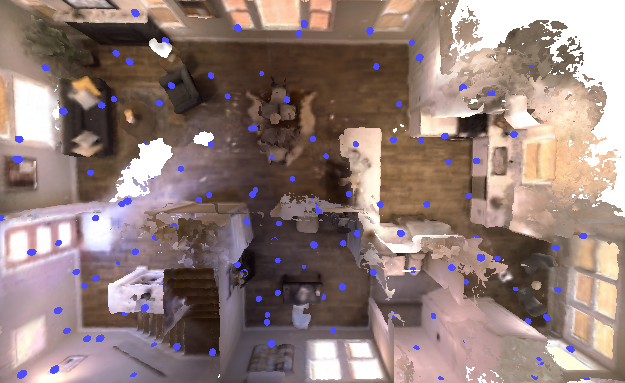}};

\coordinate (close1) at ($(overview.south west) + (0.5,-0.6)$);
\coordinate (close1anchor) at ($(close1) + (90:0.5)$);
\coordinate (close2) at ($(close1) + (1.1,0.0)$);
\coordinate (close2anchor) at ($(close2) + (120:0.5)$);
\coordinate (close3) at ($(close2) + (1.1,0.0)$);
\coordinate (close3anchor) at ($(close3) + (110:0.5)$);
\coordinate (close4) at ($(close3) + (1.1,0.0)$);
\coordinate (close4anchor) at ($(close4) + (90:0.5)$);

\begin{scope}
    \clip (close1) circle (0.5);
    \node[anchor=center, inner sep=0pt, outer sep=0pt] at ($(close1) + (-0.07,-0.08)$) {\includegraphics[width=0.89cm,trim=0 0 0 0,clip]{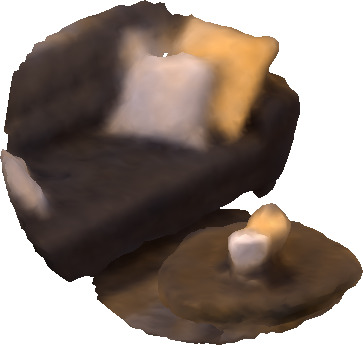}};
\end{scope}
\node[draw, circle,minimum size=1cm, inner sep=0pt, outer sep=0pt, color=orange] (close1circle) at (close1) {};

\begin{scope}
    \clip (close2) circle (0.5);
    \node[anchor=center, inner sep=0pt, outer sep=0pt] at ($(close2) + (0.02,-0.05)$) {\includegraphics[width=0.97cm,trim=0 0 0 0,clip]{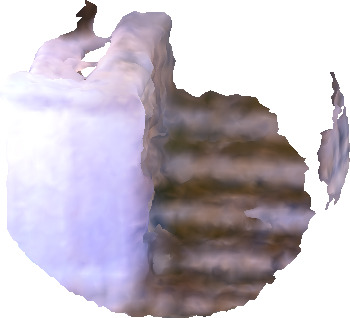}};
\end{scope}
\node[draw, circle,minimum size=1cm, inner sep=0pt, outer sep=0pt, color=orange] (close2circle) at (close2) {};

\begin{scope}
    \clip (close3) circle (0.5);
    \node[anchor=center, inner sep=0pt, outer sep=0pt] at ($(close3) + (-0.1,-0.1)$) {\includegraphics[width=0.7cm,trim=0 0 0 0,clip]{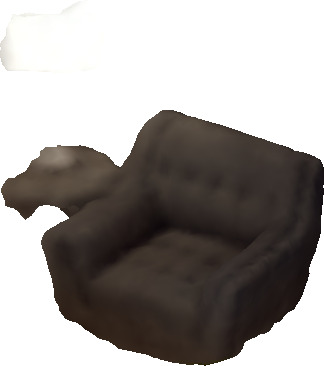}};  
\end{scope}
\node[draw, circle,minimum size=1cm, inner sep=0pt, outer sep=0pt, color=orange] (close3circle) at (close3) {};

\begin{scope}
    \node[anchor=center, inner sep=0pt, outer sep=0pt] at ($(close4) + (0.00,0.0)$) {\includegraphics[width=1cm,trim=0 0 0 0,clip]{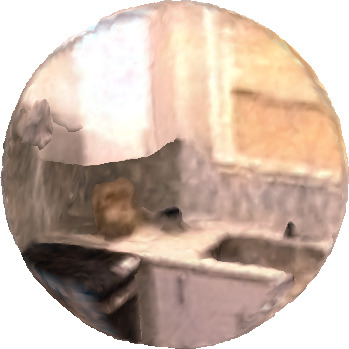}};  
\end{scope}
\node[draw, circle,minimum size=1cm, inner sep=0pt, outer sep=0pt, color=orange] (close4circle) at (close4) {};

\node[draw, color=orange, circle,minimum size=0.1cm, inner sep=0pt, outer sep=0pt] (close1over) at (-1.45,0.595) {};
\draw[color=orange] (close1over) -- (close1circle);
\node[draw, color=orange, circle,minimum size=0.1cm, inner sep=0pt, outer sep=0pt] (close2over) at (-1.29,-0.84) {};
\draw[color=orange] (close2over) -- (close2circle);
\node[draw, color=orange, circle,minimum size=0.1cm, inner sep=0pt, outer sep=0pt] (close3over) at (-0.97,0.73) {};
\draw[color=orange] (close3over) -- (close3circle);
\node[draw, color=orange, circle,minimum size=0.1cm, inner sep=0pt, outer sep=0pt] (close4over) at (1.385,0.3925) {};
\draw [color=orange](close4over) -- (close4circle);

\end{tikzpicture}%

%% file: figures/framework.tex
\scriptsize
\begin{tikzpicture}[outer sep=0pt]
\makeatletter
\tikzset{scale line widths/.style={%
/utils/exec=\def\tikz@semiaddlinewidth##1{%
\pgfgettransformentries{\tmpa}{\tmpb}{\tmpc}{\tmpd}{\tmp}{\tmp}%
\pgfmathsetmacro{\myJacobian}{sqrt(abs(\tmpa*\tmpd-\tmpb*\tmpc))}%
\pgfmathsetlength\pgflinewidth{\myJacobian*0.4pt}%
\pgfmathsetmacro{\my@lw}{\myJacobian*##1}%
\tikz@addoption{\pgfsetlinewidth{\my@lw pt}}\pgfmathsetlength\pgflinewidth{\my@lw pt}},%
thin}}
\makeatother

\tikzset{
    invclip/.style={
        clip,
        insert path={{[reset cm](-16383.99999pt,-16383.99999pt) rectangle (16383.99999pt,16383.99999pt)}}
    }
}

\coordinate (kfposes) at (1.0, 0);

\coordinate (opt) at (0.0, 0.0);
\begin{scope}[shift=(opt), local bounding box=optbb]
    \path[fill=mpltab20co3, rounded corners] (0,0) rectangle (6.6, 3.35);
\end{scope}
\node[anchor=south west] at (optbb.south west) {Independent per-field multi-view optimization (Sec.\ \ref{sec:tsdfoptimization})};

\coordinate (mfsr) at ($(optbb.south east) + (0.3, 0.0)$);
\begin{scope}[shift=(mfsr), local bounding box=mfsrbb]
    \path[fill=mpltab20cb3, rounded corners] (0,0) rectangle (7.8, 4.7);
\end{scope}
\node[anchor=south west] at (mfsrbb.south west) {Multi-field scene representation (Sec. \ref{sec:mfsr})};

\coordinate (query) at ($(mfsrbb.south east) + (0.3, 0.0)$);
\begin{scope}[fill=mpltab20cb3, shift=(query), local bounding box=querybb]
    \path[fill=mpltab20cg3, rounded corners] (0,0) rectangle (2.3, 4.7);
\end{scope}
\node[anchor=south west] at (querybb.south west) {Queries (Sec. \ref{sec:queries})};

\coordinate (mesh) at ($(querybb.north west) + (0.15, -0.15)$);
\begin{scope}[shift=(mesh), local bounding box=meshbb]
    \path[fill=white,rounded corners] (0,0) rectangle (2, -2);
    \node[anchor=north west] at (0,0) {Mesh extraction};
\end{scope}

\coordinate (novel) at ($(meshbb.south west) + (0.0, -0.15)$);
\begin{scope}[shift=(novel)]
    \path[fill=white,rounded corners] (0,0) rectangle (2, -2);
    \node[anchor=north west] at (0,0) {View synthesis};
\end{scope}

\coordinate (mfsrbg) at ($(mfsr.south west) + (0.15, 0.4)$);
\begin{scope}[shift=(mfsrbg)]
    \path[fill=white,rounded corners] (0,0) rectangle (7.5, 3.1);
\end{scope}

\node[fill=white, rounded corners, draw=none, anchor=north west, align=center, minimum height=0.9cm, minimum width=3.01cm] (fields) at ($(mfsrbb.north west) + (0.15, -0.15)$) {Posed neural fields\\ $\mathcal{F}_t=\{(f_i, \tensor[^{\mathrm{w}}]{\mathbf{T}}{_{i}})\}$};

\node[fill=white, rounded corners, draw=none, anchor=west, align=center, minimum height=0.9cm] (kfbuffer) at ($(fields.east) + (0.15,0.0)$) {Keyframes\\ $\mathcal{K}_t=\{(\mathbf{C}_k,\mathbf{D}_k)\}$};

\node[fill=white, rounded corners, draw=none, align=center,minimum height=0.9cm, minimum width=1.9cm, anchor=west] (kfposes) at ($(kfbuffer.east) + (0.15,0.0)$) {Keyframe poses\\ $\mathcal{T}_t=\{\tensor[^{\mathrm{w}}]{\mathbf{T}}{_{k}}\}$};

\node[rounded corners, draw, align=center, anchor=south west, minimum height=1.05cm, minimum width=1.2cm] (rgbd) at ($(optbb.north west) + (0.0,0.3)$) {RGB-D\\frame};

\coordinate (pgslam) at ($(rgbd.south east) + (0.3,0.0)$);
\begin{scope}[local bounding box=pgslambb]
    \draw[rounded corners] (pgslam) rectangle ++(2.9, 1.05);
\end{scope}
\node[anchor=south west] at (pgslam) {Pose graph-based SLAM};

\node[rounded corners, draw, align=center, anchor=north east] (iskf) at ($(pgslambb.north east) + (-0.2, -0.15)$) {Is keyframe?};

\node[fill=mpltab20cv3, rounded corners, draw=none, align=center, anchor=south east, minimum height=1.05cm, minimum width=1.8cm] (inf) at ($(optbb.north east) + (0.0, 0.3)$) {Instantiate new \\ fields (Sec. \ref{sec:ins})};

\draw[thick, ->] (rgbd) -> (pgslambb);
\draw[thick, ->] (iskf.east) -- (inf.west |- iskf.east);
\draw[thick, ->] (inf.east) -- (fields.west |- inf.east);
\draw[thick, <->] (mfsrbb.west |- optbb.east) -- (optbb.east);
\draw[thick, ->] (mfsrbb.east |- optbb.east) -- (querybb.west |- optbb.east);
\draw[thick, ->] (iskf.east) to [out=0, in=180] ($(inf.north west) + (0.15, 0.15)$) -- ++ (4.5, 0.0) to [out=0, in=135] ($(kfbuffer.north west) + (0.04, -0.04)$);
\draw[thick, ->] ($(pgslambb.north east) + (-0.04, -0.04)$) to [out=45, in=180] ($(inf.north west) + (0.15, 0.3)$) -- ++ (6.0, 0.0) to [out=0, in=135] ($(kfposes.north west) + (0.04, -0.04)$);

\coordinate (pg1) at ($(mfsrbg) + (2.22,1.05)$);
\coordinate (pg2) at ($(pg1) + (4.3,0.0)$);

\begin{scope}[rotate=90, scale=0.4, scale line widths=0.4, shift=(pg1), local bounding box=pg1bb]
    \coordinate (n0) at (0.0,0.0);
    \coordinate (n1) at (-0.3,1.0);
    \coordinate (n2) at (0.3,2.2);
    \coordinate (n3) at (1.2,3.0);
    \coordinate (n4) at (2.3,2.5);
    \coordinate (n5) at (2.6,1.3);
    \coordinate (n6) at (2.4,0.3);
    \coordinate (render) at (1.7,3);

    \draw (n0) -- (n1) -- (n2) -- (n3) -- (n4) -- (n5) -- (n6);
    
    \begin{scope}[rotate=190, shift=(n0)]
        \coordinate (f1) at (1.2,-0.3);
        \coordinate (f2) at (1.3,-0.8);
        \coordinate (f3) at (1.2,0.707);
    
        \path[thick, fill=blue!70, fill opacity=0.2] (f1) circle (0.707);
        \path[thick, fill=blue!70, fill opacity=0.2] (f2) circle (0.707);
        \path[thick, fill=blue!70, fill opacity=0.2] (f3) circle (0.707);
    
        \path[draw, fill, blue!70] (f1) circle (0.05);
        \path[draw, fill, blue!70] (f2) circle (0.05);
        \path[draw, fill, blue!70] (f3) circle (0.05);
    
        \draw[blue!70] (0,0) -- (f1);
        \draw[blue!70] (0,0) -- (f2);
        \draw[blue!70] (0,0) -- (f3);
        
        \path[draw, fill, thick] (0,0) circle (0.05);
        \draw[thick] (0,0) -- (0.3,0.15) --++ (0, -0.3) -- cycle;
    \end{scope}
    
    \begin{scope}[rotate=140, shift=(n1)]
        \coordinate (f1) at (1.2,-0.3);
        \coordinate (f2) at (1.2,0.707);
    
        \path[thick, fill=blue!70, fill opacity=0.2] (f1) circle (0.707);
        \path[thick, fill=blue!70, fill opacity=0.2] (f2) circle (0.707);
    
        \path[draw, fill, blue!70] (f1) circle (0.05);
        \path[draw, fill, blue!70] (f2) circle (0.05);
    
        \draw[blue!70] (0,0) -- (f1);
        \draw[blue!70] (0,0) -- (f2);
        
        \path[draw, fill, thick] (0,0) circle (0.05);
        \draw[thick] (0,0) -- (0.3,0.15) --++ (0, -0.3) -- cycle;
    \end{scope}
    
    \begin{scope}[rotate=140, shift=(n2)]
        \coordinate (f1) at (1,0.0);
        \coordinate (f2) at (1.3,-0.8);
        \coordinate (f3) at (1.2,0.707);
    
        \path[thick, fill=blue!70, fill opacity=0.2] (f1) circle (0.707);
        \path[thick, fill=blue!70, fill opacity=0.2] (f2) circle (0.707);
        \path[thick, fill=blue!70, fill opacity=0.2] (f3) circle (0.707);
    
        \path[draw, fill, blue!70] (f1) circle (0.05);
        \path[draw, fill, blue!70] (f2) circle (0.05);
        \path[draw, fill, blue!70] (f3) circle (0.05);
    
        \draw[blue!70] (0,0) -- (f1);
        \draw[blue!70] (0,0) -- (f2);
        \draw[blue!70] (0,0) -- (f3);
        
        \path[draw, fill, thick] (0,0) circle (0.05);
        \draw[thick] (0,0) -- (0.3,0.15) --++ (0, -0.3) -- cycle;
    \end{scope}
    
    \begin{scope}[rotate=100, shift=(n3)]
        \coordinate (f1) at (1.1,0.1);
        \coordinate (f2) at (1.3,-0.8);
        \coordinate (f3) at (1.2,0.707);
    
        \path[thick, fill=blue!70, fill opacity=0.2] (f1) circle (0.707);
        \path[thick, fill=blue!70, fill opacity=0.2] (f2) circle (0.707);
        \path[thick, fill=blue!70, fill opacity=0.2] (f3) circle (0.707);
    
        \path[draw, fill, blue!70] (f1) circle (0.05);
        \path[draw, fill, blue!70] (f2) circle (0.05);
        \path[draw, fill, blue!70] (f3) circle (0.05);
    
        \draw[blue!70] (0,0) -- (f1);
        \draw[blue!70] (0,0) -- (f2);
        \draw[blue!70] (0,0) -- (f3);
        
        \path[draw, fill, thick] (0,0) circle (0.05);
        \draw[thick] (0,0) -- (0.3,0.15) --++ (0, -0.3) -- cycle;
    \end{scope}
    
    \begin{scope}[rotate=50, shift=(n4)]
        \coordinate (f1) at (1.2,-0.3);
        \coordinate (f2) at (1.3,-0.8);
        \coordinate (f3) at (1.2,0.707);
    
        \path[thick, fill=blue!70, fill opacity=0.2] (f1) circle (0.707);
        \path[thick, fill=blue!70, fill opacity=0.2] (f2) circle (0.707);
        \path[thick, fill=blue!70, fill opacity=0.2] (f3) circle (0.707);
    
        \path[draw, fill, blue!70] (f1) circle (0.05);
        \path[draw, fill, blue!70] (f2) circle (0.05);
        \path[draw, fill, blue!70] (f3) circle (0.05);
    
        \draw[blue!70] (0,0) -- (f1);
        \draw[blue!70] (0,0) -- (f2);
        \draw[blue!70] (0,0) -- (f3);
        
        \path[draw, fill, thick] (0,0) circle (0.05);
        \draw[thick] (0,0) -- (0.3,0.15) --++ (0, -0.3) -- cycle;
    \end{scope}
    
    \begin{scope}[rotate=20, shift=(n5)]
        \coordinate (f1) at (1.1,-0.4);
        \coordinate (f2) at (1.2,0.707);
        \coordinate (f3) at (1.3,-0.9);
    
        \path[thick, fill=blue!70, fill opacity=0.2] (f1) circle (0.707);
        \path[thick, fill=blue!70, fill opacity=0.2] (f2) circle (0.707);
        \path[thick, fill=blue!70, fill opacity=0.2] (f3) circle (0.707);
    
        \path[draw, fill, blue!70] (f1) circle (0.05);
        \path[draw, fill, blue!70] (f2) circle (0.05);
        \path[draw, fill, blue!70] (f3) circle (0.05);
    
        \draw[blue!70] (0,0) -- (f1);
        \draw[blue!70] (0,0) -- (f2);
        \draw[blue!70] (0,0) -- (f3);
        
        \path[draw, fill, thick] (0,0) circle (0.05);
        \draw[thick] (0,0) -- (0.3,0.15) --++ (0, -0.3) -- cycle;
    \end{scope}
    
    \begin{scope}[rotate=-10, shift=(n6)]
        \coordinate (f1) at (1.1,-0.4);
        \coordinate (f2) at (1.2,0.707);
        \coordinate (f3) at (1.3,-0.9);
    
        \path[thick, fill=blue!70, fill opacity=0.2] (f1) circle (0.707);
        \path[thick, fill=blue!70, fill opacity=0.2] (f2) circle (0.707);
        \path[thick, fill=blue!70, fill opacity=0.2] (f3) circle (0.707);
    
        \path[draw, fill, blue!70] (f1) circle (0.05);
        \path[draw, fill, blue!70] (f2) circle (0.05);
        \path[draw, fill, blue!70] (f3) circle (0.05);
    
        \draw[blue!70] (0,0) -- (f1);
        \draw[blue!70] (0,0) -- (f2);
        \draw[blue!70] (0,0) -- (f3);
        
        \path[draw, fill, thick] (0,0) circle (0.05);
        \draw[thick] (0,0) -- (0.3,0.15) --++ (0, -0.3) -- cycle;
    \end{scope}
\end{scope}

\begin{scope}[rotate=90, scale=0.4, scale line widths=0.4, shift=(pg2)]
    \coordinate (n0) at (0.0,0.0);
    \coordinate (n1) at (-0.3,1.0);
    \coordinate (n2) at (0.3,2.2);
    \coordinate (n3) at (1.2,3.0);
    \coordinate (n4) at (1.9,2.4);
    \coordinate (n5) at (1.8,1.2);
    \coordinate (n6) at (1,0.3);
    
    \draw (n0) -- (n1) -- (n2) -- (n3) -- (n4) -- (n5) -- (n6);

    \draw[red!80!black, thick] (n0) -- (n6);
    
    \begin{scope}[rotate=190, shift=(n0)]
        \coordinate (f1) at (1.2,-0.3);
        \coordinate (f2) at (1.3,-0.8);
        \coordinate (f3) at (1.2,0.707);
    
        \path[thick, fill=blue!70, fill opacity=0.2] (f1) circle (0.707);
        \path[thick, fill=blue!70, fill opacity=0.2] (f2) circle (0.707);
        \path[thick, fill=blue!70, fill opacity=0.2] (f3) circle (0.707);
    
        \path[draw, fill, blue!70] (f1) circle (0.05);
        \path[draw, fill, blue!70] (f2) circle (0.05);
        \path[draw, fill, blue!70] (f3) circle (0.05);
    
        \draw[blue!70] (0,0) -- (f1);
        \draw[blue!70] (0,0) -- (f2);
        \draw[blue!70] (0,0) -- (f3);
        
        \path[draw, fill, thick] (0,0) circle (0.05);
        \draw[thick] (0,0) -- (0.3,0.15) --++ (0, -0.3) -- cycle;
    \end{scope}
    
    \begin{scope}[rotate=140, shift=(n1)]
        \coordinate (f1) at (1.2,-0.3);
        \coordinate (f2) at (1.2,0.707);
    
        \path[thick, fill=blue!70, fill opacity=0.2] (f1) circle (0.707);
        \path[thick, fill=blue!70, fill opacity=0.2] (f2) circle (0.707);
    
        \path[draw, fill, blue!70] (f1) circle (0.05);
        \path[draw, fill, blue!70] (f2) circle (0.05);
    
        \draw[blue!70] (0,0) -- (f1);
        \draw[blue!70] (0,0) -- (f2);
        
        \path[draw, fill, thick] (0,0) circle (0.05);
        \draw[thick] (0,0) -- (0.3,0.15) --++ (0, -0.3) -- cycle;
    \end{scope}
    
    \begin{scope}[rotate=140, shift=(n2)]
        \coordinate (f1) at (1,0.0);
        \coordinate (f2) at (1.3,-0.8);
        \coordinate (f3) at (1.2,0.707);
    
        \path[thick, fill=blue!70, fill opacity=0.2] (f1) circle (0.707);
        \path[thick, fill=blue!70, fill opacity=0.2] (f2) circle (0.707);
        \path[thick, fill=blue!70, fill opacity=0.2] (f3) circle (0.707);
    
        \path[draw, fill, blue!70] (f1) circle (0.05);
        \path[draw, fill, blue!70] (f2) circle (0.05);
        \path[draw, fill, blue!70] (f3) circle (0.05);
    
        \draw[blue!70] (0,0) -- (f1);
        \draw[blue!70] (0,0) -- (f2);
        \draw[blue!70] (0,0) -- (f3);
        
        \path[draw, fill, thick] (0,0) circle (0.05);
        \draw[thick] (0,0) -- (0.3,0.15) --++ (0, -0.3) -- cycle;
    \end{scope}
    
    \begin{scope}[rotate=100, shift=(n3)]
        \coordinate (f1) at (1.1,0.1);
        \coordinate (f2) at (1.3,-0.8);
        \coordinate (f3) at (1.2,0.707);
    
        \path[thick, fill=blue!70, fill opacity=0.2] (f1) circle (0.707);
        \path[thick, fill=blue!70, fill opacity=0.2] (f2) circle (0.707);
        \path[thick, fill=blue!70, fill opacity=0.2] (f3) circle (0.707);
    
        \path[draw, fill, blue!70] (f1) circle (0.05);
        \path[draw, fill, blue!70] (f2) circle (0.05);
        \path[draw, fill, blue!70] (f3) circle (0.05);
    
        \draw[blue!70] (0,0) -- (f1);
        \draw[blue!70] (0,0) -- (f2);
        \draw[blue!70] (0,0) -- (f3);
        
        \path[draw, fill, thick] (0,0) circle (0.05);
        \draw[thick] (0,0) -- (0.3,0.15) --++ (0, -0.3) -- cycle;
    \end{scope}
    
    \begin{scope}[rotate=25, shift=(n4)]
        \coordinate (f1) at (1.2,-0.3);
        \coordinate (f2) at (1.3,-0.8);
        \coordinate (f3) at (1.2,0.707);
    
        \path[thick, fill=blue!70, fill opacity=0.2] (f1) circle (0.707);
        \path[thick, fill=blue!70, fill opacity=0.2] (f2) circle (0.707);
        \path[thick, fill=blue!70, fill opacity=0.2] (f3) circle (0.707);
    
        \path[draw, fill, blue!70] (f1) circle (0.05);
        \path[draw, fill, blue!70] (f2) circle (0.05);
        \path[draw, fill, blue!70] (f3) circle (0.05);
    
        \draw[blue!70] (0,0) -- (f1);
        \draw[blue!70] (0,0) -- (f2);
        \draw[blue!70] (0,0) -- (f3);
        
        \path[draw, fill, thick] (0,0) circle (0.05);
        \draw[thick] (0,0) -- (0.3,0.15) --++ (0, -0.3) -- cycle;
    \end{scope}
    
    \begin{scope}[rotate=-20, shift=(n5)]
        \coordinate (f1) at (1.1,-0.4);
        \coordinate (f2) at (1.2,0.707);
        \coordinate (f3) at (1.3,-0.9);
    
        \path[thick, fill=blue!70, fill opacity=0.2] (f1) circle (0.707);
        \path[thick, fill=blue!70, fill opacity=0.2] (f2) circle (0.707);
        \path[thick, fill=blue!70, fill opacity=0.2] (f3) circle (0.707);
    
        \path[draw, fill, blue!70] (f1) circle (0.05);
        \path[draw, fill, blue!70] (f2) circle (0.05);
        \path[draw, fill, blue!70] (f3) circle (0.05);
    
        \draw[blue!70] (0,0) -- (f1);
        \draw[blue!70] (0,0) -- (f2);
        \draw[blue!70] (0,0) -- (f3);
        
        \path[draw, fill, thick] (0,0) circle (0.05);
        \draw[thick] (0,0) -- (0.3,0.15) --++ (0, -0.3) -- cycle;
    \end{scope}
    
    \begin{scope}[rotate=-90, shift=(n6)]
        \coordinate (f1) at (1.2,-0.3);
        \coordinate (f2) at (1.1,0.707);
        \coordinate (f3) at (1.3,-0.9);
    
        \path[thick, fill=blue!70, fill opacity=0.2] (f1) circle (0.707);
        \path[thick, fill=blue!70, fill opacity=0.2] (f2) circle (0.707);
        \path[thick, fill=blue!70, fill opacity=0.2] (f3) circle (0.707);
    
        \path[draw, fill, blue!70] (f1) circle (0.05);
        \path[draw, fill, blue!70] (f2) circle (0.05);
        \path[draw, fill, blue!70] (f3) circle (0.05);
    
        \draw[blue!70] (0,0) -- (f1);
        \draw[blue!70] (0,0) -- (f2);
        \draw[blue!70] (0,0) -- (f3);
        
        \path[draw, fill, thick] (0,0) circle (0.05);
        \draw[thick] (0,0) -- (0.3,0.15) --++ (0, -0.3) -- cycle;
    \end{scope}
\end{scope}

\draw[->] (pg1bb.east) -- ++ (1.5,0) node[midway, above, align=center] (arrow) {$\mathcal{T}_{t-1} \to \mathcal{T}_{t}$};

\node[draw=blue!70, text=black, rounded corners, anchor=north, align=center, minimum height=0.6cm, thick] (fields) at ($(arrow) +(0,-0.3)$) {Anchoring\\$p(i)$ (\cref{sec:ins})};

\coordinate (meshextract) at ($(mesh) + (1.35,-1.3)$);
\begin{scope}[rotate=90, scale=0.2, scale line widths=0.2, shift=(meshextract)]
    \coordinate (n0) at (0.0,0.0);
    \coordinate (n1) at (-0.3,1.0);
    \coordinate (n2) at (0.3,2.2);
    \coordinate (n3) at (1.2,3.0);
    \coordinate (n4) at (1.9,2.4);
    \coordinate (n5) at (1.8,1.2);
    \coordinate (n6) at (1,0.3);

    \begin{scope}[rotate=190, shift=(n0)]
        \coordinate (f1) at (1.2,-0.3);
        \coordinate (f2) at (1.3,-0.8);
        \coordinate (f3) at (1.2,0.707);
    
        \path[thick, fill=blue!70, fill opacity=0.2] (f1) circle (0.707);
        \path[thick, fill=blue!70, fill opacity=0.2] (f2) circle (0.707);
        \path[thick, fill=blue!70, fill opacity=0.2] (f3) circle (0.707);
    
        \path[draw, fill, blue!70] (f1) circle (0.05);
        \path[draw, fill, blue!70] (f2) circle (0.05);
        \path[draw, fill, blue!70] (f3) circle (0.05);
    \end{scope}
    
    \begin{scope}[rotate=140, shift=(n1)]
        \coordinate (f4) at (1.2,-0.3);
        \coordinate (f5) at (1.2,0.707);
    
        \path[thick, fill=blue!70, fill opacity=0.2] (f4) circle (0.707);
        \path[thick, fill=blue!70, fill opacity=0.2] (f5) circle (0.707);
    
        \path[draw, fill, blue!70] (f4) circle (0.05);
        \path[draw, fill, blue!70] (f5) circle (0.05);
    \end{scope}
    
    \begin{scope}[rotate=140, shift=(n2)]
        \coordinate (f6) at (1,0.0);
        \coordinate (f7) at (1.3,-0.8);
        \coordinate (f8) at (1.2,0.707);
    
        \path[thick, fill=blue!70, fill opacity=0.2] (f6) circle (0.707);
        \path[thick, fill=blue!70, fill opacity=0.2] (f7) circle (0.707);
        \path[thick, fill=blue!70, fill opacity=0.2] (f8) circle (0.707);
    
        \path[draw, fill, blue!70] (f6) circle (0.05);
        \path[draw, fill, blue!70] (f7) circle (0.05);
        \path[draw, fill, blue!70] (f8) circle (0.05);
    \end{scope}
    
    \begin{scope}[rotate=100, shift=(n3)]
        \coordinate (f9) at (1.1,0.1);
        \coordinate (f10) at (1.3,-0.8);
        \coordinate (f11) at (1.2,0.707);
    
        \path[thick, fill=blue!70, fill opacity=0.2] (f9) circle (0.707);
        \path[thick, fill=blue!70, fill opacity=0.2] (f10) circle (0.707);
        \path[thick, fill=blue!70, fill opacity=0.2] (f11) circle (0.707);
    
        \path[draw, fill, blue!70] (f9) circle (0.05);
        \path[draw, fill, blue!70] (f10) circle (0.05);
        \path[draw, fill, blue!70] (f11) circle (0.05);
    \end{scope}
    
    \begin{scope}[rotate=25, shift=(n4)]
        \coordinate (f12) at (1.2,-0.3);
        \coordinate (f13) at (1.3,-0.8);
        \coordinate (f14) at (1.2,0.707);
    
        \path[thick, fill=blue!70, fill opacity=0.2] (f12) circle (0.707);
        \path[thick, fill=blue!70, fill opacity=0.2] (f13) circle (0.707);
        \path[thick, fill=blue!70, fill opacity=0.2] (f14) circle (0.707);
    
        \path[draw, fill, blue!70] (f12) circle (0.05);
        \path[draw, fill, blue!70] (f13) circle (0.05);
        \path[draw, fill, blue!70] (f14) circle (0.05);
    \end{scope}
    
    \begin{scope}[rotate=-20, shift=(n5)]
        \coordinate (f15) at (1.1,-0.4);
        \coordinate (f16) at (1.2,0.707);
        \coordinate (f17) at (1.3,-0.9);
    
        \path[thick, fill=blue!70, fill opacity=0.2] (f15) circle (0.707);
        \path[thick, fill=blue!70, fill opacity=0.2] (f16) circle (0.707);
        \path[thick, fill=blue!70, fill opacity=0.2] (f17) circle (0.707);
    
        \path[draw, fill, blue!70] (f15) circle (0.05);
        \path[draw, fill, blue!70] (f16) circle (0.05);
        \path[draw, fill, blue!70] (f17) circle (0.05);
    \end{scope}
    
    \begin{scope}[rotate=-90, shift=(n6)]
        \coordinate (f18) at (1.2,-0.3);
        \coordinate (f19) at (1.1,0.707);
        \coordinate (f20) at (1.3,-0.9);
    
        \path[thick, fill=blue!70, fill opacity=0.2] (f18) circle (0.707);
        \path[thick, fill=blue!70, fill opacity=0.2] (f19) circle (0.707);
        \path[thick, fill=blue!70, fill opacity=0.2] (f20) circle (0.707);
    
        \path[draw, fill, blue!70] (f18) circle (0.05);
        \path[draw, fill, blue!70] (f19) circle (0.05);
        \path[draw, fill, blue!70] (f20) circle (0.05);
    \end{scope}

    \foreach \x in {4.15, 3.85, ..., -2.6}  
        \foreach \y in {5.15, 4.85, ..., -2.0} 
            \path[fill, fill opacity=0.4] (\x,\y) circle (0.02);

    \begin{scope}
        \clip (f1) circle (0.707) 
              (f2) circle (0.707) 
              (f3) circle (0.707) 
              (f4) circle (0.707) 
              (f5) circle (0.707) 
              (f6) circle (0.707) 
              (f7) circle (0.707) 
              (f8) circle (0.707) 
              (f9) circle (0.707) 
              (f10) circle (0.707) 
              (f11) circle (0.707) 
              (f12) circle (0.707) 
              (f13) circle (0.707) 
              (f14) circle (0.707) 
              (f15) circle (0.707) 
              (f16) circle (0.707) 
              (f17) circle (0.707) 
              (f18) circle (0.707) 
              (f19) circle (0.707) 
              (f20) circle (0.707);

        \foreach \x in {4.15, 3.85, ..., -2.6}  
            \foreach \y in {5.15, 4.85, ..., -2.0} 
                \path[fill, orange] (\x,\y) circle (0.04);
    \end{scope}

\end{scope}

\coordinate (nvs) at ($(novel) + (1.35,-1.3)$);
\begin{scope}[rotate=90, scale=0.2, scale line widths=0.2, shift=(nvs)]
    \coordinate (n0) at (0.0,0.0);
    \coordinate (n1) at (-0.3,1.0);
    \coordinate (n2) at (0.3,2.2);
    \coordinate (n3) at (1.2,3.0);
    \coordinate (n4) at (1.9,2.4);
    \coordinate (n5) at (1.8,1.2);
    \coordinate (n6) at (1,0.3);

    \begin{scope}[rotate=190, shift=(n0)]
        \coordinate (f1) at (1.2,-0.3);
        \coordinate (f2) at (1.3,-0.8);
        \coordinate (f3) at (1.2,0.707);
    
        \path[thick, fill=blue!70, fill opacity=0.2] (f1) circle (0.707);
        \path[thick, fill=blue!70, fill opacity=0.2] (f2) circle (0.707);
        \path[thick, fill=blue!70, fill opacity=0.2] (f3) circle (0.707);
    
        \path[draw, fill, blue!70] (f1) circle (0.05);
        \path[draw, fill, blue!70] (f2) circle (0.05);
        \path[draw, fill, blue!70] (f3) circle (0.05);
    \end{scope}
    
    \begin{scope}[rotate=140, shift=(n1)]
        \coordinate (f4) at (1.2,-0.3);
        \coordinate (f5) at (1.2,0.707);
    
        \path[thick, fill=blue!70, fill opacity=0.2] (f4) circle (0.707);
        \path[thick, fill=blue!70, fill opacity=0.2] (f5) circle (0.707);
    
        \path[draw, fill, blue!70] (f4) circle (0.05);
        \path[draw, fill, blue!70] (f5) circle (0.05);
    \end{scope}
    
    \begin{scope}[rotate=140, shift=(n2)]
        \coordinate (f6) at (1,0.0);
        \coordinate (f7) at (1.3,-0.8);
        \coordinate (f8) at (1.2,0.707);
    
        \path[thick, fill=blue!70, fill opacity=0.2] (f6) circle (0.707);
        \path[thick, fill=blue!70, fill opacity=0.2] (f7) circle (0.707);
        \path[thick, fill=blue!70, fill opacity=0.2] (f8) circle (0.707);
    
        \path[draw, fill, blue!70] (f6) circle (0.05);
        \path[draw, fill, blue!70] (f7) circle (0.05);
        \path[draw, fill, blue!70] (f8) circle (0.05);
    \end{scope}
    
    \begin{scope}[rotate=100, shift=(n3)]
        \coordinate (f9) at (1.1,0.1);
        \coordinate (f10) at (1.3,-0.8);
        \coordinate (f11) at (1.2,0.707);
    
        \path[thick, fill=blue!70, fill opacity=0.2] (f9) circle (0.707);
        \path[thick, fill=blue!70, fill opacity=0.2] (f10) circle (0.707);
        \path[thick, fill=blue!70, fill opacity=0.2] (f11) circle (0.707);
    
        \path[draw, fill, blue!70] (f9) circle (0.05);
        \path[draw, fill, blue!70] (f10) circle (0.05);
        \path[draw, fill, blue!70] (f11) circle (0.05);
    \end{scope}
    
    \begin{scope}[rotate=25, shift=(n4)]
        \coordinate (f12) at (1.2,-0.3);
        \coordinate (f13) at (1.3,-0.8);
        \coordinate (f14) at (1.2,0.707);
    
        \path[thick, fill=blue!70, fill opacity=0.2] (f12) circle (0.707);
        \path[thick, fill=blue!70, fill opacity=0.2] (f13) circle (0.707);
        \path[thick, fill=blue!70, fill opacity=0.2] (f14) circle (0.707);
    
        \path[draw, fill, blue!70] (f12) circle (0.05);
        \path[draw, fill, blue!70] (f13) circle (0.05);
        \path[draw, fill, blue!70] (f14) circle (0.05);
    \end{scope}
    
    \begin{scope}[rotate=-20, shift=(n5)]
        \coordinate (f15) at (1.1,-0.4);
        \coordinate (f16) at (1.2,0.707);
        \coordinate (f17) at (1.3,-0.9);
    
        \path[thick, fill=blue!70, fill opacity=0.2] (f15) circle (0.707);
        \path[thick, fill=blue!70, fill opacity=0.2] (f16) circle (0.707);
        \path[thick, fill=blue!70, fill opacity=0.2] (f17) circle (0.707);
    
        \path[draw, fill, blue!70] (f15) circle (0.05);
        \path[draw, fill, blue!70] (f16) circle (0.05);
        \path[draw, fill, blue!70] (f17) circle (0.05);
    \end{scope}
    
    \begin{scope}[rotate=-90, shift=(n6)]
        \coordinate (f18) at (1.2,-0.3);
        \coordinate (f19) at (1.1,0.707);
        \coordinate (f20) at (1.3,-0.9);
    
        \path[thick, fill=blue!70, fill opacity=0.2] (f18) circle (0.707);
        \path[thick, fill=blue!70, fill opacity=0.2] (f19) circle (0.707);
        \path[thick, fill=blue!70, fill opacity=0.2] (f20) circle (0.707);
    
        \path[draw, fill, blue!70] (f18) circle (0.05);
        \path[draw, fill, blue!70] (f19) circle (0.05);
        \path[draw, fill, blue!70] (f20) circle (0.05);
    \end{scope}

    \coordinate (cam) at (1.2,0.8);
    \begin{scope}[rotate=-20, shift=(cam), scale=1.0]
        \path[draw, fill, thick] (0,0) circle (0.05);
        \draw[thick] (0:0) -- (-30:0.3) -- (30:0.3) -- cycle;
        \foreach \d [count=\di] in {0,0.2,...,3.5}
            \foreach \a [count=\da] in {-30,-25, ..., 30}
            {
                \pgfmathparse{\d + 0.2 * random()}
                \coordinate (p_\di_\da) at (\a:\pgfmathresult);
                \path[fill, fill opacity=0.4] (p_\di_\da) circle (0.02);
            }

        \clip (f1) circle (0.707) 
              (f2) circle (0.707) 
              (f3) circle (0.707) 
              (f4) circle (0.707) 
              (f5) circle (0.707) 
              (f6) circle (0.707) 
              (f7) circle (0.707) 
              (f8) circle (0.707) 
              (f9) circle (0.707) 
              (f10) circle (0.707) 
              (f11) circle (0.707) 
              (f12) circle (0.707) 
              (f13) circle (0.707) 
              (f14) circle (0.707) 
              (f15) circle (0.707) 
              (f16) circle (0.707) 
              (f17) circle (0.707) 
              (f18) circle (0.707) 
              (f19) circle (0.707) 
              (f20) circle (0.707);
        \foreach \d [count=\di] in {0,0.2,...,3.5}
            \foreach \a [count=\da] in {-30,-25, ..., 30}
                \path[fill, orange] (p_\di_\da) circle (0.04);
    \end{scope}

\end{scope}

\coordinate (opt1) at ($(optbb.north west) + (0.15, -0.15)$);
\begin{scope}[shift=(opt1), scale=0.8, scale line widths=0.8, local bounding box=opt1bb]
    \coordinate (kf1) at (0.3, -0.7);
    \coordinate (kf2) at (0.3, -1.5);
    \coordinate (kf3) at (0.3, -2.0);
    \coordinate (wallcorner) at (1.8,-3.2);
    \coordinate (field) at (1.25, -1.75);

    \path[fill=white,rounded corners] (0.0, 0.0) rectangle ++ (2.5,-3.5);
    \path[name path=rayendline] (2.2, 0.0) -- (2.2, -3.5);
    
    \path[name path=wall] (1.8,-0.3) -- (wallcorner) -- (0.3, -3.2);
    
    \begin{scope}[rotate=-20, shift=(kf1), scale=1.0]
        \path[draw, fill, thick] (0,0) circle (0.05);
        \draw[thick] (0:0) -- (-30:0.3) -- (30:0.3) -- cycle;
        \begin{scope}[overlay]
            \path[name path global=kf1-right] (0:0) -- (-30:5.0);
            \path[name path global=kf1-left] (0:0) -- (30:5.0);
        \end{scope}

        \begin{scope}[rotate=-10, overlay]
            \coordinate (ray) at (0:3.0);
            \path[name path=ray] (0:0) -- (ray);
            \path[name intersections={of=rayendline and ray, by=rayend}];
            \coordinate (rayop) at ($(kf1)!(field)!(rayend)$);
            \coordinate (segmentstart) at ($(rayop) + (0:0.6363)$);
            \coordinate (segmentend) at ($(rayop) - (0:0.6363)$);
            \draw (segmentstart) -- (segmentend);
            \draw ($(segmentstart) + (0.0, 0.05)$) -- ++ (0.0, -0.1);
            \draw ($(segmentend) + (0.0, 0.05)$) -- ++ (0.0, -0.1);
            \draw[opacity=0.5, ->] (0:0) -- (rayend);
        \end{scope}
        
        \path [name intersections={of=kf1-right and wall, by=kf1wall1}];
        \path [name intersections={of=kf1-left and wall, by=kf1wall2}];
        \draw[thick, line cap=round, dash pattern=on 0pt off 1.5pt, dash phase=1.8pt] (kf1wall1) -- (kf1wall2);
        \path[fill=gray, fill opacity=0.1] (kf1) -- (kf1wall1) -- (kf1wall2) -- cycle;
    \end{scope}
    \begin{scope}[rotate=0, shift=(kf2), scale=1.0]
        \path[draw, fill, thick] (0,0) circle (0.05);
        \draw[thick] (0:0) -- (-30:0.3) -- (30:0.3) -- cycle;
        \begin{scope}[overlay]
            \path[name path global=kf2-right] (0:0) -- (-30:5.0);
            \path[name path global=kf2-left] (0:0) -- (30:5.0);
        \end{scope}

        \begin{scope}[rotate=10, overlay]
            \coordinate (ray) at (0:3.0);
            \path[name path=ray] (0:0) -- (ray);
            \path[name intersections={of=rayendline and ray, by=rayend}];
            \coordinate (rayop) at ($(kf2)!(field)!(rayend)$);
            \coordinate (segmentstart) at ($(rayop) + (0:0.6363)$);
            \coordinate (segmentend) at ($(rayop) - (0:0.6363)$);
            \draw (segmentstart) -- (segmentend);
            \draw ($(segmentstart) + (0.0, 0.05)$) -- ++ (0.0, -0.1);
            \draw ($(segmentend) + (0.0, 0.05)$) -- ++ (0.0, -0.1);
            \draw[opacity=0.5, ->] (0:0) -- (rayend);
        \end{scope}

        \begin{scope}[rotate=-10, overlay]
            \coordinate (ray) at (0:3.0);
            \path[name path=ray] (0:0) -- (ray);
            \path[name intersections={of=rayendline and ray, by=rayend}];
            \coordinate (rayop) at ($(kf2)!(field)!(rayend)$);
            \coordinate (segmentstart) at ($(rayop) + (0:0.6363)$);
            \coordinate (segmentend) at ($(rayop) - (0:0.6363)$);
            \draw (segmentstart) -- (segmentend);
            \draw ($(segmentstart) + (0.0, 0.05)$) -- ++ (0.0, -0.1);
            \draw ($(segmentend) + (0.0, 0.05)$) -- ++ (0.0, -0.1);
            \draw[opacity=0.5, ->] (0:0) -- (rayend);
        \end{scope}
    
        \path [name intersections={of=kf2-right and wall, by=kf2wall1}];
        \path [name intersections={of=kf2-left and wall, by=kf2wall2}];
        \draw[thick, line cap=round, dash pattern=on 0pt off 1.5pt] (kf2wall1) -- (kf2wall2);
        \path[fill=gray, fill opacity=0.1] (kf2) -- (kf2wall1) -- (kf2wall2) -- cycle;
    \end{scope}
    \begin{scope}[rotate=-20, shift=(kf3), scale=1.0]
        \path[draw, fill, thick] (0,0) circle (0.05);
        \draw[thick] (0:0) -- (-30:0.3) -- (30:0.3) -- cycle;
        \begin{scope}[overlay]
            \path[name path global=kf3-right] (0:0) -- (-30:5.0);
            \path[name path global=kf3-left] (0:0) -- (30:5.0);
        \end{scope}

        \begin{scope}[rotate=23, overlay]
            \coordinate (ray) at (0:3.0);
            \path[name path=ray] (0:0) -- (ray);
            \path[name intersections={of=rayendline and ray, by=rayend}];
            \coordinate (rayop) at ($(kf3)!(field)!(rayend)$);
            \coordinate (segmentstart) at ($(rayop) + (0:0.6363)$);
            \coordinate (segmentend) at ($(rayop) - (0:0.6363)$);
            \draw (segmentstart) -- (segmentend);
            \draw ($(segmentstart) + (0.0, 0.05)$) -- ++ (0.0, -0.1);
            \draw ($(segmentend) + (0.0, 0.05)$) -- ++ (0.0, -0.1);
            \draw[opacity=0.5, ->] (0:0) -- (rayend);
        \end{scope}
        
        \path [name intersections={of=kf3-right and wall, by=kf3wall1}];
        \path [name intersections={of=kf3-left and wall, by=kf3wall2}];
        \draw[thick, line cap=round, dash pattern=on 0pt off 1.5pt] (kf3wall1) -- (wallcorner) -- (kf3wall2);
        \path[fill=gray, fill opacity=0.1] (kf3) -- (kf3wall1) -- (wallcorner) -- (kf3wall2) -- cycle;
    \end{scope}

    \begin{scope}[rotate=0, shift=(field), scale=0.9]
        \path[fill=blue!70] (0,0) circle (0.05);
        \path[thick, fill=blue!70, fill opacity=0.2] (0,0) circle (0.707);
    \end{scope}
\end{scope}

\coordinate (opt2) at ($(opt1bb.north east) + (0.15, 0.0)$);
\begin{scope}[shift=(opt2), scale=0.8, scale line widths=0.8, local bounding box=opt2bb]
    \coordinate (kf1) at (0.5, -0.5);
    \coordinate (kf2) at (1.7, -0.8);
    \coordinate (kf3) at (1.5, -3.1);
    \coordinate (kf4) at (0.7, -2.5);
    \coordinate (field) at (1.25, -1.75);

    \path[fill=white,rounded corners] (0.0, 0.0) rectangle ++ (2.5,-3.5);
    
    \path[name path=wall] (0.0,-1.55) -- (2.5, -1.25);
    \path[name path=wall2] (0.0,-1.75) -- (2.5, -1.45);
    
    \begin{scope}[rotate=-60, shift=(kf1), scale=1.0]
        \path[draw, fill, thick] (0,0) circle (0.05);
        \draw[thick] (0:0) -- (-30:0.3) -- (30:0.3) -- cycle;
        \begin{scope}[overlay]
            \path[name path global=kf-right] (0:0) -- (-30:5.0);
            \path[name path global=kf-left] (0:0) -- (30:5.0);
        \end{scope}

        \begin{scope}[rotate=-10, overlay]
            \coordinate (rayend) at (0:2.3);
            \coordinate (rayop) at ($(kf1)!(field)!(rayend)$);
            \coordinate (segmentstart) at ($(rayop) + (0:0.6363)$);
            \coordinate (segmentend) at ($(rayop) - (0:0.6363)$);
            \draw (segmentstart) -- (segmentend);
            \draw ($(segmentstart) + (0.0, 0.05)$) -- ++ (0.0, -0.1);
            \draw ($(segmentend) + (0.0, 0.05)$) -- ++ (0.0, -0.1);
            \draw[opacity=0.5, ->] (0:0) -- (rayend);
        \end{scope}
        
        \path [name intersections={of=kf-right and wall, by=kfwall1}];
        \path [name intersections={of=kf-left and wall, by=kfwall2}];
        \draw[thick, line cap=round, dash pattern=on 0pt off 1.5pt, dash phase=1.8pt] (kfwall1) -- (kfwall2);
        \path[fill=gray, fill opacity=0.1] (kf1) -- (kfwall1) -- (kfwall2) -- cycle;
    \end{scope}
    
    \begin{scope}[rotate=-110, shift=(kf2), scale=1.0]
        \path[draw, fill, thick] (0,0) circle (0.05);
        \draw[thick] (0:0) -- (-30:0.3) -- (30:0.3) -- cycle;
        \begin{scope}[overlay]
            \path[name path global=kf-right] (0:0) -- (-30:5.0);
            \path[name path global=kf-left] (0:0) -- (30:5.0);
        \end{scope}

        \begin{scope}[rotate=-20, overlay]
            \coordinate (rayend) at (0:2.0);
            \coordinate (rayop) at ($(kf2)!(field)!(rayend)$);
            \coordinate (segmentstart) at ($(rayop) + (0:0.6363)$);
            \coordinate (segmentend) at ($(rayop) - (0:0.6363)$);
            \draw (segmentstart) -- (segmentend);
            \draw ($(segmentstart) + (0.0, 0.05)$) -- ++ (0.0, -0.1);
            \draw ($(segmentend) + (0.0, 0.05)$) -- ++ (0.0, -0.1);
            \draw[opacity=0.5, ->] (0:0) -- (rayend);
        \end{scope}
        
        \path [name intersections={of=kf-right and wall, by=kfwall1}];
        \path [name intersections={of=kf-left and wall, by=kfwall2}];
        \draw[thick, line cap=round, dash pattern=on 0pt off 1.5pt, dash phase=1.8pt] (kfwall1) -- (kfwall2);
        \path[fill=gray, fill opacity=0.1] (kf2) -- (kfwall1) -- (kfwall2) -- cycle;
    \end{scope}
    
    \begin{scope}[rotate=95, shift=(kf3), scale=1.0]
        \path[draw, fill, thick] (0,0) circle (0.05);
        \draw[thick] (0:0) -- (-30:0.3) -- (30:0.3) -- cycle;
        \begin{scope}[overlay]
            \path[name path global=kf-right] (0:0) -- (-30:5.0);
            \path[name path global=kf-left] (0:0) -- (30:5.0);
        \end{scope}

        \begin{scope}[rotate=4, overlay]
            \coordinate (rayend) at (0:2.3);
            \coordinate (rayop) at ($(kf3)!(field)!(rayend)$);
            \coordinate (segmentstart) at ($(rayop) + (0:0.6363)$);
            \coordinate (segmentend) at ($(rayop) - (0:0.6363)$);
            \draw (segmentstart) -- (segmentend);
            \draw ($(segmentstart) + (0.0, 0.05)$) -- ++ (0.0, -0.1);
            \draw ($(segmentend) + (0.0, 0.05)$) -- ++ (0.0, -0.1);
            \draw[opacity=0.5, ->] (0:0) -- (rayend);
        \end{scope}
        
        \path [name intersections={of=kf-right and wall2, by=kfwall1}];
        \path [name intersections={of=kf-left and wall2, by=kfwall2}];
        \draw[thick, line cap=round, dash pattern=on 0pt off 1.5pt, dash phase=1.8pt] (kfwall1) -- (kfwall2);
        \path[fill=gray, fill opacity=0.1] (kf3) -- (kfwall1) -- (kfwall2) -- cycle;
    \end{scope}
    
    \begin{scope}[rotate=80, shift=(kf4), scale=1.0]
        \path[draw, fill, thick] (0,0) circle (0.05);
        \draw[thick] (0:0) -- (-30:0.3) -- (30:0.3) -- cycle;
        \begin{scope}[overlay]
            \path[name path global=kf-right] (0:0) -- (-30:5.0);
            \path[name path global=kf-left] (0:0) -- (30:5.0);
        \end{scope}

        \begin{scope}[rotate=-1, overlay]
            \coordinate (rayend) at (0:1.8);
            \coordinate (rayop) at ($(kf4)!(field)!(rayend)$);
            \coordinate (segmentstart) at ($(rayop) + (0:0.6363)$);
            \coordinate (segmentend) at ($(rayop) - (0:0.6363)$);
            \draw (segmentstart) -- (segmentend);
            \draw ($(segmentstart) + (0.0, 0.05)$) -- ++ (0.0, -0.1);
            \draw ($(segmentend) + (0.0, 0.05)$) -- ++ (0.0, -0.1);
            \draw[opacity=0.5, ->] (0:0) -- (rayend);
        \end{scope}
        
        \path [name intersections={of=kf-right and wall2, by=kfwall1}];
        \path [name intersections={of=kf-left and wall2, by=kfwall2}];
        \draw[thick, line cap=round, dash pattern=on 0pt off 1.5pt, dash phase=1.8pt] (kfwall1) -- (kfwall2);
        \path[fill=gray, fill opacity=0.1] (kf4) -- (kfwall1) -- (kfwall2) -- cycle;
    \end{scope}
    
    \begin{scope}[rotate=0, shift=(field), scale=0.9]
        \path[fill=blue!70] (0,0) circle (0.05);
        \path[thick, fill=blue!70, fill opacity=0.2] (0,0) circle (0.707);
    \end{scope}
\end{scope}

\coordinate (opt3) at ($(opt2bb.north east) + (0.15, 0.0)$);
\begin{scope}[shift=(opt3), scale=0.8, scale line widths=0.8, local bounding box=opt3bb]
    \path[fill=white,rounded corners] (0.0, 0.0) rectangle ++ (2.5,-3.5);
    \draw[fill] (1.25, -1.75) circle (0.05);
    \draw[fill] (1.05, -1.75) circle (0.05);
    \draw[fill] (1.45, -1.75) circle (0.05);
\end{scope}
\end{tikzpicture}%

%% file: figures/instantiation.tikz
\begin{tikzpicture}

\tikzset{
    invclip/.style={
        clip,
        insert path={{[reset cm](-16383.99999pt,-16383.99999pt) rectangle (16383.99999pt,16383.99999pt)}}
    }
}

\coordinate (grid) at (0.0,0.0);
\coordinate (camera) at (0.6,3.3);
\coordinate (field1) at (3.4,2.6);
\coordinate (field2) at (3.6,1.4);
\coordinate (field3) at (3.5,0.5);
\coordinate (field4) at (2.5,0.5);
\coordinate (field5) at (1.5,0.5);

\begin{scope}[rotate=0, shift=(grid)]
    \foreach \y in {0,1,2,3}
      \draw[gray, opacity=0.5] (-0.5,\y) -- (5.5,\y);
    \foreach \x in {0,1,2,3,4,5}
      \draw[gray, opacity=0.5] (\x,-0.5) -- (\x,3.5);
    \draw[<->] (5.2, 2.0) -- node[right] {$g=2r/\sqrt{3}$} (5.2, 3.0);
    \path[pattern={crosshatch}, opacity=0.5, pattern color=orange] (1,0) rectangle (2,1);
    \path[pattern={crosshatch}, opacity=0.5, pattern color=orange] (2,0) rectangle (3,1);
    \path[pattern={crosshatch}, opacity=0.5, pattern color=orange] (3,0) rectangle (4,1);
\end{scope}

\begin{scope}[rotate=0, shift=(field1), scale=1.0]
    \path[fill=blue!70] (0,0) circle (0.05);
    \path[thick, fill=blue!70, fill opacity=0.2] (0,0) circle (0.707);
    \path[draw,<->] (0:0) -- node[above] {$r$} (15:0.707);
\end{scope}

\begin{scope}[rotate=0, shift=(field2), scale=1.0]
    \path[fill=blue!70] (0,0) circle (0.05);
    \path[thick, fill=blue!70, fill opacity=0.2] (0,0) circle (0.707);
\end{scope}

\begin{scope}[rotate=0, shift=(field3), scale=1.0]
    \path[fill=orange] (0,0) circle (0.05);
    \path[thick, fill=orange, fill opacity=0.2] (0,0) circle (0.707);
\end{scope}

\begin{scope}[rotate=0, shift=(field4), scale=1.0]
    \path[fill=orange] (0,0) circle (0.05);
    \path[thick, fill=orange, fill opacity=0.2] (0,0) circle (0.707);
\end{scope}

\begin{scope}[rotate=0, shift=(field5), scale=1.0]
    \path[fill=orange] (0,0) circle (0.05);
    \path[thick, fill=orange, fill opacity=0.2] (0,0) circle (0.707);
\end{scope}

\begin{scope}[rotate=-40, shift=(camera), scale=2.0]
    \path[draw, fill, thick] (0,0) circle (0.05);
    \draw[thick] (0:0) -- (-30:0.3) -- (30:0.3) -- cycle;
    \draw[thick, line cap=round, dash pattern=on 0pt off 1.5pt] plot [smooth, tension=0.3] coordinates {(-30:1.5) (0:2.0) (30:1.5)};
\end{scope}

\begin{scope}[rotate=-40, shift=(camera), scale=2.0]
    \begin{pgfinterruptboundingbox}
        \path [invclip] (field1) circle (0.707);
        \path [invclip] (field2) circle  (0.707);
    \end{pgfinterruptboundingbox}
    \draw[very thick, red!80!black, line cap=round, dash pattern=on 0pt off 1.5pt] plot [smooth, tension=0.3] coordinates {(-30:1.5) (0:2.0) (30:1.5)};
\end{scope}

\begin{scope}[rotate=0, shift=(grid)]
    \node at (3.9, 1.8) [black] {$\mathcal{X}$};
    \node at (2.5, 0.8) [red!80!black] {$\mathcal{X}_\mathrm{unc}$};
    \node at (2.8, 1.8) [blue] {$\mathcal{F}$};
    \node at (2.5, -0.4) [orange] {$\mathcal{F}_\mathrm{new}$};
\end{scope}

\end{tikzpicture}

%% file: figures/sampling.tikz
\scriptsize%
\begin{tikzpicture}
\tikzset{
    invclip/.style={
        clip,
        insert path={{[reset cm](-16383.99999pt,-16383.99999pt) rectangle (16383.99999pt,16383.99999pt)}}
    }
}

\coordinate (step1) at (0.0,4.0);
\coordinate (step2) at (4.0,4.0);
\coordinate (step3) at (0.0,1.0);

\begin{scope}[shift=(step1)]
    \coordinate (camera) at (2.5,-1.7);
    \coordinate (field1) at (3.0,-2.5);
    \coordinate (field2) at (3.4,-2.5);
    \coordinate (field3) at (3.4,-2.1);
    \coordinate (field4) at (3.4,-1.6);
    \coordinate (field5) at (3.3,-1.2);
    \coordinate (field6) at (3.3,-0.8);
    \coordinate (field7) at (2.9,-1.0);
    \coordinate (field8) at (2.5,-1.0);
    \coordinate (field9) at (2.7,-0.7);
    \coordinate (field10) at (2.3,-0.8);
    \coordinate (field11) at (1.9,-0.7);
    \coordinate (field12) at (1.5,-0.9);
    \coordinate (field13) at (1.3,-0.6);
    \coordinate (field14) at (1.0,-0.8);
    \coordinate (field15) at (1.0,-1.2);
    \node[anchor=north west] at (0.0, 0.0) [black] {(1)};
    
    \path [general shadow/.default={draw=orange, line width=2pt, fill=none, shadow xshift=0pt, shadow yshift=0pt},
        my shadow/.append style={fill=white, general shadow},my shadow]
          (field1) circle (0.2828)
          (field2) circle (0.2828)
          (field3) circle (0.2828)
          (field4) circle (0.2828);
          
    \begin{scope}[rotate=0, shift=(field1), scale=0.4]
        \path[fill=orange] (0,0) circle (0.05);
        \path[thick,fill=orange, fill opacity=0.2] (0,0) circle (0.707);
    \end{scope}
    \begin{scope}[rotate=0, shift=(field2), scale=0.4]
        \path[fill=blue!70] (0,0) circle (0.05);
        \path[thick, fill=blue!70, fill opacity=0.2] (0,0) circle (0.707);
    \end{scope}
    \begin{scope}[rotate=0, shift=(field3), scale=0.4]
        \path[fill=orange] (0,0) circle (0.05);
        \path[thick, fill=orange, fill opacity=0.2] (0,0) circle (0.707);
    \end{scope}
    \begin{scope}[rotate=0, shift=(field4), scale=0.4]
        \path[fill=orange] (0,0) circle (0.05);
        \path[thick, fill=orange, fill opacity=0.2] (0,0) circle (0.707);
    \end{scope}
    \begin{scope}[rotate=0, shift=(field5), scale=0.4]
        \path[fill=blue!70] (0,0) circle (0.05);
        \path[thick, fill=blue!70, fill opacity=0.2] (0,0) circle (0.707);
    \end{scope}
    \begin{scope}[rotate=0, shift=(field6), scale=0.4]
        \path[fill=blue!70] (0,0) circle (0.05);
        \path[thick, fill=blue!70, fill opacity=0.2] (0,0) circle (0.707);
    \end{scope}
    \begin{scope}[rotate=0, shift=(field7), scale=0.4]
        \path[fill=orange] (0,0) circle (0.05);
        \path[thick, fill=orange, fill opacity=0.2] (0,0) circle (0.707);
    \end{scope}
    \begin{scope}[rotate=0, shift=(field8), scale=0.4]
        \path[fill=blue!70] (0,0) circle (0.05);
        \path[thick, fill=blue!70, fill opacity=0.2] (0,0) circle (0.707);
    \end{scope}
    \begin{scope}[rotate=0, shift=(field9), scale=0.4]
        \path[fill=blue!70] (0,0) circle (0.05);
        \path[thick, fill=blue!70, fill opacity=0.2] (0,0) circle (0.707);
    \end{scope}
    \begin{scope}[rotate=0, shift=(field10), scale=0.4]
        \path[fill=blue!70] (0,0) circle (0.05);
        \path[thick, fill=blue!70, fill opacity=0.2] (0,0) circle (0.707);
    \end{scope}
    \begin{scope}[rotate=0, shift=(field11), scale=0.4]
        \path[fill=blue!70] (0,0) circle (0.05);
        \path[thick, fill=blue!70, fill opacity=0.2] (0,0) circle (0.707);
    \end{scope}
    \begin{scope}[rotate=0, shift=(field12), scale=0.4]
        \path[fill=orange] (0,0) circle (0.05);
        \path[thick, fill=orange, fill opacity=0.2] (0,0) circle (0.707);
    \end{scope}
    \begin{scope}[rotate=0, shift=(field13), scale=0.4]
        \path[fill=blue!70] (0,0) circle (0.05);
        \path[thick, fill=blue!70, fill opacity=0.2] (0,0) circle (0.707);
    \end{scope}
    \begin{scope}[rotate=0, shift=(field14), scale=0.4]
        \path[fill=blue!70] (0,0) circle (0.05);
        \path[thick, fill=blue!70, fill opacity=0.2] (0,0) circle (0.707);
    \end{scope}
    \begin{scope}[rotate=0, shift=(field15), scale=0.4]
        \path[fill=orange] (0,0) circle (0.05);
        \path[thick, fill=orange, fill opacity=0.2] (0,0) circle (0.707);
    \end{scope}
    
    \begin{scope}[rotate=-30, shift=(camera), scale=0.6]
        \path[draw, fill, thick] (0,0) circle (0.05);
        \draw[thick] (0:0) -- (-30:0.3) -- (30:0.3) -- cycle;
        \draw[thick, line cap=round, dash pattern=on 0pt off 1.5pt] plot [smooth, tension=0.3] coordinates {(-30:1.5) (0:2.0) (30:1.5)};
    \end{scope}
\end{scope}
\begin{scope}[shift=(step2)]
    \coordinate (field) at (3, -1.5);

    \coordinate (kf1) at (2.0, -0.7);
    \coordinate (kf2) at (2.0, -1.5);
    \coordinate (kf3) at (1.7, -2.0);
    \coordinate (kf4) at (1.9, -2.5);
    \coordinate (kf5) at (0.7, -0.8);
    \coordinate (kf6) at (0.8, -1.5);
    \coordinate (wallcorner) at (3.5,-3.2);
    
    \path[name path=wall] (3.5,-0.3) -- (wallcorner) -- (1.0, -3.2);
    \path[name path=wall2] (1.5,-0.3) -- (1.5, -3.2);
    
    \node[anchor=north west] at (0.0, 0.0) [black] {(2)};
    
    \begin{scope}[rotate=-20, shift=(kf1), scale=1.0]
        \path[draw, fill, thick] (0,0) circle (0.05);
        \draw[thick] (0:0) -- (-30:0.3) -- (30:0.3) -- cycle;
        \begin{scope}[overlay]
            \path[name path global=kf1-right] (0:0) -- (-30:5.0);
            \path[name path global=kf1-left] (0:0) -- (30:5.0);
        \end{scope}

        \begin{scope}[rotate=-10]
            \coordinate (rayend) at (0:2.15);
            \coordinate (rayop) at ($(kf1)!(field)!(rayend)$);
            \coordinate (segmentstart) at ($(rayop) + (0:0.6363)$);
            \coordinate (segmentend) at ($(rayop) - (0:0.6363)$);
            \draw (segmentstart) -- (segmentend);
            \draw ($(segmentstart) + (0.0, 0.05)$) -- ++ (0.0, -0.1);
            \draw ($(segmentend) + (0.0, 0.05)$) -- ++ (0.0, -0.1);
            \draw[opacity=0.5, ->] (0:0) -- (rayend);
        \end{scope}
        
        \path [name intersections={of=kf1-right and wall, by=kf1wall1}];
        \path [name intersections={of=kf1-left and wall, by=kf1wall2}];
        \draw[thick, line cap=round, dash pattern=on 0pt off 1.5pt, dash phase=1.8pt] (kf1wall1) -- (kf1wall2);
        \path[fill=gray, fill opacity=0.1] (kf1) -- (kf1wall1) -- (kf1wall2) -- cycle;
    \end{scope}
    \begin{scope}[rotate=0, shift=(kf2), scale=1.0]
        \path[draw, fill, thick] (0,0) circle (0.05);
        \draw[thick] (0:0) -- (-30:0.3) -- (30:0.3) -- cycle;
        \begin{scope}[overlay]
            \path[name path global=kf2-right] (0:0) -- (-30:5.0);
            \path[name path global=kf2-left] (0:0) -- (30:5.0);
        \end{scope}

        \begin{scope}[rotate=10]
            \coordinate (rayend) at (0:1.89);
            \coordinate (rayop) at ($(kf2)!(field)!(rayend)$);
            \coordinate (segmentstart) at ($(rayop) + (0:0.6363)$);
            \coordinate (segmentend) at ($(rayop) - (0:0.6363)$);
            \draw (segmentstart) -- (segmentend);
            \draw ($(segmentstart) + (0.0, 0.05)$) -- ++ (0.0, -0.1);
            \draw ($(segmentend) + (0.0, 0.05)$) -- ++ (0.0, -0.1);
            \draw[opacity=0.5, ->] (0:0) -- (rayend);
        \end{scope}

        \begin{scope}[rotate=-10]
            \coordinate (rayend) at (0:1.89);
            \coordinate (rayop) at ($(kf2)!(field)!(rayend)$);
            \coordinate (segmentstart) at ($(rayop) + (0:0.6363)$);
            \coordinate (segmentend) at ($(rayop) - (0:0.6363)$);
            \draw (segmentstart) -- (segmentend);
            \draw ($(segmentstart) + (0.0, 0.05)$) -- ++ (0.0, -0.1);
            \draw ($(segmentend) + (0.0, 0.05)$) -- ++ (0.0, -0.1);
            \draw[opacity=0.5, ->] (0:0) -- (rayend);
        \end{scope}
    
        \path [name intersections={of=kf2-right and wall, by=kf2wall1}];
        \path [name intersections={of=kf2-left and wall, by=kf2wall2}];
        \draw[thick, line cap=round, dash pattern=on 0pt off 1.5pt] (kf2wall1) -- (kf2wall2);
        \path[fill=gray, fill opacity=0.1] (kf2) -- (kf2wall1) -- (kf2wall2) -- cycle;
    \end{scope}
    \begin{scope}[rotate=-30, shift=(kf3), scale=1.0]
        \path[draw, fill, thick] (0,0) circle (0.05);
        \draw[thick] (0:0) -- (-30:0.3) -- (30:0.3) -- cycle;
        \begin{scope}[overlay]
            \path[name path global=kf3-right] (0:0) -- (-30:5.0);
            \path[name path global=kf3-left] (0:0) -- (30:5.0);
        \end{scope}

        \begin{scope}[rotate=27]
            \coordinate (rayend) at (0:2.15);
            \coordinate (rayop) at ($(kf3)!(field)!(rayend)$);
            \coordinate (segmentstart) at ($(rayop) + (0:0.6363)$);
            \coordinate (segmentend) at ($(rayop) - (0:0.6363)$);
            \draw (segmentstart) -- (segmentend);
            \draw ($(segmentstart) + (0.0, 0.05)$) -- ++ (0.0, -0.1);
            \draw ($(segmentend) + (0.0, 0.05)$) -- ++ (0.0, -0.1);
            \draw[opacity=0.5, ->] (0:0) -- (rayend);
        \end{scope}
        
        \path [name intersections={of=kf3-right and wall, by=kf3wall1}];
        \path [name intersections={of=kf3-left and wall, by=kf3wall2}];
        \draw[thick, line cap=round, dash pattern=on 0pt off 1.5pt] (kf3wall1) -- (wallcorner) -- (kf3wall2);
        \path[fill=gray, fill opacity=0.1] (kf3) -- (kf3wall1) -- (wallcorner) -- (kf3wall2) -- cycle;
    \end{scope}
    \begin{scope}[rotate=-50, shift=(kf4), scale=1.0, transparency group, opacity=0.1]
        \path[draw, fill, thick] (0,0) circle (0.05);
        \draw[thick] (0:0) -- (-30:0.3) -- (30:0.3) -- cycle;
        \begin{scope}[overlay]
            \path[name path global=kf4-right] (0:0) -- (-30:5.0);
            \path[name path global=kf4-left] (0:0) -- (30:5.0);
        \end{scope}
    
        \path [name intersections={of=kf4-right and wall, by=kf4wall1}];
        \path [name intersections={of=kf4-left and wall, by=kf4wall2}];
        \draw[thick, line cap=round, dash pattern=on 0pt off 1.5pt] (kf4wall1) -- (wallcorner) -- (kf4wall2);
        \path[fill=gray, fill opacity=0.1] (kf4) -- (kf4wall1) -- (wallcorner) -- (kf4wall2) -- cycle;
    \end{scope}
    \begin{scope}[rotate=-10, shift=(kf5), scale=1.0, transparency group, opacity=0.1]
        \path[draw, fill, thick] (0,0) circle (0.05);
        \draw[thick] (0:0) -- (-30:0.3) -- (30:0.3) -- cycle;
        \begin{scope}[overlay]
            \path[name path global=kf5-right] (0:0) -- (-30:5.0);
            \path[name path global=kf5-left] (0:0) -- (30:5.0);
        \end{scope}
    
        \path [name intersections={of=kf5-right and wall2, by=kf5wall1}];
        \path [name intersections={of=kf5-left and wall2, by=kf5wall2}];
        \draw[thick, line cap=round, dash pattern=on 0pt off 1.5pt] (kf5wall1) -- (kf5wall2);
        \path[fill=gray, fill opacity=0.1] (kf5) -- (kf5wall1) -- (kf5wall2) -- cycle;
    \end{scope}
    \begin{scope}[rotate=-20, shift=(kf6), scale=1.0, transparency group, opacity=0.1]
        \path[draw, fill, thick] (0,0) circle (0.05);
        \draw[thick] (0:0) -- (-30:0.3) -- (30:0.3) -- cycle;
        \begin{scope}[overlay]
            \path[name path global=kf6-right] (0:0) -- (-30:5.0);
            \path[name path global=kf6-left] (0:0) -- (30:5.0);
        \end{scope}
    
        \path [name intersections={of=kf6-right and wall2, by=kf6wall1}];
        \path [name intersections={of=kf6-left and wall2, by=kf6wall2}];
        \draw[thick, line cap=round, dash pattern=on 0pt off 1.5pt] (kf6wall1) -- (kf6wall2);
        \path[fill=gray, fill opacity=0.1] (kf6) -- (kf6wall1) -- (kf6wall2) -- cycle;
    \end{scope}

    \begin{scope}[rotate=0, shift=(field), scale=0.9]
        \path[fill=orange] (0,0) circle (0.05);
        \path[thick, fill=orange, fill opacity=0.2] (0,0) circle (0.707);
        \path[fill=red!80!black] (0:0.707) circle (0.05);
        \path[fill=red!80!black] (50:0.707) circle (0.05);
        \path[fill=red!80!black] (120:0.707) circle (0.05);
        \path[fill=red!80!black] (170:0.707) circle (0.05);
        \path[fill=red!80!black] (210:0.707) circle (0.05);
        \path[fill=red!80!black] (270:0.707) circle (0.05);
        \path[fill=red!80!black] (300:0.707) circle (0.05);
    \end{scope}
    
\end{scope}

\begin{scope}[shift=(step3)]
    \node[anchor=north west] at (0.0, 0.0) [black] {(3)};
    \coordinate (camera) at (0.3,-1.36);
    \begin{scope}[rotate=0, shift=(camera), scale=1.0]
        \coordinate (obs) at (3.6,0.0);
        \path[draw, fill, thick] (0,0) circle(0.05)  node[xshift=0.37cm, yshift=0.4cm]{$\boldsymbol{o}+l\boldsymbol{d}$};
        \draw[thick] (0:0) -- (-30:0.3) -- (30:0.3) -- cycle;
        \draw[semithick] (0:0) -- (0:0.4); 
        \draw[semithick, dotted, line cap=round, dash pattern=on 0pt off 1.5pt] (0:0.4) -- (0:0.83);
        \draw[semithick, line cap=round, ->] (0:0.825) -- (0:8.0);
        \draw[thick] (1.0, 0.1) --++ (0.0, -0.2) node (lmin) [below] {$l_\mathrm{min}$};
        \draw[thick] (7.6, 0.1) --++ (0.0, -0.2) node (lmax) [below] {$l_\mathrm{max}$};
        \draw[thick,blue!70,decorate,decoration={brace,mirror,amplitude=4pt}] 
            (lmin.south) -- (lmax.south) node[midway,anchor=north,yshift=-0.5em,blue!70]{$N_\mathrm{up}$};
        
        \path[fill=blue!70] (1.7, 0.0) circle (0.07);
        \path[fill=blue!70] (2.7, 0.0) circle (0.07);
        \path[fill=blue!70] (4.0, 0.0) circle (0.07);
        \path[fill=blue!70] (5.2, 0.0) circle (0.07);
        \path[fill=blue!70] (6.6, 0.0) circle (0.07);
        
        \begin{scope}[rotate=0, shift=(obs), scale=1.0]
            \draw[thick] (0.0, 0.1) --++ (0.0, -0.2) node [below] {$l_\mathrm{obs}$};
            \draw[thick] (-0.6, -0.1) --++ (0.0, 0.5);
            \draw[thick] (0.6, -0.1) --++ (0.0, 0.5);
            \draw[semithick, <->] (-0.6, 0.25) -- node[above] {$2\tau$} (0.6, 0.25);
            \draw[thick,orange,decorate,decoration={brace,amplitude=4pt,raise=5ex}] 
                (-0.6,0) -- (0.6,0) node[midway,yshift=3.7em,orange]{$N_\mathrm{dp}$};

            \path[fill=orange] (-0.48, 0.0) circle (0.07);
            \path[fill=orange] (-0.29, 0.0) circle (0.07);
            \path[fill=orange] (-0.09, 0.0) circle (0.07);
            \path[fill=orange] (0.1, 0.0) circle (0.07);
            \path[fill=orange] (0.33, 0.0) circle (0.07);
            \path[fill=orange] (0.53, 0.0) circle (0.07);
        \end{scope}
    \end{scope}
\end{scope}

\end{tikzpicture}

%% file: figures/kintinuous.tikz
\begin{tikzpicture}
\normalsize

\node[inner sep=0pt, outer sep=0pt, anchor=north west] (ngmb_3855) at (0,0) {\includegraphics[width=2cm, trim={2mm 0 2mm 0},clip]{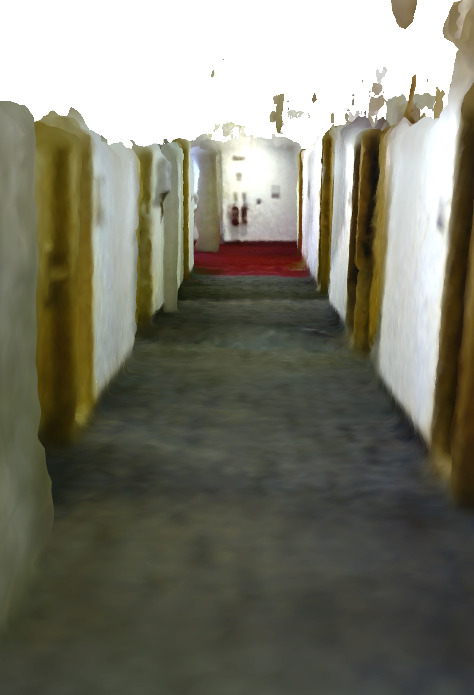}};
\node[inner sep=0pt, outer sep=0pt, anchor=south west] (ngmb_3856) at (ngmb_3855.south east) {\includegraphics[width=2cm, trim={2mm 0 2mm 0},clip]{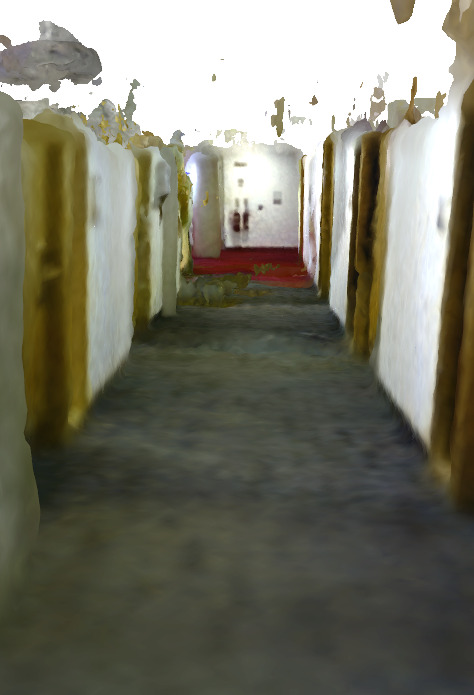}};

\node[inner sep=0pt, outer sep=0pt, anchor=south] (ngma_3856) at (ngmb_3855.north) {\includegraphics[width=2cm, trim={2mm 0 2mm 0},clip]{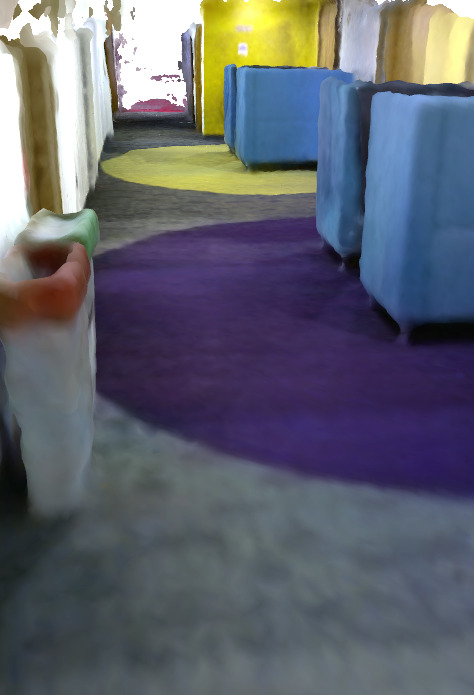}};
\node[inner sep=0pt, outer sep=0pt, anchor=south] (ngma_3855) at (ngma_3856.north) {\includegraphics[width=2cm, trim={2mm 0 2mm 0},clip]{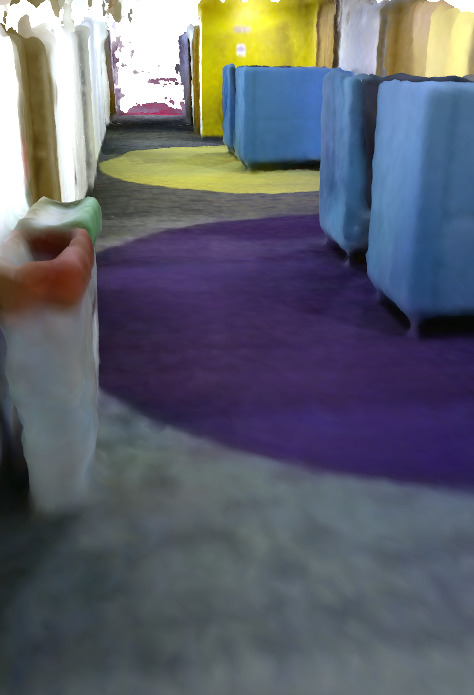}};

\path[preaction={clip,postaction={draw=cyan, line width=1.5mm}}] (ngma_3855.north west) rectangle (ngma_3855.south east);
\path[preaction={clip,postaction={draw=cyan, line width=1.5mm}}] (ngma_3856.north west) rectangle (ngma_3856.south east);
\path[preaction={clip,postaction={draw=orange, line width=1.5mm}}] (ngmb_3855.north west) rectangle (ngmb_3855.south east);
\path[preaction={clip,postaction={draw=orange, line width=1.5mm}}] (ngmb_3856.north west) rectangle (ngmb_3856.south east);

\begin{scope}[on background layer]
    \node[inner sep=0pt, outer sep=0pt, anchor=north west] (over_ngm_3855) at (ngma_3855.north east) {\includegraphics[width=5cm, trim={1cm 1cm 4cm 1cm},clip]{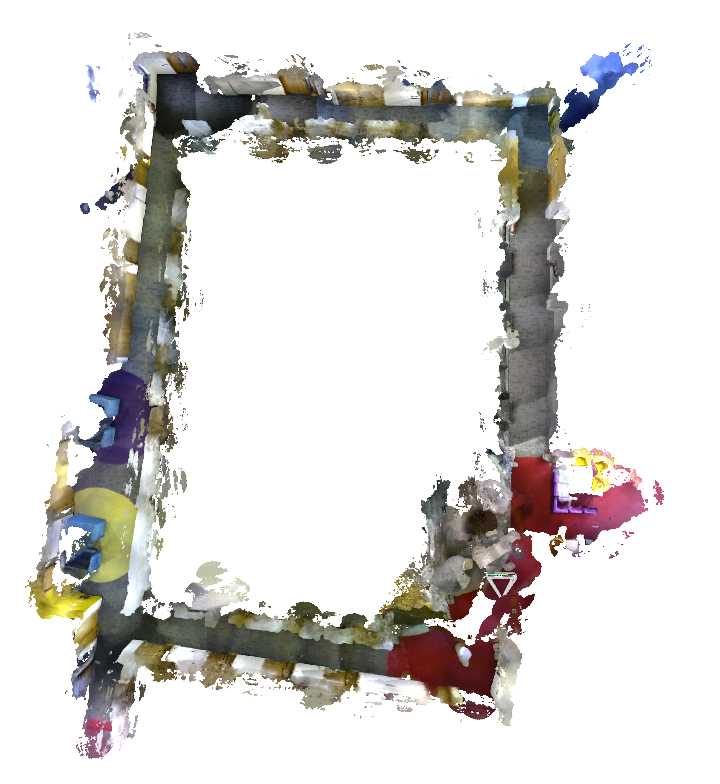}};
    \node[inner sep=0pt, outer sep=0pt, anchor=north west] (over_ngm_3856) at ($(over_ngm_3855) + (-1.3cm, 0.3cm)$) {\includegraphics[width=5cm, trim={0 1cm 4cm 1cm},clip]{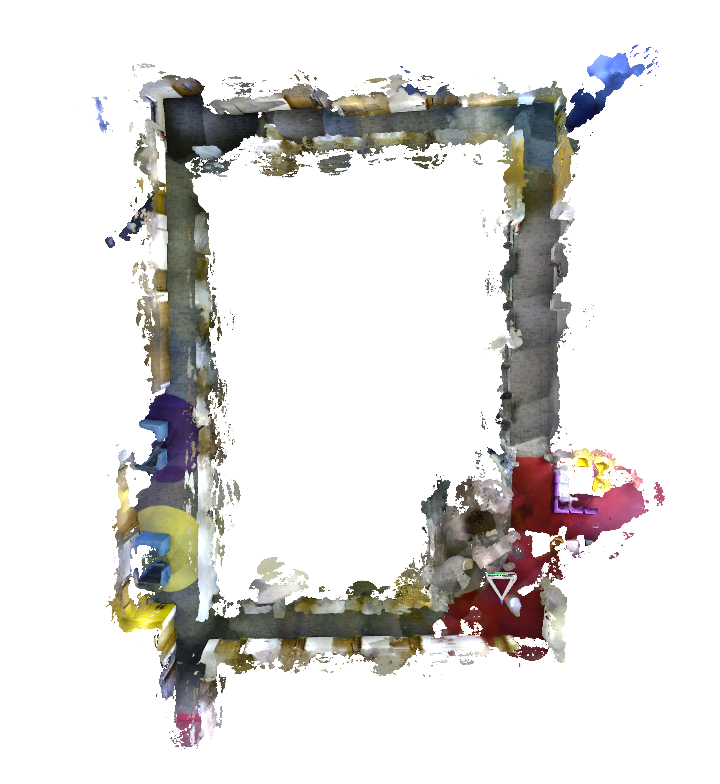}};

\end{scope}

\begin{scope}[color=cyan, shift={(3,3.4)}, rotate=-105]
    \path[draw, fill, very thick] coordinate (cama_3855) at (0,0) circle (0.05);
    \draw[very thick, fill, fill opacity=0.6] (0:0) -- (-30:0.5) -- (30:0.5) -- cycle;
\end{scope}
\begin{scope}[color=orange, shift={(3.4,0.76)}, rotate=-10]
    \path[draw, fill, very thick] coordinate (camb_3855) at  (0,0) circle (0.05);
    \draw[very thick, fill, fill opacity=0.6] (0:0) -- (-30:0.5) -- (30:0.5) -- cycle;
\end{scope}
\begin{scope}[color=cyan, shift={(4.69,0.56)}, rotate=-90]
    \path[draw, fill, very thick] coordinate (cama_3856) at  (0,0) circle (0.05);
    \draw[very thick, fill, fill opacity=0.6] (0:0) -- (-30:0.5) -- (30:0.5) -- cycle;
\end{scope}
\begin{scope}[color=orange, shift={(5.3,-1.75)}, rotate=0]
    \path[draw, fill, very thick] coordinate (camb_3856) at  (0,0) circle (0.05);
    \draw[very thick, fill, fill opacity=0.6] (0:0) -- (-30:0.5) -- (30:0.5) -- cycle;
\end{scope}

\draw[cyan, very thick] (ngma_3855.east) to[out=0,in=90] (cama_3855);

\draw[cyan, very thick] (ngma_3856) to[out=0,in=90] (cama_3856);

\draw[orange, very thick] (ngmb_3855) to[out=90,in=180] (camb_3855);

\draw[orange, very thick] (ngmb_3856) to[out=0,in=180] (camb_3856);

\draw[->, very thick] ($(over_ngm_3855.east) + (0.0,1.5)$) to[out=0,in=90] node[midway,above,yshift=1mm,xshift=-1mm,rotate=-40] {$+63\,\mathrm{ms}$} ($(over_ngm_3856.north east) + (-0.3,0.0)$);
\end{tikzpicture}%

%% file: figures/kintinuous_mono.tikz
\begin{tikzpicture}
\normalsize

\node[inner sep=0pt, outer sep=0pt, anchor=north west] (monob_3855) at (0,0) {\includegraphics[width=2cm, trim={2mm 0 2mm 0},clip]{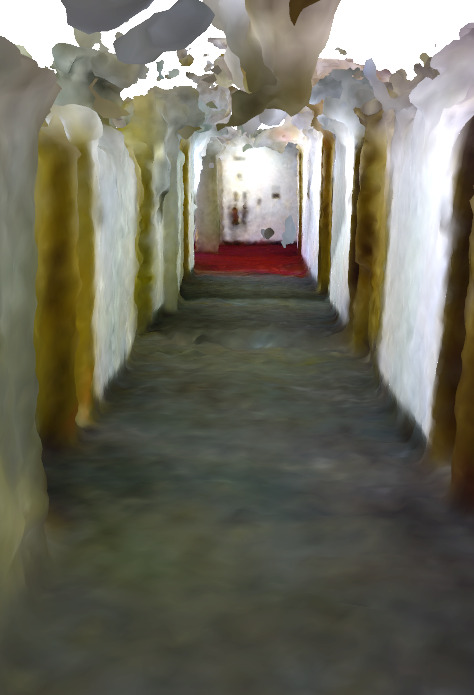}};
\node[inner sep=0pt, outer sep=0pt, anchor=south west] (monob_3856) at (monob_3855.south east) {\includegraphics[width=2cm, trim={2mm 0 2mm 0},clip]{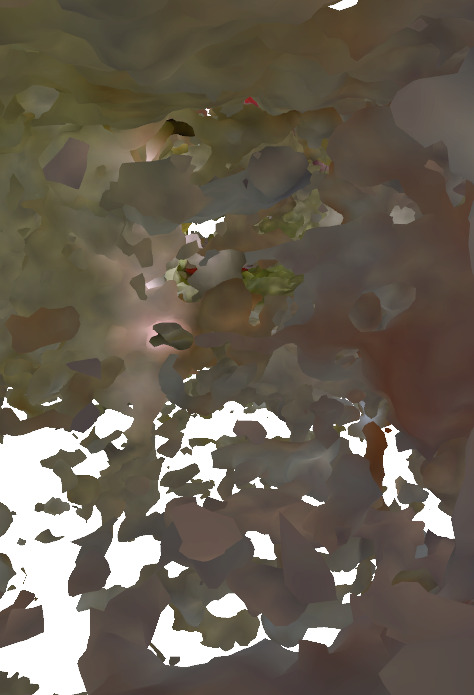}};

\node[inner sep=0pt, outer sep=0pt, anchor=south] (monoa_3856) at (monob_3855.north) {\includegraphics[width=2cm, trim={2mm 0 2mm 0},clip]{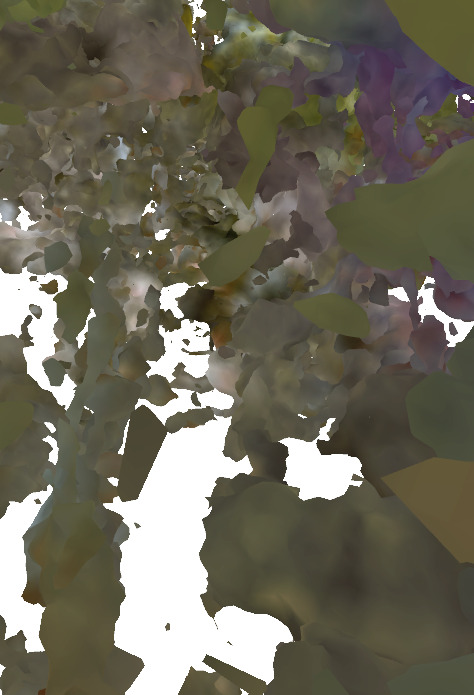}};
\node[inner sep=0pt, outer sep=0pt, anchor=south] (monoa_3855) at (monoa_3856.north) {\includegraphics[width=2cm, trim={2mm 0 2mm 0},clip]{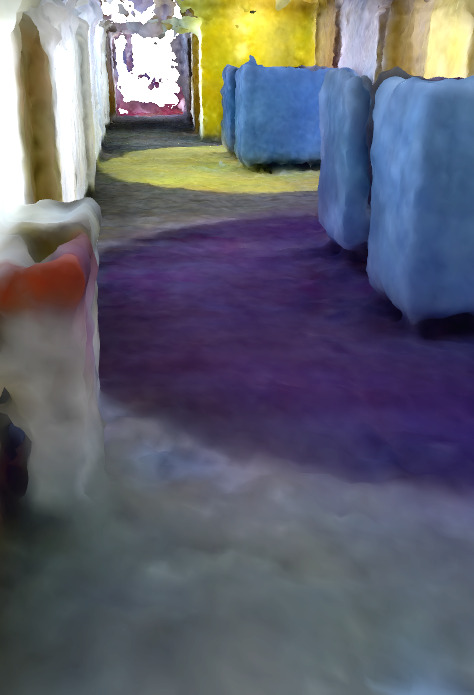}};

\path[preaction={clip,postaction={draw=cyan, line width=1.5mm}}] (monoa_3855.north west) rectangle (monoa_3855.south east);
\path[preaction={clip,postaction={draw=cyan, line width=1.5mm}}] (monoa_3856.north west) rectangle (monoa_3856.south east);
\path[preaction={clip,postaction={draw=orange, line width=1.5mm}}] (monob_3855.north west) rectangle (monob_3855.south east);
\path[preaction={clip,postaction={draw=orange, line width=1.5mm}}] (monob_3856.north west) rectangle (monob_3856.south east);

\begin{scope}[on background layer]
    \node[inner sep=0pt, outer sep=0pt, anchor=north west] (over_mono_3855) at (monoa_3855.north east) {\includegraphics[width=5cm, trim={1cm 1cm 4cm 1cm},clip]{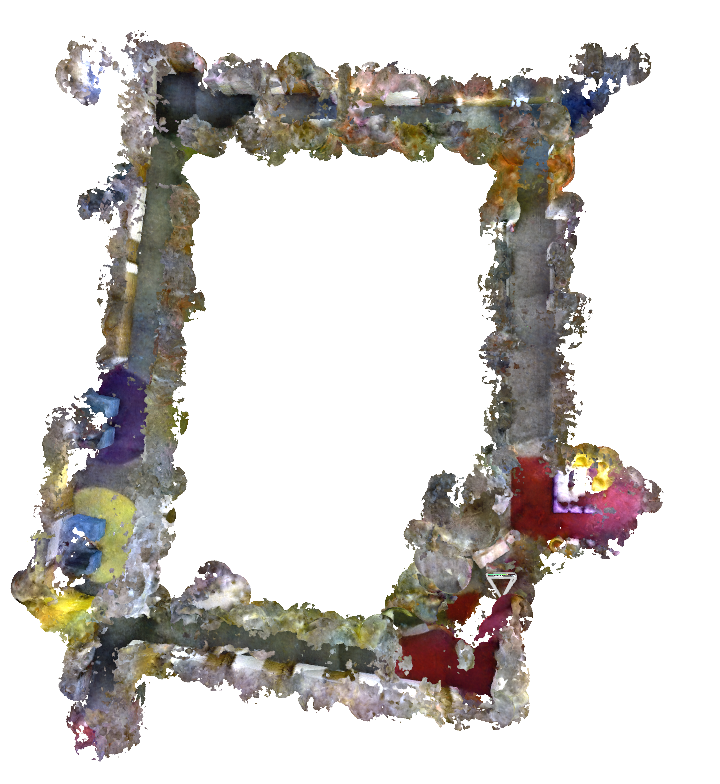}};
    \node[inner sep=0pt, outer sep=0pt, anchor=north west] (over_mono_3856) at ($(over_mono_3855) + (-1.3cm, 0.3cm)$) {\includegraphics[width=5cm, trim={0 1cm 4cm 1cm},clip]{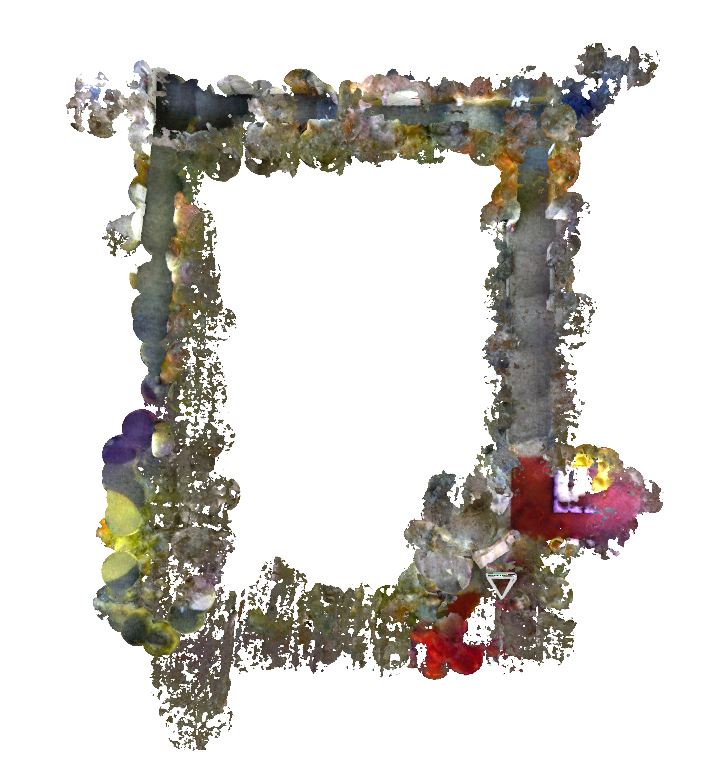}};
\end{scope}

\node[inner sep=0pt, outer sep=0pt, anchor=south west] (monob_3860) at ($(monob_3855.south -| over_mono_3856.east) + (0.05,0.8)$) {\includegraphics[width=2cm, trim={2mm 0 2mm 0},clip]{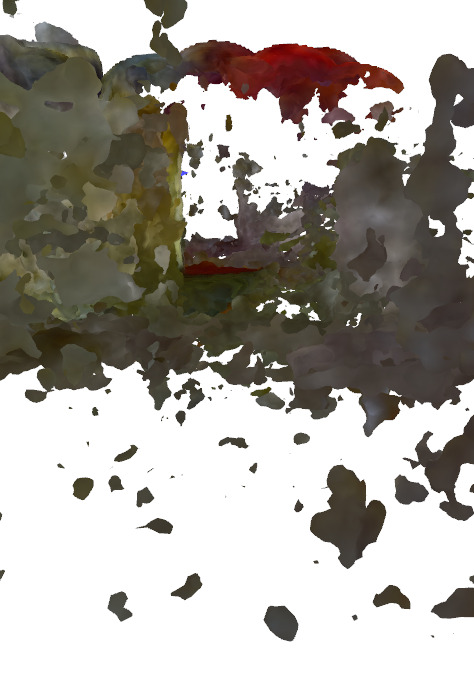}};
\node[inner sep=0pt, outer sep=0pt, anchor=south west] (monob_3870) at (monob_3860.south east) {\includegraphics[width=2cm, trim={2mm 0 2mm 0},clip]{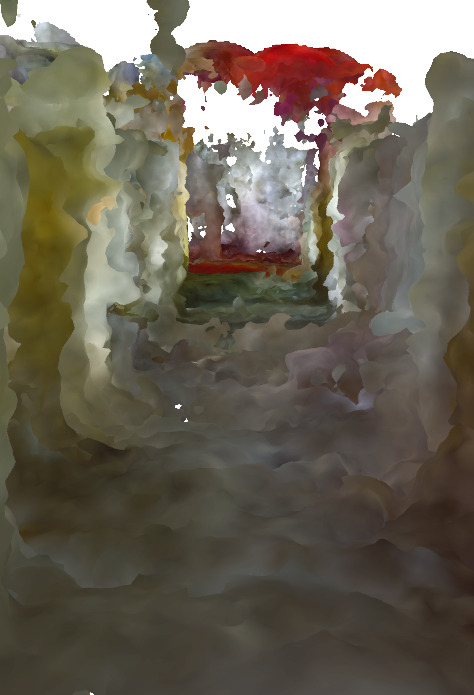}};
\node[inner sep=0pt, outer sep=0pt, anchor=south west] (monob_3900) at (monob_3870.south east) {\includegraphics[width=2cm, trim={2mm 0 2mm 0},clip]{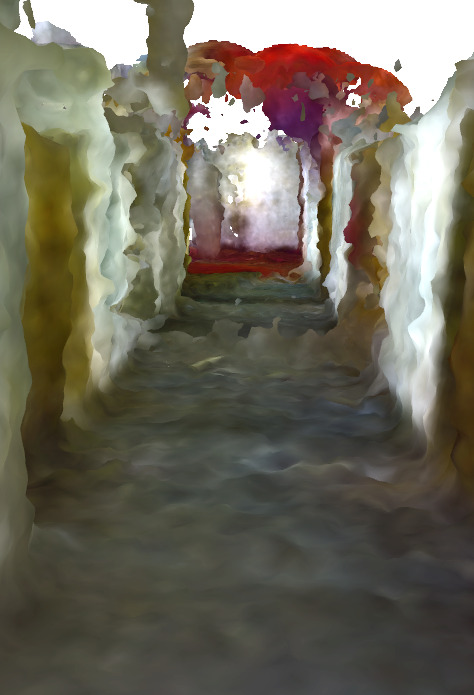}};
\node[inner sep=0pt, outer sep=0pt, anchor=south west] (monob_4000) at (monob_3900.south east) {\includegraphics[width=2cm, trim={2mm 0 2mm 0},clip]{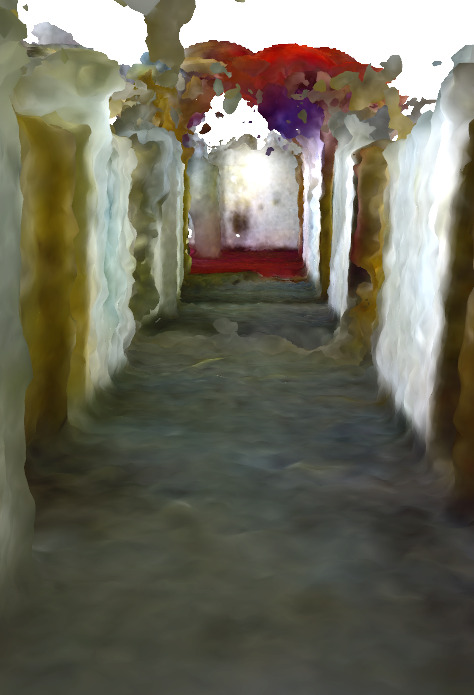}};

\node[inner sep=0pt, outer sep=0pt, anchor=north west] (monoa_3860) at ($(monoa_3855.north -| monob_3860.west) + (0.0,-0.8)$) {\includegraphics[width=2cm, trim={2mm 0 2mm 0},clip]{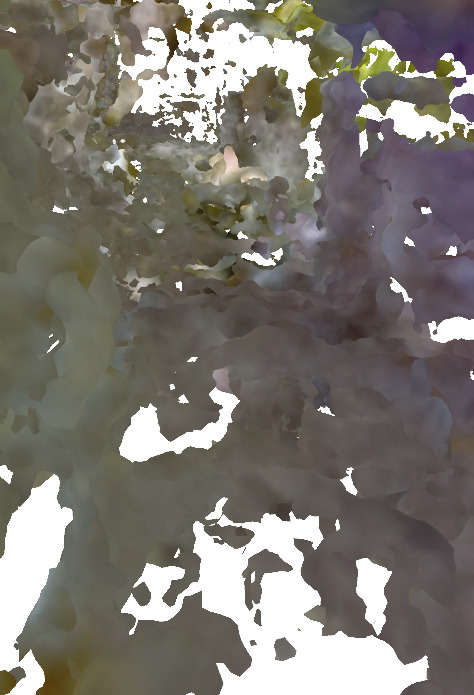}};
\node[inner sep=0pt, outer sep=0pt, anchor=south west] (monoa_3870) at (monoa_3860.south east) {\includegraphics[width=2cm, trim={2mm 0 2mm 0},clip]{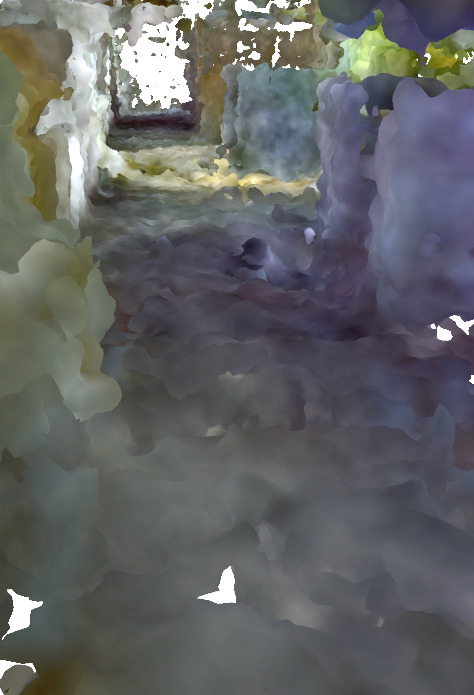}};
\node[inner sep=0pt, outer sep=0pt, anchor=south west] (monoa_3900) at (monoa_3870.south east) {\includegraphics[width=2cm, trim={2mm 0 2mm 0},clip]{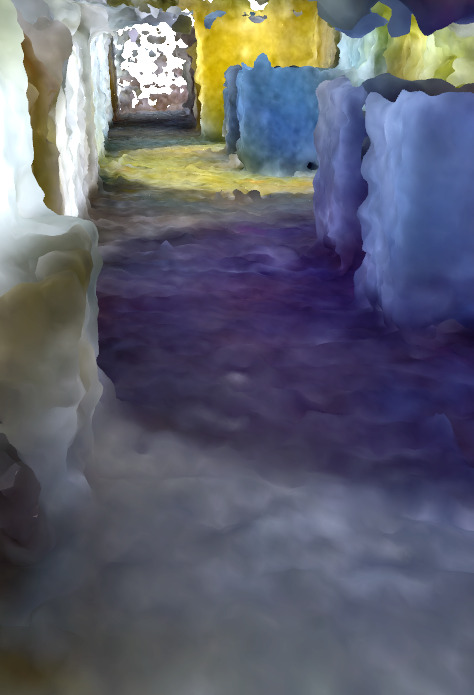}};
\node[inner sep=0pt, outer sep=0pt, anchor=south west] (monoa_4000) at (monoa_3900.south east) {\includegraphics[width=2cm, trim={2mm 0 2mm 0},clip]{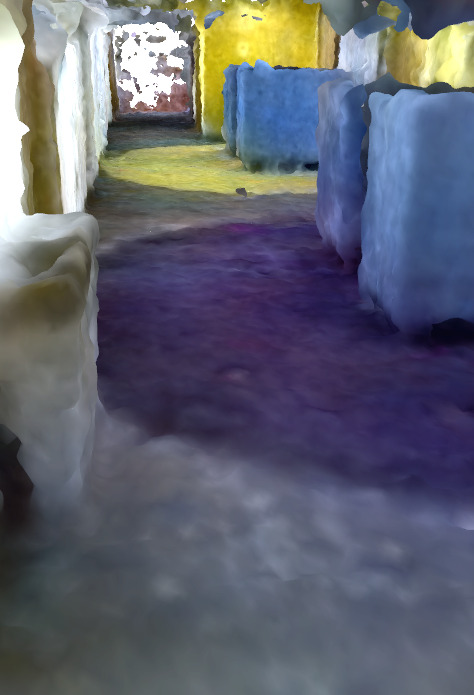}};

\path[preaction={clip,postaction={draw=cyan, line width=1.5mm}}] (monoa_3860.north west) rectangle (monoa_4000.south east);
\path[preaction={clip,postaction={draw=orange, line width=1.5mm}}] (monob_3860.north west) rectangle (monob_4000.south east);

\draw[->, very thick] ($(monoa_3860.west)!0.5!(monob_3860.west) + (0.3,0.0)$) -- ($(monob_4000.east)!0.5!(monoa_4000.east) + (-0.3,0.0)$);
\draw[very thick] ($(monob_3860.south)!0.5!(monoa_3860.north) + (0.0,0.1)$) -- ($(monob_3860.south)!0.5!(monoa_3860.north) + (0.0,-0.1)$) node[below] {$+0.26\,\mathrm{s}$};
\draw[very thick] ($(monob_3870.south)!0.5!(monoa_3870.north) + (0.0,0.1)$) -- ($(monob_3870.south)!0.5!(monoa_3870.north) + (0.0,-0.1)$) node[below] {$+0.79\,\mathrm{s}$};
\draw[very thick] ($(monob_3900.south)!0.5!(monoa_3900.north) + (0.0,0.1)$) -- ($(monob_3900.south)!0.5!(monoa_3900.north) + (0.0,-0.1)$) node[below] {$+2.38\,\mathrm{s}$};
\draw[very thick] ($(monob_4000.south)!0.5!(monoa_4000.north) + (0.0,0.1)$) -- ($(monob_4000.south)!0.5!(monoa_4000.north) + (0.0,-0.1)$) node[below] {$+7.68\,\mathrm{s}$};
\draw[->, very thick] ($(over_mono_3855.east) + (0.0,1.5)$) to[out=0,in=90] node[midway,above,yshift=1mm,xshift=-1mm,rotate=-40] {$+53\,\mathrm{ms}$} ($(over_mono_3856.north east) + (-0.3,0.0)$);

\begin{scope}[on background layer]
    \node[inner sep=0pt, outer sep=0pt, anchor=west] (over_mono_4000) at ($(0,1.5 -| monoa_4000.east) + (0.05,0.0)$) {\includegraphics[width=5cm, trim={3cm 1cm 4cm 2cm},clip]{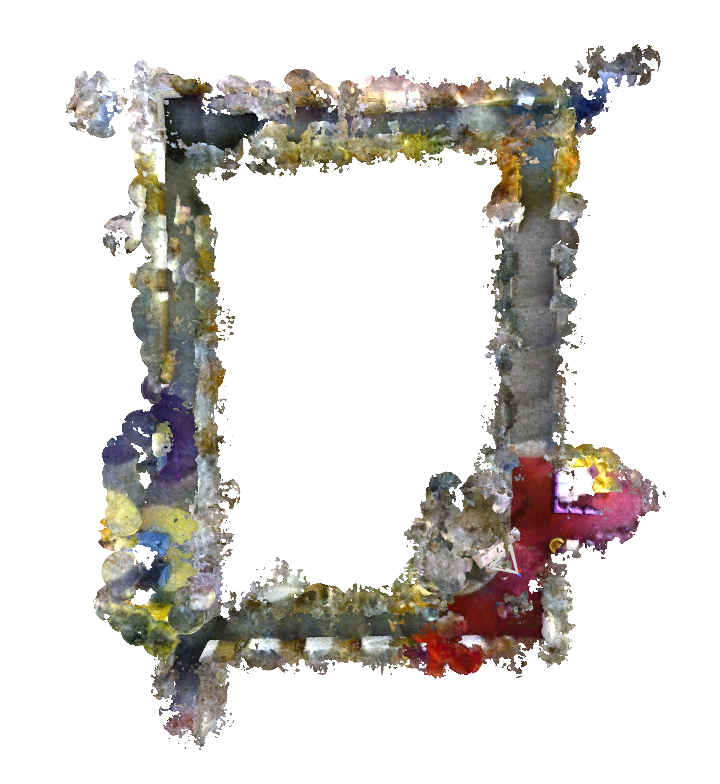}};
\end{scope}

\begin{scope}[color=cyan, shift={(3,3.4)}, rotate=-105]
    \path[draw, fill, very thick] coordinate (cama_3855) at (0,0) circle (0.05);
    \draw[very thick, fill, fill opacity=0.6] (0:0) -- (-30:0.5) -- (30:0.5) -- cycle;
\end{scope}
\begin{scope}[color=orange, shift={(3.4,0.76)}, rotate=-10]
    \path[draw, fill, very thick] coordinate (camb_3855) at  (0,0) circle (0.05);
    \draw[very thick, fill, fill opacity=0.6] (0:0) -- (-30:0.5) -- (30:0.5) -- cycle;
\end{scope}
\begin{scope}[color=cyan, shift={(4.69,0.56)}, rotate=-90]
    \path[draw, fill, very thick] coordinate (cama_3856) at  (0,0) circle (0.05);
    \draw[very thick, fill, fill opacity=0.6] (0:0) -- (-30:0.5) -- (30:0.5) -- cycle;
\end{scope}
\begin{scope}[color=orange, shift={(5.3,-1.75)}, rotate=0]
    \path[draw, fill, very thick] coordinate (camb_3856) at  (0,0) circle (0.05);
    \draw[very thick, fill, fill opacity=0.6] (0:0) -- (-30:0.5) -- (30:0.5) -- cycle;
\end{scope}

\begin{scope}[shift={(12.6,1.2)}]
    \begin{scope}[color=cyan, shift={(4.69,0.56)}, rotate=-90]
        \path[draw, fill, very thick] coordinate (cama_4000) at  (0,0) circle (0.05);
        \draw[very thick, fill, fill opacity=0.6] (0:0) -- (-30:0.5) -- (30:0.5) -- cycle;
    \end{scope}
    \begin{scope}[color=orange, shift={(5.3,-1.75)}, rotate=0]
        \path[draw, fill, very thick] coordinate (camb_4000) at  (0,0) circle (0.05);
        \draw[very thick, fill, fill opacity=0.6] (0:0) -- (-30:0.5) -- (30:0.5) -- cycle;
    \end{scope}
\end{scope}

\draw[cyan, very thick] (monoa_3855.east) to[out=0,in=90] (cama_3855);

\draw[cyan, very thick] (monoa_3856) to[out=0,in=90] (cama_3856);

\draw[orange, very thick] (monob_3855) to[out=90,in=180] (camb_3855);

\draw[orange, very thick] (monob_3856) to[out=0,in=180] (camb_3856);

\draw[cyan, very thick] (monoa_4000) to[out=0,in=90] (cama_4000);

\draw[orange, very thick] (monob_4000) to[out=0,in=180] (camb_4000);

\end{tikzpicture}%

%% file: figures/svmv.tikz
\begin{tikzpicture}
\scriptsize

\node[inner sep=0pt, outer sep=0pt] (sv) at (0,0) {\includegraphics[width=0.4\linewidth]{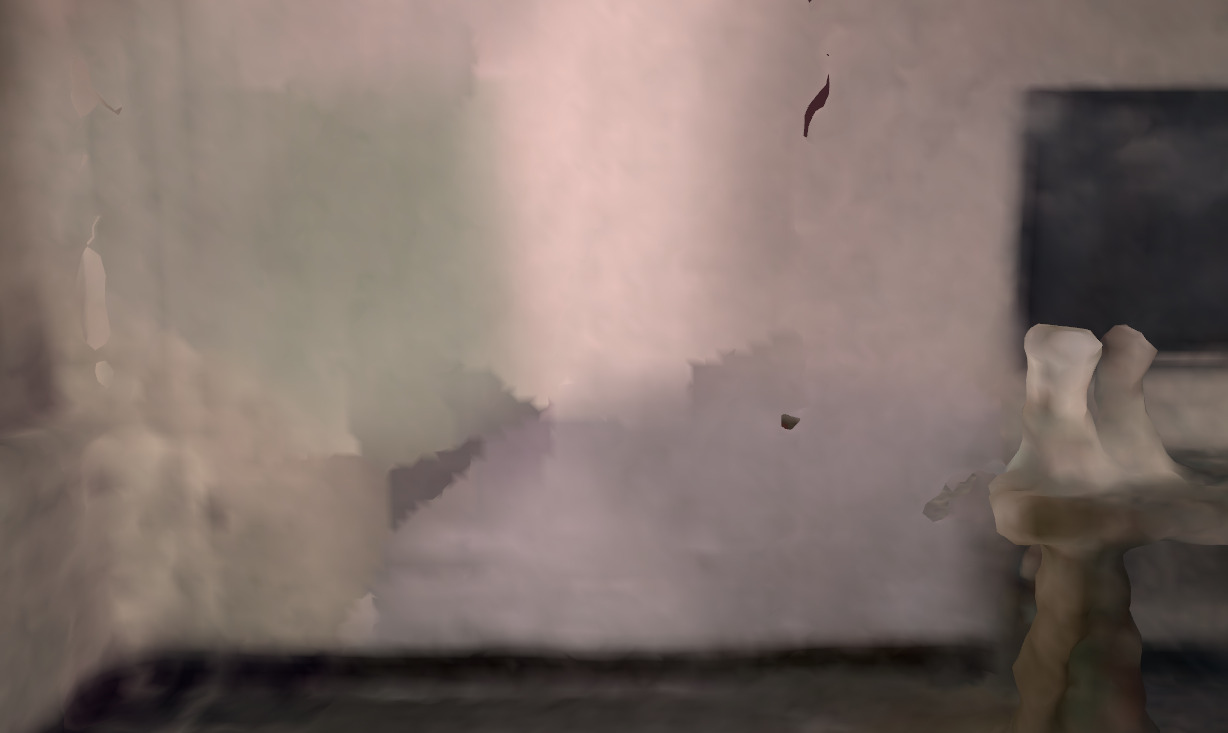}};
\node[inner sep=3pt, outer sep=3pt, anchor=north west, fill=white, fill opacity=0.5, text opacity=1.0] (svlabel) at (sv.north west) {Single-view};

\node[inner sep=0pt, outer sep=0pt, anchor=north west] (mv) at ($(sv.north east) + (0.01\linewidth,0.0)$ ) {\includegraphics[width=0.4\linewidth]{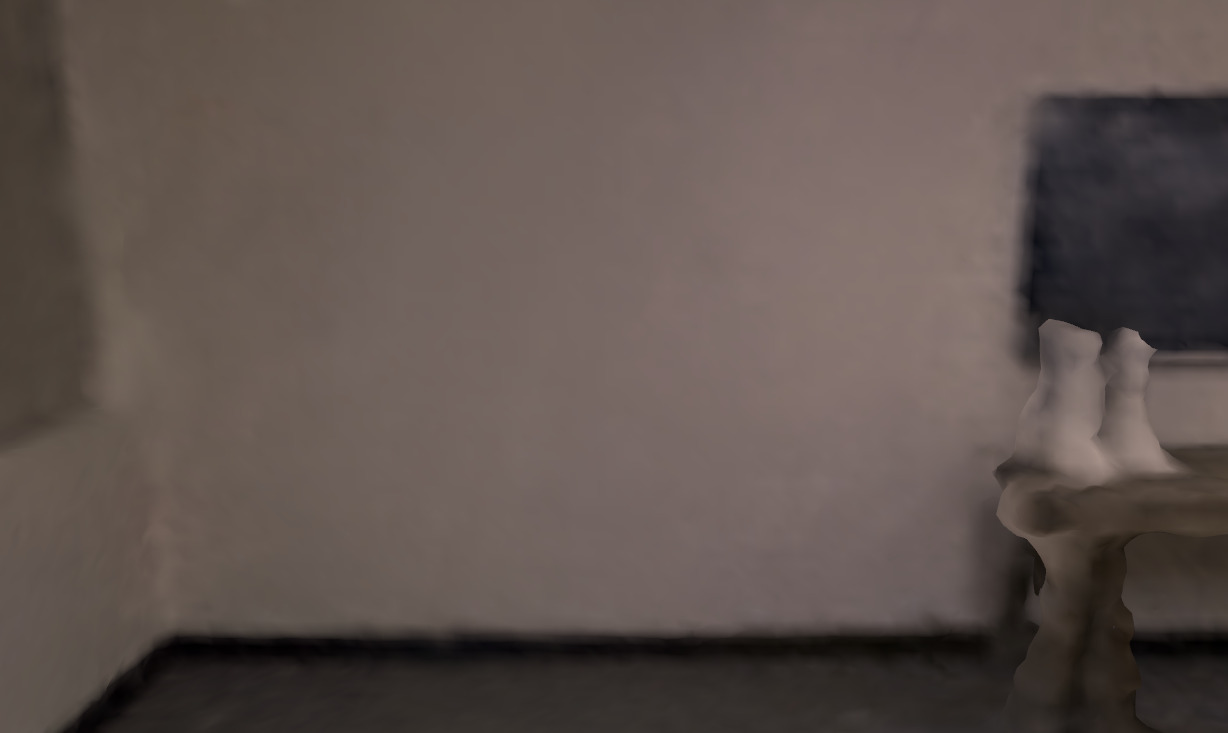}};
\node[inner sep=3pt, outer sep=3pt, anchor=north west, fill=white, fill opacity=0.5, text opacity=1.0] (mvlabel) at (mv.north west) {Multi-view};

\end{tikzpicture}%

%% file: supplementary_content.tex
\section{Supplementary Video} In the supplementary video (\url{https://roym899.github.io/videos/ngm_supplementary.mp4}), we first showcase the result on the Kintinuous scene visualizing the field centers and their movement during loop closure. The second part shows a comparison with an ablated, single-field version of our method (supplementing Fig.\ 6 in the main paper). In the last part, we show the mapping on the large-scale \texttt{apt0} scene which includes two smaller loop closures. More videos can be found on our project page \url{https://kth-rpl.github.io/neural_graph_mapping/}.

\section{Limitations and Discussion} While our approach enables efficient integration of loop closure into the volumetric map, it is not without drawbacks. In particular, the multi-field representation is less memory-efficient compared to monolithic neural field representations, since those can use the available network capacity relatively unconstrained, whereas this adaptiveness is constrained to the local spheres in our approach. Sharing hash tables among multiple fields might be one direction to reduce the memory overhead.

Another downside of our method is that the neural scene representation is currently not used to improve the SLAM result and a tighter integration of dense mapping and sparse tracking could lead to improved robustness. A similar limitation applies to GO-SLAM \cite{zhang2023go}, which also does not use the neural map for pose estimation. In \cite{liso2024loopy} this property is referred to as ``decoupled''. While ``coupled'' approaches that use the map for frame-to-map alignment appear elegant they are not without drawbacks. Current methods for frame-to-map alignment using neural representation have a small basin of convergence and hence only work for slow sequences with small baselines. They also typically rely on depth measurements to be available. In contrast, feature-based methods can potentially estimtate poses even for large baselines and can directly benefit from progress in image matching. In the future we want to investigate how to combine sparse features and neural maps in a principled way.

\section{Method Details}

\subsection{Sampling Strategy}

As described in Sec. 3.3 and illustrated in Fig.\ 4 in the main paper, a three-stage sampling procedure is used. First, a subset of fields is sampled, then rays are sampled for each field, and finally points are sampled along each ray. In the following we describe the details of each stage.

\paragraph{Fields} Especially with larger scenes, it is important that recently added fields and those currently observed are optimized with a higher rate than out-of-view fields that have already been optimized before. To achieve this, the currently observed fields $\mathcal{F}^{\mathrm{obs}}_t\subseteq \mathcal{F}_t$ are determined and sampled with a higher probability. Specifically, a total of $N_\mathrm{f}$ fields are sampled; half from the currently observed fields and the remaining ones from all fields $\mathcal{F}_t$ discarding duplicates.

\paragraph{Rays} To sample supervision targets, each sampled field $i$ is approximated by a set of points $\boldsymbol{q}_j^i, j=1,...,N_\mathrm{approx}$ sampled uniformly on the field's sphere of radius $r$. These points are projected into all keyframes. A field is considered visible in a keyframe, if at least one of the field's points $\boldsymbol{q}_j^i$ is inside the keyframe's frustum and the projected depth of $\boldsymbol{q}_j^i$ is smaller than the observed depth at the projected 2D point. This yields a set of keyframes $\mathcal{K}_t^i \subseteq \mathcal{K}_t$. $N_\mathrm{r}$ rays per field (i.e., target rays) are then sampled via the 2D bounding boxes of the projected points $\boldsymbol{q}^i_j$ in the keyframes $\mathcal{K}_t^i$. For each target ray $(\boldsymbol{o},\boldsymbol{d})$ the closest point to the field center is computed as $\boldsymbol{o}+l_\mathrm{c}\boldsymbol{d}$ and only a ray segment $[l_\mathrm{c}-r,l_\mathrm{c}+r]$ covering the sphere will be considered for optimization.

\paragraph{Points} Given a ray segment $[l_\mathrm{min},l_\mathrm{max}]$, $N_\mathrm{up}$ points are uniformly sampled across the segment, and $N_\mathrm{dp}$ points are uniformly sampled in the truncation interval $\tau$ around the observed depth, that is, in the interval $[l_\mathrm{obs}-\tau, l_\mathrm{obs}+\tau]$; the full ray interval is used, if there is no depth measurement or $l_\mathrm{obs} \notin [l_\mathrm{min},l_\mathrm{max}]$. This yields a total of $N_\mathrm{p}=N_\mathrm{up}+N_\mathrm{dp}$ query points per ray segment during optimization.

In total, each optimization iteration will contain a maximum of $N_\mathrm{f}N_\mathrm{r}N_\mathrm{p}$ query points.

\subsection{K-Nearest Neighbors Queries}\label{sec:knn}

We compute the color and signed distance at a query point as the weighted average of the $k$ nearest fields. Specifically, let $\boldsymbol{x}\in\mathbb{R}^3$ denote the query point. Let $\boldsymbol{c}_i, s_i, d_i, i=1,...,k$ denote the returned color, signed distance, and distance to the field center for the $k$ nearest fields for the query point $\boldsymbol{x}$. We compute weights based on the softmax of the negative distances, that is, 
\begin{equation}
    u_i=\frac{e^{-\xi d_i}}{\sum_{j=1}^{k}e^{-\xi d_j}},
\end{equation}
where $\xi$ determines the transition speed. The combined color and signed distance are then computed as a weighted sum, that is, $\boldsymbol{c}=\sum_{i=1}^{k} u_i \boldsymbol{c}_i$ and  $s=\sum_{i=1}^{k} u_i s_i$.

We always use the $k$ nearest fields even when only the closest field is within radius $r$. However, we set $\xi$ sufficiently high such that the transition region becomes small and fields with $d_i\gg r$ will have no significant contribution to the final value. This is a feasible strategy, since fields are optimized for all ray segments intersecting them even when the segment is terminating outside the sphere. Hence, each field will in practice capture a region larger than a sphere with radius $r$. 

\begin{figure}
    \scriptsize
    \centering
    \begin{subfigure}{0.3\linewidth}
        \centering
        \includegraphics[width=\linewidth]{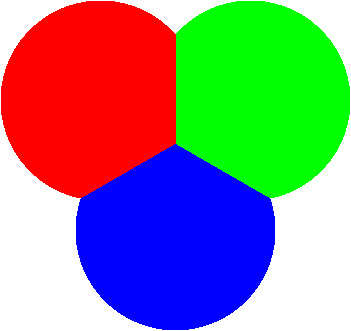}
        $k=1$
    \end{subfigure}
    \begin{subfigure}{0.3\linewidth}
        \centering
        \includegraphics[width=\linewidth]{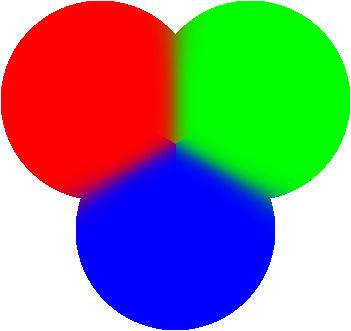}
        $k=2$
    \end{subfigure}
    \begin{subfigure}{0.3\linewidth}
        \centering
        \includegraphics[width=\linewidth]{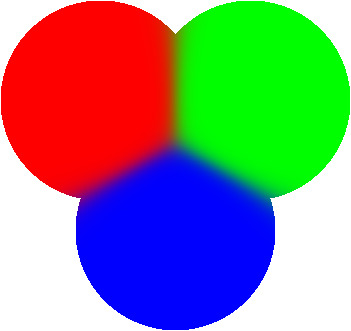}
        $k=3$
    \end{subfigure}
    
    \vspace{1em}
    
    \begin{subfigure}{0.3\linewidth}
        \centering
        \includegraphics[width=\linewidth]{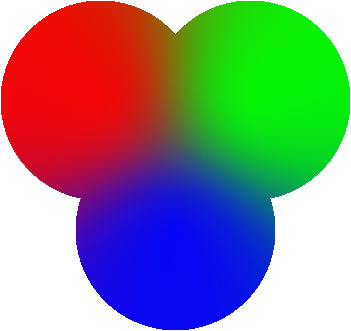}
        $\xi=2.5$
    \end{subfigure}
    \begin{subfigure}{0.3\linewidth}
        \centering
        \includegraphics[width=\linewidth]{figures/knn_queries/knn_3_10.0_False.jpg}
        $\xi=10.0$
    \end{subfigure}
    \begin{subfigure}{0.3\linewidth}
        \centering
        \includegraphics[width=\linewidth]{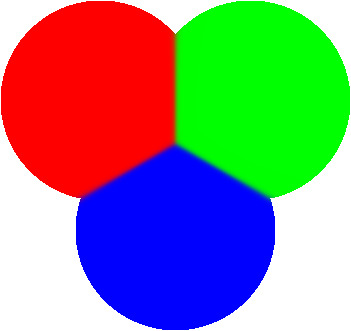}
        $\xi=40.0$
    \end{subfigure}
    
    \vspace{1em}
    
    \begin{subfigure}{0.3\linewidth}
        \centering
        \includegraphics[width=\linewidth]{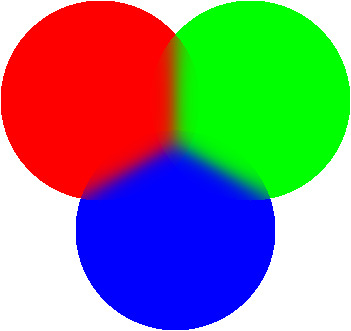}
        Excluding $d_i>r$
    \end{subfigure}
    \begin{subfigure}{0.3\linewidth}
        \centering
        \includegraphics[width=\linewidth]{figures/knn_queries/knn_3_10.0_False.jpg}
        Including $d_i>r$
    \end{subfigure}
    \caption{Visualization of k-nearest neighbor distance-based averaging. The top row shows the effect of varying $k$. The second row shows the effect of varying $\xi$. The last row shows the effect of excluding fields with distances $d_i$ greater than the field radius $r$ from the averaging.}
    \label{fig:knnvis}
    \vspace*{-\baselineskip}
\end{figure}

\begin{table*}[htb!]
    \centering
    \scriptsize
    \renewcommand{\arraystretch}{1.2}
    \begin{threeparttable}
        \caption{Overview of parameters.}
        \begin{tabular}{lcm{13cm}}
            \toprule
            Parameter & Value & Description \\
            \midrule
            $r$ & $1\,\mathrm{m}$ & Field radius\\
            \rowcolor{LightGray} $N_\mathrm{f}$ & $2$ & Number of fields optimized in parallel in each iterations\\
            $N_\mathrm{r}$ & $512$ & Number of ray segments sampled per field during optimization\\
            \rowcolor{LightGray} $N_\mathrm{up}$ & $8$ &  Number of uniformly sampled points distributed along each ray segment during optimization\\
            $N_\mathrm{dp}$ & $16$ & Number of depth-guided points distributed along each ray segment during optimization\\
            \rowcolor{LightGray} $\tau$ & $0.1\,\mathrm{m}\text{ or }0.2\,\mathrm{m}\smash{^\ast}$ & Truncation distance; used for scaling depth-guided sampling and for capping the supervision range (i.e., dividing samples into free-space samples and TSDF samples)\\
            $\eta$ & $20.0$ & Determines how fast occupancy probability decays around surfaces\\
            \rowcolor{LightGray} $k$ & $2$ & Number of nearest neighbors used during evaluation\\
            $\xi$ & $10.0$ & Distance weighing determining transition speed between two fields\\
            \rowcolor{LightGray} $L$ & $1$ & Number of MLP layers following the permutohedral encoding; excluding the final linear layer\\
            $T$ & $2^{12}$ & Hash table size for permutohedral encoding\\
            \rowcolor{LightGray} $N_\mathrm{levels}$ & $16$ & Number of resolution levels for permutohedral encoding \\ 
            $N_\mathrm{fpl}$ & $16$ & Number of features per level for permutohedral encoding \\ 
            \rowcolor{LightGray} $r_\mathrm{coarse}$ & $0.1$ & Coarsest resolution for permutohedral encoding \\ 
            $r_\mathrm{fine}$ & $0.0001$ & Finest resolution for permutohedral encoding \\ 
            \rowcolor{LightGray} $\lambda_\mathrm{color}$ & $1.0$ & Weight of color loss\\
            $\lambda_\mathrm{depth}$ & $1.0$ & Weight of depth loss\\
            \rowcolor{LightGray} $\lambda_\mathrm{fs}$ & $40.0$ & Weight of free-space loss\\
            $\lambda_\mathrm{tsdf}$ & $50.0$ & Weight of TSDF loss\\
            \rowcolor{LightGray} $\delta$ & $5\,\mathrm{cm}$ & Huber loss threshold\\
            $\gamma$ & $1\times10^{-3}$ & Learning rate used for Adam optimizer \cite{kingma2014adam}\\
            \rowcolor{LightGray} $\lambda$ & $1\times10^{-5}$ & Weight decay used for Adam optimizer \cite{kingma2014adam}\\
             \bottomrule
        \end{tabular}
        \begin{tablenotes}
            \item[$^\ast$] The truncation distance is increased for real-world datasets to account for the increase in depth noise.
        \end{tablenotes}
        \label{tab:parameters}
    \end{threeparttable}
    \vspace*{-\baselineskip}
\end{table*}

\Cref{fig:knnvis} illustrates the effect of this weighted averaging for different values of $\xi$, $k$, and with and without $u_i=0$ for $d_i>r$ on a 2D toy example with three fields of fixed color. Note that the distance-weighted averaging leads to smooth transitions in the overlapping regions. When forcing $u_i=0$ for $d_i > r$ transitions on the boundaries are unavoidable, hence we opt for the strategy described in the previous paragraph. For the experiments we use $k=2$ and $\xi=10$.

\subsection{Parameters}

In \cref{tab:parameters} we provide a full list of parameters, the used value to achieve the experimental results, and a brief description. Parameters were tuned manually and the same setting is used for all experiments (with the exception of $\tau$, which is increased for the real-world datasets).

\section{Experiment Details}

\paragraph{Baseline Setup} All baselines are evaluated using the parameters published as part of the published code. For additional datasets for which no parameters were provided, the most similar dataset's parameters were adopted (i.e., for Replica-Big the provided setup for Replica is used; for Kintinuous the setup for ScanNet). Scene boundaries were manually adjusted to cover the observed area with extra margin to account for errors in positioning. 

\begin{figure}[htb!]
    \centering
    \begin{subfigure}{0.49\linewidth}
        \centering
        \includegraphics[width=\linewidth]{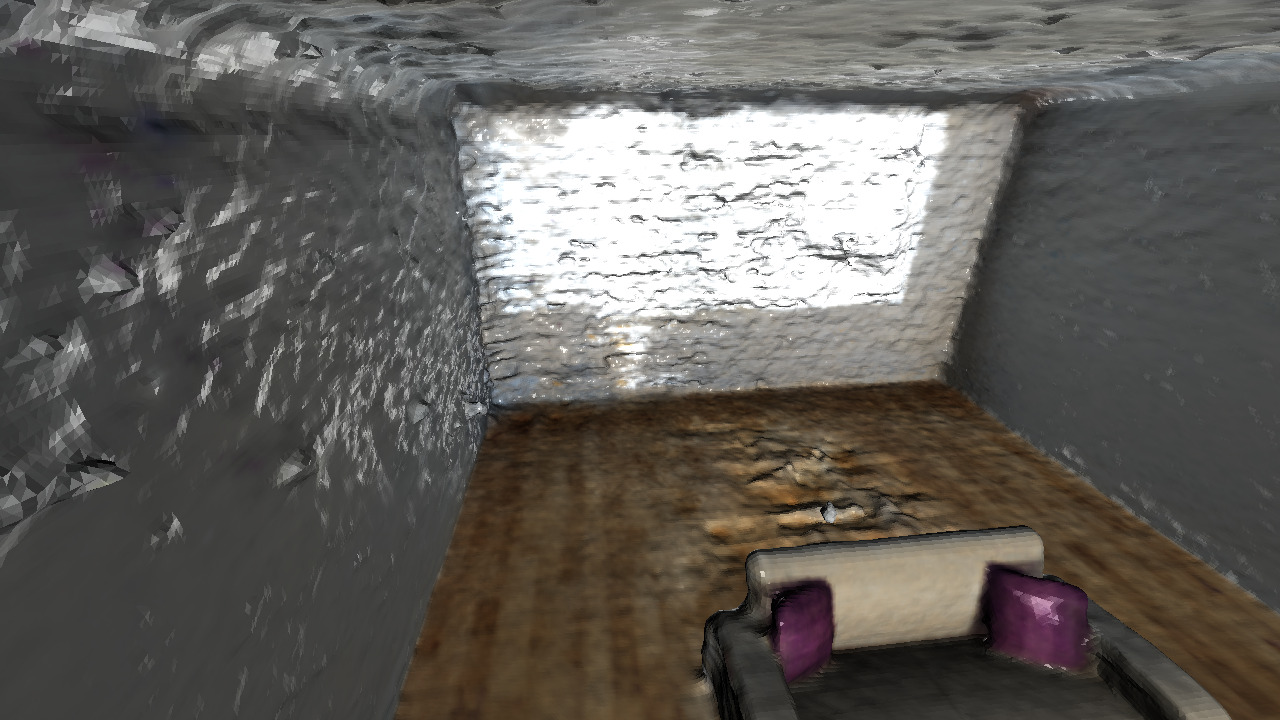}
        \caption{With axis-alignment}
    \end{subfigure}
    \begin{subfigure}{0.49\linewidth}
        \centering
        \includegraphics[width=\linewidth]{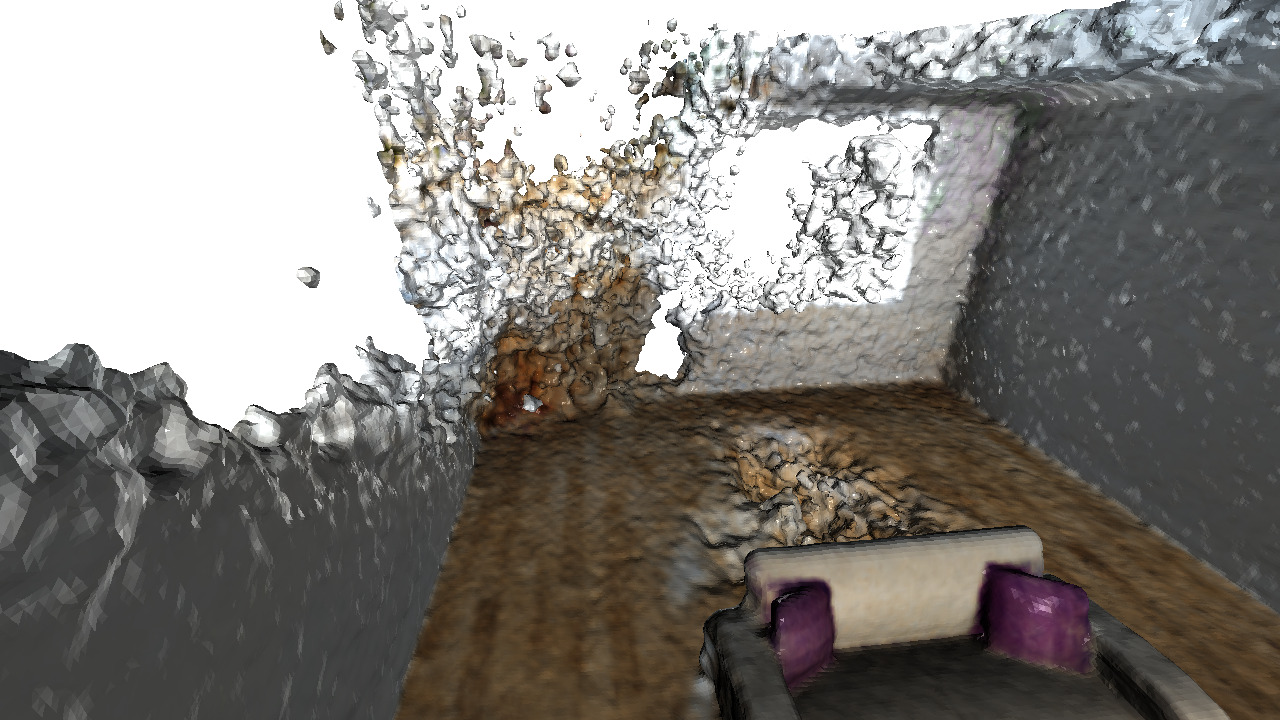}
        \caption{Without axis-alignment}
    \end{subfigure}
    \caption{Co-SLAM result with and without ground-truth initialization. Ground-truth initialization leads to axis-aligned walls and floors, which in turn leads to significantly better scene completion.}\label{fig:axisalignment}
    \vspace*{-\baselineskip}
\end{figure}

In our experiments, we noticed that Co-SLAM \cite{wang2023co} uses the ground-truth pose of the first frame to initialize the SLAM system, which leads to axis-aligned planes. We found that planes (such as walls, floors, and ceilings) which are axis-aligned are significantly better completed using the one-blob encoding \cite{muller2019neural} than generally-oriented planes. Therefore, for a fair comparison, we modified Co-SLAM's implementation to start from a random orientation instead. We note that this mainly reduces qualitative scene completion, however, on one of the Replica scenes it leads to tracking issues and hence poor reconstruction results. \Cref{fig:axisalignment} shows an example of the scene completion capability of Co-SLAM with and without ground-truth initialization (i.e., with and without axis-aligned planes).

\subsection{Evaluation Protocol}\label{sec:reconeval}

Let $\tensor*[^{\mathrm{w}}]{\mathcal{M}}{_{\mathrm{gt}}}$ and $\tensor*[^{\tilde{\mathrm{w}}}]{\mathcal{M}}{_{\mathrm{est}}}$ denote the ground-truth mesh and estimated mesh, respectively. We assume that $\tensor*[^{\tilde{\mathrm{w}}}]{\mathcal{M}}{_{\mathrm{est}}}$ has already been globally aligned with $\tensor*[^{\mathrm{w}}]{\mathcal{M}}{_{\mathrm{gt}}}$, which is typically achieved by either aligning the first frame in the sequence or by aligning the trajectories using the Umeyama algorithm. Starting from $\tensor*[^{\mathrm{w}}]{\mathcal{M}}{_{\mathrm{gt}}}$ and $\tensor*[^{\tilde{\mathrm{w}}}]{\mathcal{M}}{_{\mathrm{est}}}$ further preprocessing steps are performed before the evaluation metrics are computed.
\begin{enumerate}
    \item Unobserved parts of the ground-truth mesh $\tensor*[^{\mathrm{w}}]{\mathcal{M}}{_{\mathrm{gt}}}$ are removed. In particular, we apply two removal steps. First, vertices falling more than $\mathrm{2}\,\mathrm{cm}$ outside the scene bounding box are removed. The scene bounding box is computed as the intersection of a manually-set bounding box (used to exclude outliers in the depth map present in some scenes) and an automatically computed bounding box (based on the ground-truth trajectory and depth maps). Second, vertices that are not in front or up to $3\,\mathrm{cm}$ behind any rendered depth map are removed. These depth maps are rendered from the ground-truth trajectory and from additional virtual views manually placed to improve the evaluation of scene completion (same as in Co-SLAM \cite{wang2023co}). This yields a culled ground-truth mesh used for evaluation $\tensor*[^{\mathrm{w}}]{{\mathcal{M}}}{^{\ast}_{\mathrm{gt}}}$.
    \item To further equalize slight differences in alignment between different methods, we perform another alignment step using point-to-plane-based iterative closest point from $\tensor*[^{\tilde{\mathrm{w}}}]{\mathcal{M}}{_{\mathrm{est}}}$'s vertices to $\tensor*[^{\mathrm{w}}]{\mathcal{M}}{_{\mathrm{gt}}}$'s vertices yielding an aligned estimated mesh $\tensor*[^{\mathrm{w}}]{\mathcal{M}}{_{\mathrm{est}}}$.
    \item The aligned estimated mesh $\tensor*[^{\mathrm{w}}]{{\mathcal{M}}}{_{\mathrm{est}}}$ follows the same removal process as the ground-truth mesh (see step 1 above) yielding the culled estimated mesh used for evaluation $\tensor*[^{\mathrm{w}}]{{\mathcal{M}}}{^{\ast}_{\mathrm{est}}}$.
\end{enumerate}

For the evaluation, $N_\mathrm{samples}=200\,000$ points are uniformly sampled on both meshes yielding the point sets
$\mathcal{G}=\{\boldsymbol{x}_i\sim \mathcal{U}(\tensor*[^{\mathrm{w}}]{\mathcal{M}}{^{\ast}_{\mathrm{gt}}})\mid i=1,...,N_\mathrm{samples}\}$ and $\mathcal{E}=\{\boldsymbol{y}_i\sim \mathcal{U}(\tensor*[^{\mathrm{w}}]{\mathcal{M}}{^{\ast}_{\mathrm{est}}})\mid i=1,...,N_\mathrm{samples}\}$, where $\mathcal{U}(\cdot)$ denotes the uniform distribution. The point sets are used to compute accuracy, completion, accuracy ratio, and completion ratio as
\begin{align}
    \mathrm{Acc}(\mathcal{G},\mathcal{E})&=\frac{1}{|\mathcal{E}|} \sum_{\boldsymbol{y}\in \mathcal{E}}\min_{\boldsymbol{x}\in\mathcal{G}}\lVert \boldsymbol{y} - \boldsymbol{x} \rVert\\ 
    \mathrm{Comp}(\mathcal{G},\mathcal{E})&=\frac{1}{|\mathcal{G}|} \sum_{\boldsymbol{x}\in \mathcal{G}}\min_{\boldsymbol{y}\in\mathcal{E}}\lVert \boldsymbol{x} - \boldsymbol{y} \rVert\\ 
    \mathrm{AR}(\mathcal{G},\mathcal{E})&=\frac{1}{|\mathcal{E}|} \sum_{\boldsymbol{y}\in \mathcal{E}} \left[\min_{\boldsymbol{x}\in\mathcal{G}}\lVert\boldsymbol{y} - \boldsymbol{x} \rVert < \Delta \right]\\ 
    \mathrm{CR}(\mathcal{G},\mathcal{E})&=\frac{1}{|\mathcal{G}|} \sum_{\boldsymbol{x}\in \mathcal{G}}\left[\min_{\boldsymbol{y}\in\mathcal{E}}\lVert \boldsymbol{x} - \boldsymbol{y} \rVert < \Delta \right],
\end{align}
where $\left[\cdot\right]$ denotes the Iverson bracket and $\Delta=5\,\mathrm{cm}$ in our experiments. Since accuracy ratio and completion ratio can be interpreted as precision and recall of the reconstruction, we further use the F1-score 
\begin{equation}
    F_1(\mathcal{G},\mathcal{E})=\frac{2}{\mathrm{AR}(\mathcal{G},\mathcal{E})^{-1} + \mathrm{CR}(\mathcal{G},\mathcal{E})^{-1}}
\end{equation}
to summarize reconstruction performance in one metric.

\section{Additional Results}

\paragraph{Detailed Results on Replica and NRGBD} \Cref{tab:replicaresults} and \Cref{tab:nrgbdresults} shows per-scene results on the Replica and NRGBD dataset, respectively. Overall, Co-SLAM achieves the best result, albeit with small margins; it fails on \texttt{room1} leading to worse average results. Loopy-SLAM achieves near perfect accuracy at the cost of worse completion. As noted in the paper it uses the ground-truth depth for rendering and mesh extraction and is hence not fairly comparable to the other methods. Overall, this experiment highlights that our contributions aiming for improved loop closure integration do not significantly worsen performance on small scenes.s

\paragraph{Additional Qualitative Results} In \cref{fig:qualitativereplica,fig:qualitativereplicab,fig:qualitativenrgbd,fig:qualitativenrgbdb,fig:qualitativescannet} additional qualitative results on the Replica, NRGBD, and ScanNet datasets are shown. In most scenes our approach performs close to the best performing method Co-SLAM. Co-SLAM achieves slightly more detailed and smoother results, which might be due to their more effective hash table use and the additional smoothness loss. Loopy-SLAM sometimes produces meshes with wrong colors; MIPSFusion shows transition and streaking artifacts; and GO-SLAM produces more noisy results particularly in terms of appearance.

Our method performs most consistent compared to other methods that support loop closure.

\paragraph{K-Nearest Neighbor Averaging} \Cref{fig:knn} shows renderings of our model with varying values for $k$ (see \cref{sec:knn}). For higher $k$ visible transitions between fields are reduced and smoothed out. Note that due to our independent training scheme, all fields are trained in overlapping regions making averaging at the query point a viable strategy. While higher values for $k$ lead to improved results, it also multiplies the number of queries required for rendering and mesh extraction (note that optimization time is unaffected by $k$). %

\begin{figure}[tb!]
    \centering
    \input{figures/knn.tikz}
    \caption{Comparison of rendered views with different number of neighbors $k$. By increasing the number of nearest neighbors taken into account, the number of visible field transitions decreases. These transitions are most visible in unobserved regions, such as the underside of the countertop shown here (see \textcolor{orange}{highlighted} regions).}
    \label{fig:knn}
    \vspace*{-\baselineskip}
\end{figure}

\begin{table*}[p!]
    \centering
    \caption{Comparison of mesh quality on Replica (\textbf{best} \protect\gold, second best \protect\silver, third best \protect\bronze).}\label{tab:replicaresults}
    
    \renewcommand{\arraystretch}{1.1}
    \setlength{\tabcolsep}{3pt}
    \scriptsize
    \begin{tabular}{
      @{}
      ll
      S[table-format=2.2]@{\hspace{0.3\tabcolsep}}c
      S[table-format=2.2]@{\hspace{0.3\tabcolsep}}c
      S[table-format=2.2]@{\hspace{0.3\tabcolsep}}c
      S[table-format=2.2]@{\hspace{0.3\tabcolsep}}c
      S[table-format=2.2]@{\hspace{0.3\tabcolsep}}c
      S[table-format=2.2]@{\hspace{0.3\tabcolsep}}c
      S[table-format=2.2]@{\hspace{0.3\tabcolsep}}c
      S[table-format=2.2]@{\hspace{0.3\tabcolsep}}c
      S[table-format=2.2]@{\hspace{0.3\tabcolsep}}c
      @{}
      }
        \toprule
        & & \texttt{offi0} & & \texttt{offi1} & & \texttt{offi2} & & \texttt{offi3} & & \texttt{offi4} & & \texttt{room0} & & \texttt{room1} & & \texttt{room2} & & \text{Avg.} & \\
        \midrule
        \multirow{4}{*}{NICE-SLAM \cite{zhu2022nice}} 
            & Acc ($\mathrm{cm}$)   & 1.90 &  & 1.61 & \silver & 3.13 &  & 2.92 & \bronze & 2.60 &  & 2.47 &  & 2.21 & \silver & 2.17 & \silver & 2.38 & \bronze \\
            & Acc.-Ratio (\%)       & 94.87 &  & 95.30 &  & 89.78 &  & 90.17 & \bronze & \bfseries 93.21 & \gold & 93.38 &  & 94.92 & \silver & \bfseries 93.75 & \gold & \bfseries 93.17 & \gold \\
            & Comp. ($\mathrm{cm}$) & 2.37 &  & 2.15 &  & 2.89 &  & 3.42 &  & 3.91 &  & 2.93 &  & 2.31 &  & 2.77 & \bronze & 2.84 &  \\
            & Comp. Ratio (\%)      & 92.58 &  & 92.15 &  & 88.78 &  & 86.20 &  & 85.72 &  & 90.90 &  & \bfseries 93.57 & \gold & 90.97 & \silver & 90.11 &  \\
            & F1-Score (\%)         & 93.71 &  & 93.70 & \bronze & 89.28 &  & 88.14 &  & 89.31 &  & 92.12 &  & \bfseries 94.24 & \gold & 92.34 & \silver & 91.60 &  \\
        \midrule
        \multirow{4}{*}{Co-SLAM \cite{wang2023co}} 
            & Acc ($\mathrm{cm}$)   & \bfseries 1.55 & \gold & \bfseries 1.33 & \gold & 2.76 & \bronze & \bfseries 2.61 & \gold & \bfseries 2.22 & \gold & \bfseries 1.99 & \gold & 19.90 &  & \bfseries 1.92 & \gold & 4.28 &  \\
            & Acc.-Ratio (\%)       & 96.15 &  & 96.75 & \silver & 90.92 &  & \bfseries 92.04 & \gold & 92.70 &  & \bfseries 95.37 & \gold & 38.68 &  & 93.51 & \silver & 87.02 &  \\
            & Comp. ($\mathrm{cm}$) & \bfseries 1.54 & \gold & \bfseries 1.68 & \gold & \bfseries 2.39 & \gold & \bfseries 2.73 & \gold & \bfseries 2.47 & \gold & \bfseries 2.37 & \gold & 17.47 &  & \bfseries 2.08 & \gold & 4.09 &  \\
            & Comp. Ratio (\%)      & \bfseries 96.04 & \gold & \bfseries 94.57 & \gold & \bfseries 91.99 & \gold & \bfseries 90.92 & \gold & \bfseries 90.96 & \gold & \bfseries 93.43 & \gold & 40.03 &  & \bfseries 93.16 & \gold & 86.39 &  \\
            & F1-Score (\%)         & \bfseries 96.09 & \gold & \bfseries 95.65 & \gold & 91.46 & \silver & \bfseries 91.48 & \gold & \bfseries 91.82 & \gold & \bfseries 94.39 & \gold & 39.34 &  & \bfseries 93.34 & \gold & 86.70 &  \\
        \midrule
        \multirow{4}{*}{GO-SLAM \cite{zhang2023go}} 
            & Acc ($\mathrm{cm}$)   & 1.88 &  & 1.72 &  & 3.51 &  & 3.97 &  & 3.56 &  & 3.45 &  & \bfseries 2.15 & \gold & 2.90 &  & 2.89 &  \\
            & Acc.-Ratio (\%)       & \bfseries 96.31 & \gold & \bfseries 97.55 & \gold & 84.49 &  & 79.64 &  & 86.34 &  & 84.59 &  & \bfseries 96.04 & \gold & 89.50 &  & 89.31 &  \\
            & Comp. ($\mathrm{cm}$) & 3.12 &  & 4.17 &  & 6.74 &  & 7.43 &  & 8.39 &  & 6.83 &  & 4.26 &  & 8.46 &  & 6.17 &  \\
            & Comp. Ratio (\%)      & 85.21 &  & 82.89 &  & 69.40 &  & 63.81 &  & 67.33 &  & 69.44 &  & 82.89 &  & 72.65 &  & 74.20 &  \\
            & F1-Score (\%)         & 90.42 &  & 89.63 &  & 76.20 &  & 70.85 &  & 75.66 &  & 76.27 &  & 88.98 &  & 80.20 &  & 81.03 &  \\
        \midrule
        \multirow{4}{*}{MIPS-Fusion \cite{tang2023mips}} 
            & Acc ($\mathrm{cm}$)   & 1.65 &  & 1.61 & \bronze & 3.16 &  & 3.43 &  & 2.62 &  & 2.30 &  & 2.64 &  & 2.52 & \bronze & 2.49 &  \\
            & Acc.-Ratio (\%)       & 96.16 & \bronze & 95.41 & \bronze & 91.21 & \bronze & 88.96 &  & 89.99 &  & 94.95 & \silver & 93.76 &  & 90.60 & \bronze & 92.63 &  \\
            & Comp. ($\mathrm{cm}$) & 1.78 &  & 1.86 & \silver & 2.99 &  & 3.22 &  & 2.67 & \silver & 2.78 &  & \bfseries 2.13 & \gold & 2.69 & \silver & 2.52 & \silver \\
            & Comp. Ratio (\%)      & 94.96 &  & 93.09 & \silver & 89.16 &  & 88.40 &  & 89.64 & \silver & 92.37 & \silver & 93.48 & \silver & 89.92 & \bronze & \bfseries 91.38 & \gold \\
            & F1-Score (\%)         & 95.56 & \bronze & 94.24 & \silver & 90.17 &  & 88.68 &  & 89.82 &  & 93.64 & \silver & 93.62 & \bronze & 90.26 & \bronze & 92.00 & \silver \\
        \midrule
        \multirow{4}{*}{Loopy-SLAM \cite{liso2024loopy}} 
            & Acc ($\mathrm{cm}$)   &  1.05 &  &  0.84 &  &  1.33 &  &  1.53 &  &  1.49 &  &  1.45 &  &  1.13 &  &  1.20 &  &  1.25 &  \\
            & Acc.-Ratio (\%)       &  100.00 &  &  100.00 &  &  100.00 &  &  99.98 &  &  99.99 &  &  99.99 &  &  100.00 &  &  99.99 &  &  99.99 &  \\
            & Comp. ($\mathrm{cm}$) & 1.65 &  & 2.16 &  & 3.64 &  & 3.03 &  & 3.80 &  & 3.45 &  & 2.95 &  & 2.65 &  & 2.92 &  \\
            & Comp. Ratio (\%)      & 93.43 &  & 90.80 &  & 86.33 &  & 88.67 &  & 86.32 &  & 88.58 &  & 89.78 &  & 90.32 &  & 89.28 &  \\
            & F1-Score (\%)         &  96.61 &  & 95.18 &  &  92.66 &  &  93.98 &  &  92.65 &  & 93.94 &  &  94.62 &  &  94.91 &  &  94.32 &  \\
        \midrule
        \multirow{4}{*}{Ours-SF} 
            & Acc ($\mathrm{cm}$)   & 1.55 & \silver & 1.79 &  & 2.49 & \silver & 3.09 &  & 2.38 & \bronze & 2.12 & \silver & 2.52 & \bronze & 2.53 &  & \bfseries 2.31 & \gold \\
            & Acc.-Ratio (\%)       & 96.25 & \silver & 94.20 &  & 92.51 & \silver & 89.94 &  & 92.83 & \silver & 94.51 & \bronze & 94.59 & \bronze & 90.08 &  & 93.11 & \silver \\
            & Comp. ($\mathrm{cm}$) & 1.59 & \silver & 2.01 &  & 2.66 & \silver & 2.94 & \bronze & 2.89 & \bronze & 2.46 & \silver & 2.23 & \bronze & 3.16 &  & \bfseries 2.49 & \gold \\
            & Comp. Ratio (\%)      & 95.88 & \silver & 92.32 & \bronze & 90.65 & \silver & 89.31 & \silver & 88.24 & \bronze & 91.92 & \bronze & 93.20 &  & 88.75 &  & 91.28 & \silver \\
            & F1-Score (\%)         & 96.07 & \silver & 93.25 &  & \bfseries 91.57 & \gold & 89.63 & \bronze & 90.47 & \silver & 93.20 & \bronze & 93.89 & \silver & 89.41 &  & \bfseries 92.19 & \gold \\
        \midrule
        \multirow{4}{*}{Ours} 
            & Acc ($\mathrm{cm}$)   & 1.60 & \bronze & 1.86 &  & \bfseries 2.46 & \gold & 2.81 & \silver & 2.36 & \silver & 2.14 & \bronze & 2.73 &  & 2.85 &  & 2.35 & \silver \\
            & Acc.-Ratio (\%)       & 95.74 &  & 94.20 &  & \bfseries 92.52 & \gold & 90.53 & \silver & 92.81 & \bronze & 94.46 &  & 92.55 &  & 89.26 &  & 92.76 & \bronze \\
            & Comp. ($\mathrm{cm}$) & 1.67 & \bronze & 2.00 & \bronze & 2.76 & \bronze & 2.78 & \silver & 3.01 &  & 2.71 & \bronze & 2.16 & \silver & 4.00 &  & 2.64 & \bronze \\
            & Comp. Ratio (\%)      & 95.27 & \bronze & 91.74 &  & 89.65 & \bronze & 89.01 & \bronze & 88.02 &  & 91.88 &  & 93.41 & \bronze & 87.96 &  & 90.87 & \bronze \\
            & F1-Score (\%)         & 95.50 &  & 92.95 &  & 91.06 & \bronze & 89.76 & \silver & 90.35 & \bronze & 93.15 &  & 92.98 &  & 88.61 &  & 91.80 & \bronze \\
         \bottomrule
    \end{tabular}
    \vspace*{-\baselineskip}
\end{table*}

\begin{table*}[p!]
    \centering
    \caption{Comparison of mesh quality on NRGBD (\textbf{best} \protect\gold, second best \protect\silver, third best \protect\bronze).}\label{tab:nrgbdresults}
    
    \renewcommand{\arraystretch}{1.1}
    \setlength{\tabcolsep}{3pt}
    \scriptsize
    \begin{tabular}{
      @{}
      ll
      S[table-format=2.2]@{\hspace{0.3\tabcolsep}}c
      S[table-format=2.2]@{\hspace{0.3\tabcolsep}}c
      S[table-format=2.2]@{\hspace{0.3\tabcolsep}}c
      S[table-format=2.2]@{\hspace{0.3\tabcolsep}}c
      S[table-format=2.2]@{\hspace{0.3\tabcolsep}}c
      S[table-format=2.2]@{\hspace{0.3\tabcolsep}}c
      S[table-format=2.2]@{\hspace{0.3\tabcolsep}}c
      S[table-format=2.2]@{\hspace{0.3\tabcolsep}}c
      S[table-format=2.2]@{\hspace{0.3\tabcolsep}}c
      S[table-format=2.2]@{\hspace{0.3\tabcolsep}}c
      @{}
      }
        \toprule
        & & \texttt{br} & & \texttt{ck} & & \texttt{gr} & & \texttt{gwr} & & \texttt{ki} & & \texttt{ma} & & \texttt{sc} & & \texttt{tg} & & \texttt{wa} & & \text{Avg.} & \\
        \midrule
        \multirow{4}{*}{NICE-SLAM \cite{zhu2022nice}} 
            & Acc ($\mathrm{cm}$)   & 2.46 &  & 10.76 &  & 2.33 &  & 2.71 &  & 9.18 &  & 1.70 & \silver & 4.55 &  & 8.37 &  & 7.37 &  & 5.49 &  \\
            & Acc.-Ratio (\%)       & 92.41 &  & 65.34 &  & 93.87 &  & 93.63 &  & 57.29 &  & 95.06 & \silver & 71.69 &  & 56.22 &  & 77.38 &  & 78.10 &  \\
            & Comp. ($\mathrm{cm}$) & 4.82 &  & 14.21 &  & 3.91 &  & 3.19 &  & 12.82 &  & 3.36 &  & 10.69 &  & 8.02 &  & 5.39 &  & 7.38 &  \\
            & Comp. Ratio (\%)      & 86.18 &  & 53.58 &  & 86.66 &  & 87.64 &  & 50.04 &  & 84.06 & \bronze & 58.79 &  & 59.24 &  & 69.04 &  & 70.58 &  \\
            & F1-Score (\%)         & 89.19 &  & 58.88 &  & 90.12 &  & 90.54 &  & 53.42 &  & 89.22 & \silver & 64.60 &  & 57.69 &  & 72.97 &  & 74.07 &  \\
        \midrule
        \multirow{4}{*}{Co-SLAM \cite{wang2023co}} 
            & Acc ($\mathrm{cm}$)   & 2.21 & \bronze & 4.73 &  & 1.89 & \bronze & 2.02 & \bronze & 7.40 &  & 1.74 &  & 3.30 & \bronze & 2.07 & \bronze & 6.24 &  & 3.51 &  \\
            & Acc.-Ratio (\%)       & 93.24 & \bronze & 75.16 &  & 95.17 & \silver & 94.84 &  & 77.72 &  & 93.98 &  & 78.01 & \bronze & 92.05 & \bronze & 84.92 &  & 87.23 &  \\
            & Comp. ($\mathrm{cm}$) & \bfseries 2.06 & \gold & 8.76 & \silver & 2.93 & \bronze & \bfseries 2.41 & \gold & 5.14 & \silver & \bfseries 2.75 & \gold & 4.29 & \silver & \bfseries 2.83 & \gold & 3.85 & \silver & \bfseries 3.89 & \gold \\
            & Comp. Ratio (\%)      & 93.49 & \bronze & 63.17 & \bronze & \bfseries 91.32 & \gold & \bfseries 93.96 & \gold & \bfseries 78.19 & \gold & \bfseries 86.41 & \gold & 70.90 & \bronze & \bfseries 86.62 & \gold & 81.70 & \silver & 82.86 & \bronze \\
            & F1-Score (\%)         & 93.37 & \bronze & 68.65 & \bronze & \bfseries 93.21 & \gold & \bfseries 94.40 & \gold & 77.96 &  & \bfseries 90.03 & \gold & 74.29 & \bronze & \bfseries 89.26 & \gold & 83.28 & \bronze & 84.94 & \bronze \\
        \midrule
        \multirow{4}{*}{GO-SLAM \cite{zhang2023go}} 
            & Acc ($\mathrm{cm}$)   & 3.89 &  & 4.08 &  & 2.50 &  & 2.87 &  & 3.28 & \bronze & \bfseries 1.54 & \gold & 6.46 &  & \bfseries 1.48 & \gold & 5.46 &  & 3.51 &  \\
            & Acc.-Ratio (\%)       & 77.64 &  & 81.94 &  & 91.46 &  & 86.87 &  & 84.74 & \bronze & \bfseries 97.44 & \gold & 66.62 &  & \bfseries 96.32 & \gold & 73.90 &  & 84.10 &  \\
            & Comp. ($\mathrm{cm}$) & 9.25 &  & 29.60 &  & 9.50 &  & 4.50 &  & \bfseries 5.11 & \gold & 4.60 &  & 12.35 &  & 7.26 &  & 12.57 &  & 10.53 &  \\
            & Comp. Ratio (\%)      & 64.29 &  & 54.56 &  & 71.31 &  & 75.12 &  & 72.58 &  & 75.53 &  & 54.24 &  & 70.78 &  & 57.33 &  & 66.19 &  \\
            & F1-Score (\%)         & 70.34 &  & 65.51 &  & 80.14 &  & 80.57 &  & 78.19 & \bronze & 85.10 &  & 59.80 &  & 81.59 &  & 64.57 &  & 73.98 &  \\
        \midrule
        \multirow{4}{*}{MIPS-Fusion \cite{tang2023mips}} 
            & Acc ($\mathrm{cm}$)   & 2.44 &  & 3.48 & \silver & 1.94 &  & 2.10 &  & 9.24 &  & 1.72 & \bronze & 3.49 &  & 1.61 & \silver & 3.30 & \bronze & 3.26 & \bronze \\
            & Acc.-Ratio (\%)       & 91.20 &  & 84.69 & \bronze & 94.53 &  & 95.76 & \bronze & 75.62 &  & 93.68 &  & 75.92 &  & 93.59 & \silver & 89.61 & \bronze & 88.29 & \bronze \\
            & Comp. ($\mathrm{cm}$) & 5.05 &  & 35.03 &  & 8.33 &  & 3.07 &  & 8.91 & \bronze & 3.15 & \bronze & 8.63 &  & 3.61 & \bronze & 11.52 &  & 9.70 &  \\
            & Comp. Ratio (\%)      & 83.49 &  & 55.36 &  & 83.09 &  & 89.38 &  & 70.86 &  & 82.81 &  & 64.98 &  & 79.77 &  & 72.89 &  & 75.85 &  \\
            & F1-Score (\%)         & 87.18 &  & 66.95 &  & 88.44 &  & 92.46 &  & 73.16 &  & 87.91 &  & 70.02 &  & 86.13 & \silver & 80.39 &  & 81.40 &  \\
        \midrule
        \multirow{4}{*}{Loopy-SLAM \cite{liso2024loopy}} 
            & Acc ($\mathrm{cm}$)   & 1.44 & & 3.41 & & 1.93 &  & 2.00 &  & 20.41 &  &  1.14 &  & 2.59 &  & \text{\protect\xmark} &  &  3.02 &  & 4.49 &  \\
            & Acc.-Ratio (\%)       & 99.85 & & 85.01 & &  98.99 &  &  98.52 &  & 61.61 &  &  99.93 &  & 94.86 &  & \text{\protect\xmark} &  & 85.87 &  & 90.58 &  \\
            & Comp. ($\mathrm{cm}$) & 3.24 &  & 13.57 &  & 3.41 &  & 3.14 &  & 8.31 &  & 3.17 &  &  3.39 &  & \text{\protect\xmark} &  & 4.40 &  & 5.33 &  \\
            & Comp. Ratio (\%)      & 90.29 &  & 71.09 &  & 88.58 &  & 89.86 &  & 68.23 &  & 83.98 &  & 85.03 &  & \text{\protect\xmark} &  & 76.88 &  & 81.74 &  \\
            & F1-Score (\%)         &  94.83 &  & 77.43 &  &  93.50 &  & 93.99 &  & 64.75 &  &  91.26 &  & 89.67 &  & \text{\protect\xmark} &  & 81.13 &  & 90.58 &  \\
        \midrule
        \multirow{4}{*}{Ours-SF} 
            & Acc ($\mathrm{cm}$)   & \bfseries 1.68 & \gold & 3.57 & \bronze & 1.82 & \silver & 1.95 & \silver & 2.28 & \silver & 1.79 &  & \bfseries 2.33 & \gold & 2.55 &  & 3.10 & \silver & 2.34 & \silver \\
            & Acc.-Ratio (\%)       & \bfseries 94.57 & \gold & 86.96 & \silver & 95.12 & \bronze & \bfseries 96.34 & \gold & 95.24 & \silver & 93.90 &  & \bfseries 95.66 & \gold & 89.75 &  & 90.47 & \silver & \bfseries 93.11 & \gold \\
            & Comp. ($\mathrm{cm}$) & 2.16 & \silver & \bfseries 7.49 & \gold & \bfseries 2.86 & \gold & 2.64 & \silver & 25.86 &  & 3.10 & \silver & \bfseries 3.84 & \gold & 3.79 &  & \bfseries 3.73 & \gold & 6.16 & \silver \\
            & Comp. Ratio (\%)      & \bfseries 94.87 & \gold & \bfseries 78.60 & \gold & 90.95 & \silver & 91.76 & \silver & 74.19 & \silver & 84.16 & \silver & \bfseries 88.77 & \gold & 82.27 & \bronze & \bfseries 83.45 & \gold & \bfseries 85.45 & \gold \\
            & F1-Score (\%)         & \bfseries 94.72 & \gold & \bfseries 82.57 & \gold & 92.99 & \bronze & 93.99 & \silver & \bfseries 83.41 & \gold & 88.76 & \bronze & \bfseries 92.09 & \gold & 85.85 & \bronze & \bfseries 86.82 & \gold & \bfseries 89.02 & \gold \\
        \midrule
        \multirow{4}{*}{Ours} 
            & Acc ($\mathrm{cm}$)   & 1.77 & \silver & \bfseries 3.19 & \gold & \bfseries 1.78 & \gold & \bfseries 1.94 & \gold & \bfseries 2.18 & \gold & 1.78 &  & 2.42 & \silver & 2.65 &  & \bfseries 2.92 & \gold & \bfseries 2.29 & \gold \\
            & Acc.-Ratio (\%)       & 93.83 & \silver & \bfseries 88.67 & \gold & \bfseries 95.50 & \gold & 96.30 & \silver & \bfseries 96.00 & \gold & 94.24 & \bronze & 94.95 & \silver & 87.40 &  & \bfseries 90.90 & \gold & 93.09 & \silver \\
            & Comp. ($\mathrm{cm}$) & 2.46 & \bronze & 8.98 & \bronze & 2.88 & \silver & 2.67 & \bronze & 29.26 &  & 3.25 &  & 4.33 & \bronze & 3.33 & \silver & 4.18 & \bronze & 6.82 & \bronze \\
            & Comp. Ratio (\%)      & 93.84 & \silver & 76.89 & \silver & 90.83 & \bronze & 91.31 & \bronze & 73.69 & \bronze & 82.80 &  & 87.90 & \silver & 83.53 & \silver & 81.51 & \bronze & 84.70 & \silver \\
            & F1-Score (\%)         & 93.83 & \silver & 82.36 & \silver & 93.11 & \silver & 93.74 & \bronze & 83.38 & \silver & 88.15 &  & 91.29 & \silver & 85.42 &  & 85.95 & \silver & 88.58 & \silver \\
        \bottomrule
    \end{tabular}
    \vspace*{-\baselineskip}
\end{table*}

\begin{figure*}[p!]
    \centering
    \scriptsize
    \setlength{\fboxsep}{0pt}
    \setlength{\tabcolsep}{1pt}
    \renewcommand{\arraystretch}{0.8}
    \begin{tabular}{C{1em}C{0.24\linewidth}C{0.24\linewidth}C{0.24\linewidth}C{0.24\linewidth}C{0.24\linewidth}}
        & \texttt{office0} & \texttt{office1} & \texttt{office2} & \texttt{office3} \\
        \rotatebox{90}{Ground-truth} 
        & \includegraphics[width=\linewidth]{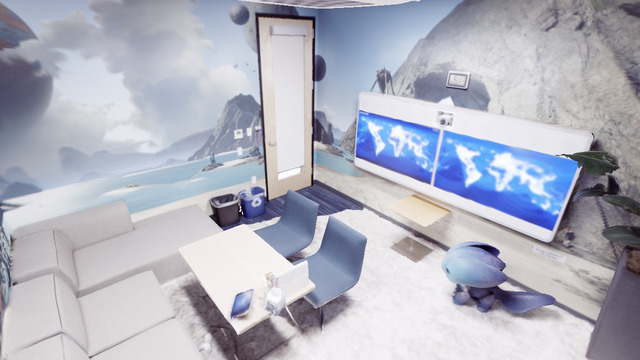} 
        & \includegraphics[width=\linewidth]{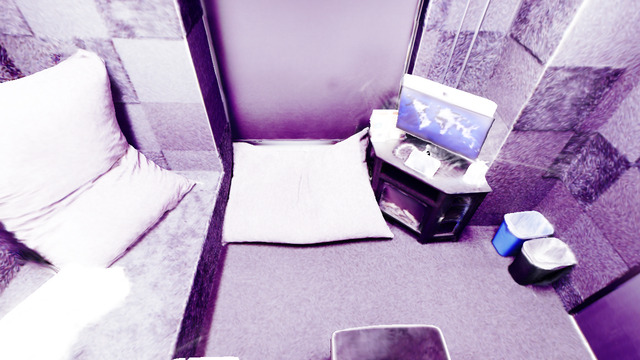}
        & \includegraphics[width=\linewidth]{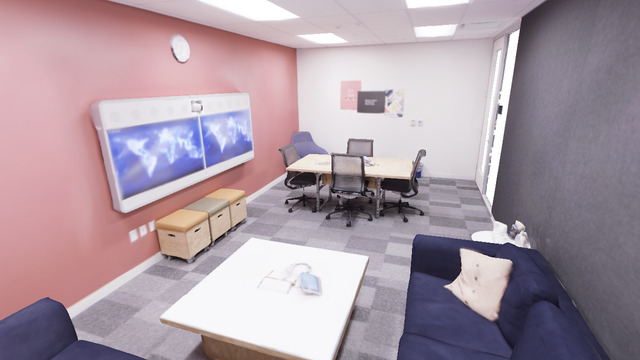}
        & \includegraphics[width=\linewidth]{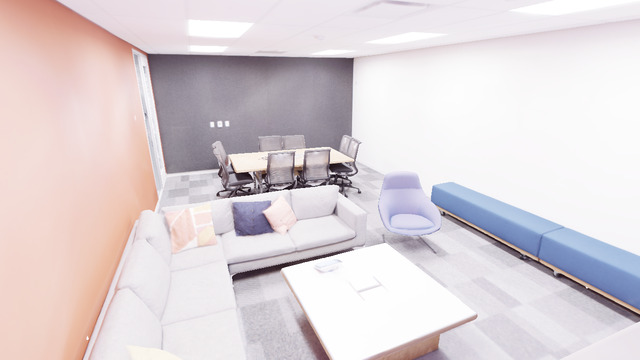} \\
        \rotatebox{90}{NICE-SLAM \cite{zhu2022nice}} 
        & \includegraphics[width=\linewidth]{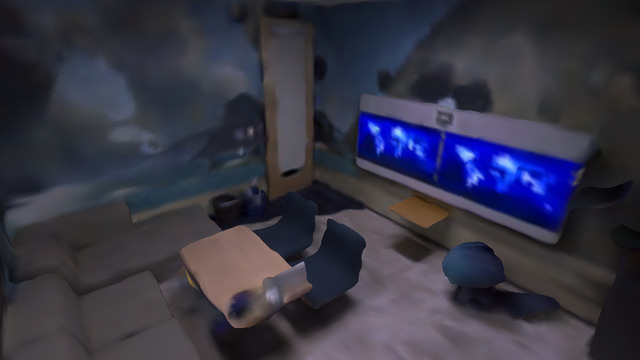}
        & \includegraphics[width=\linewidth]{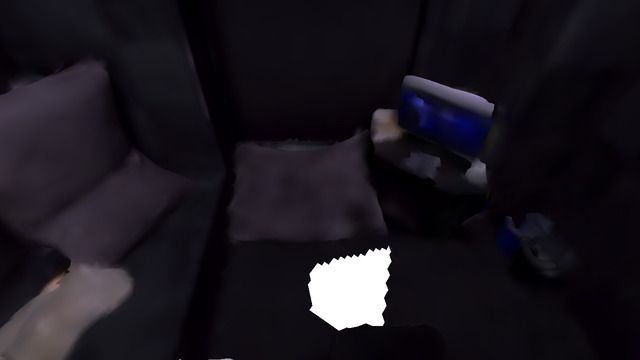}
        & \includegraphics[width=\linewidth]{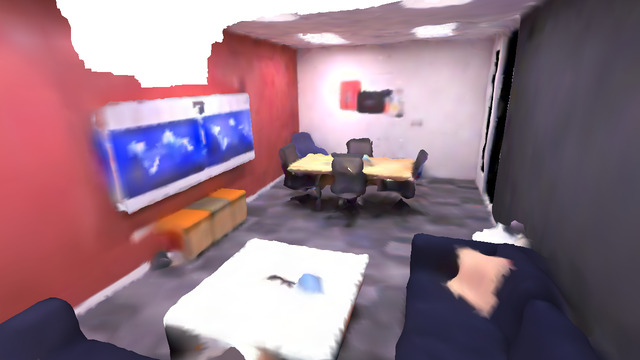}
        & \includegraphics[width=\linewidth]{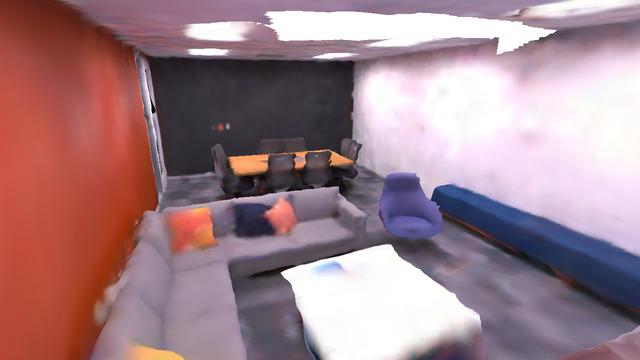} \\
        \rotatebox{90}{Co-SLAM \cite{wang2023co}}
        & \includegraphics[width=\linewidth]{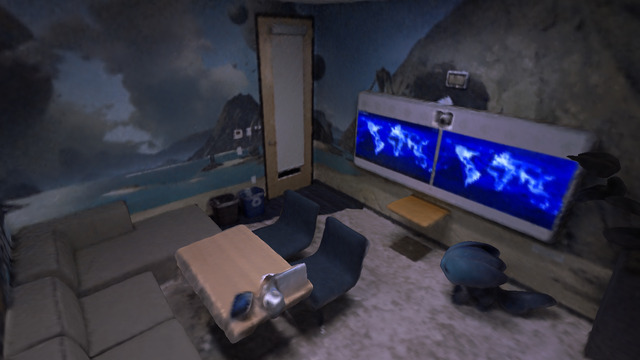}
        & \includegraphics[width=\linewidth]{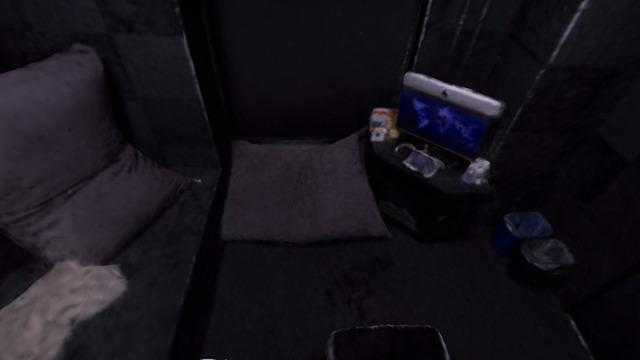}
        & \includegraphics[width=\linewidth]{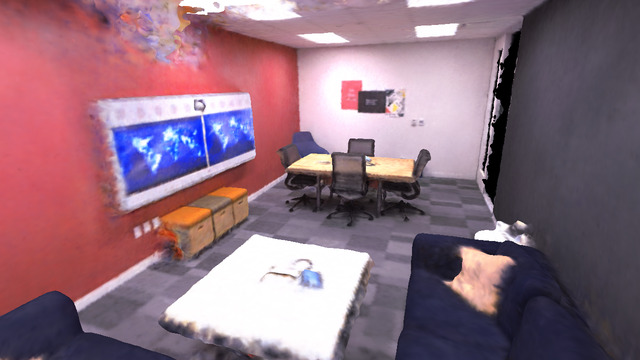}
        & \includegraphics[width=\linewidth]{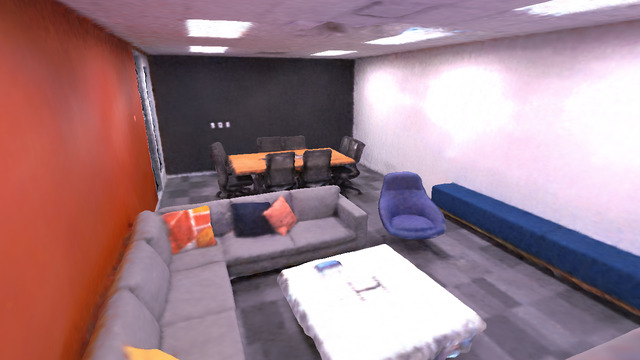} \\
        \rotatebox{90}{GO-SLAM \cite{zhang2023go}}
        & \includegraphics[width=\linewidth]{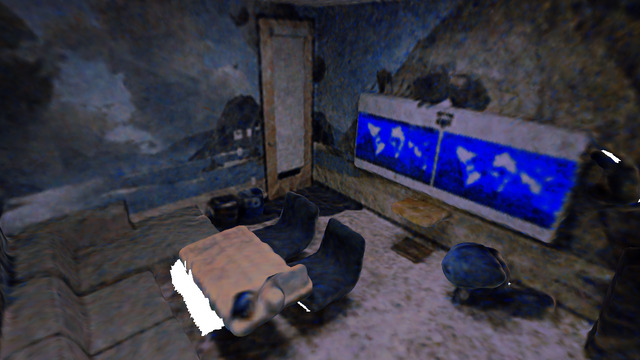}
        & \includegraphics[width=\linewidth]{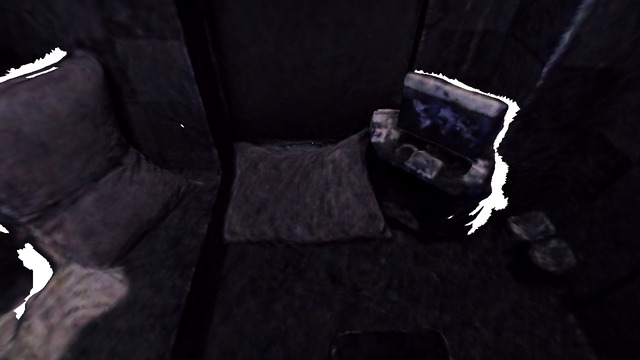}
        & \includegraphics[width=\linewidth]{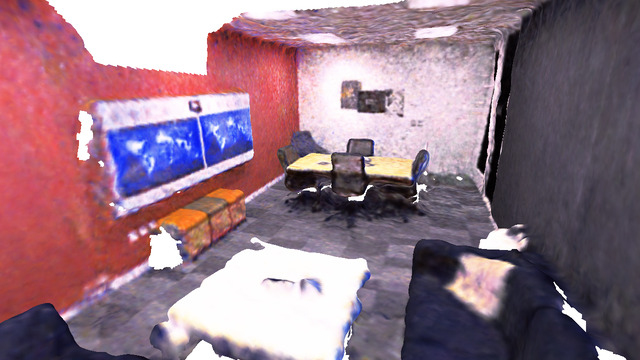}
        & \includegraphics[width=\linewidth]{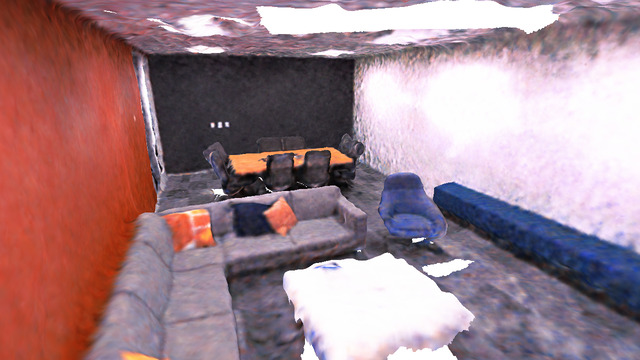} \\
        \rotatebox{90}{MIPS-Fusion \cite{tang2023mips}}
        & \includegraphics[width=\linewidth]{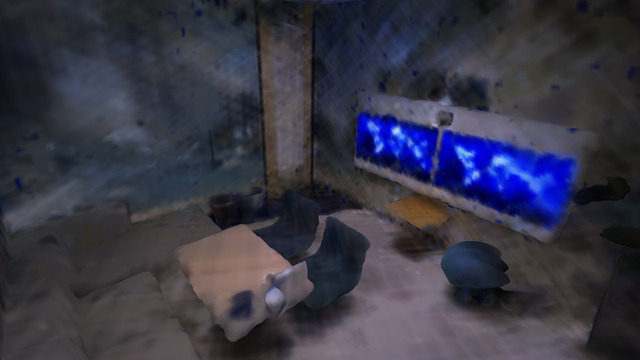}
        & \includegraphics[width=\linewidth]{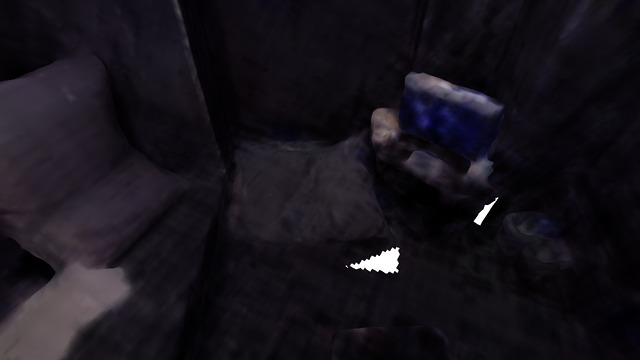}
        & \includegraphics[width=\linewidth]{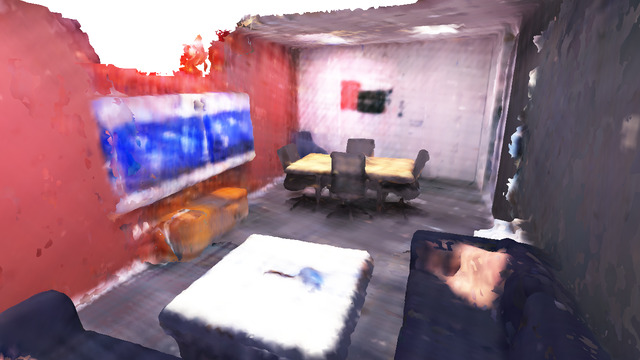}
        & \includegraphics[width=\linewidth]{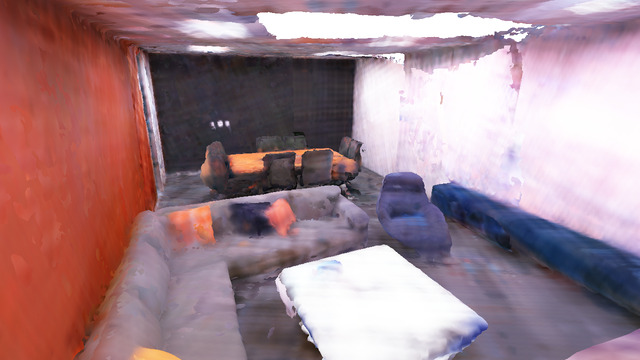} \\
        \rotatebox{90}{Loopy-SLAM \cite{liso2024loopy}}
        & \includegraphics[width=\linewidth]{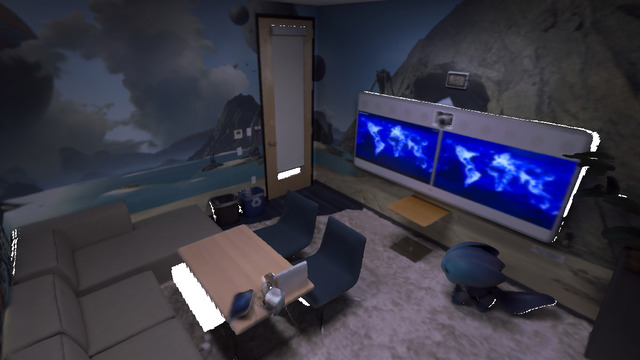}
        & \includegraphics[width=\linewidth]{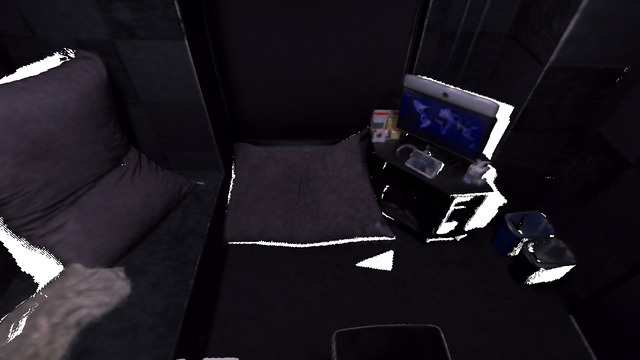}
        & \includegraphics[width=\linewidth]{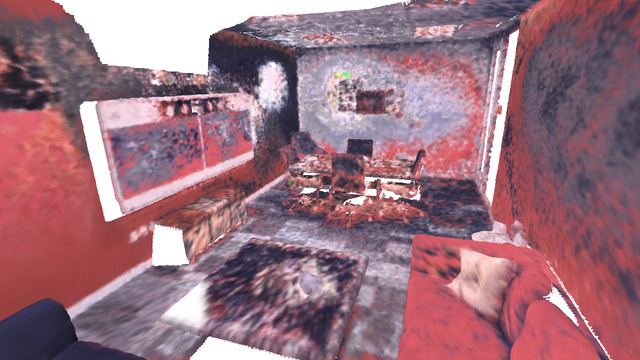}
        & \includegraphics[width=\linewidth]{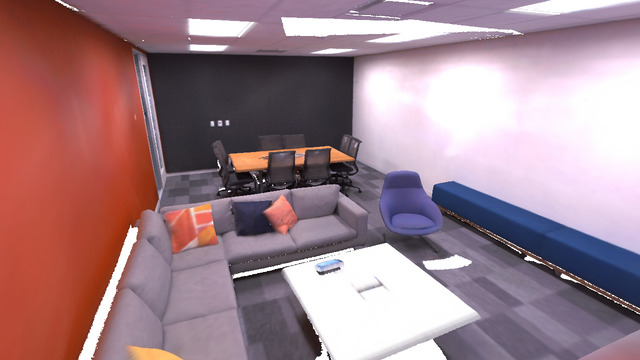} \\
        \rotatebox{90}{Ours-SF}
        & \includegraphics[width=\linewidth]{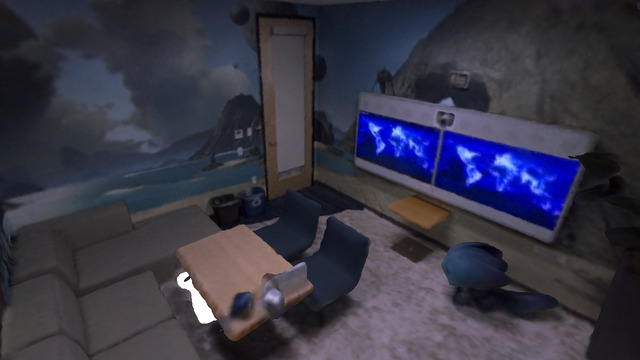}
        & \includegraphics[width=\linewidth]{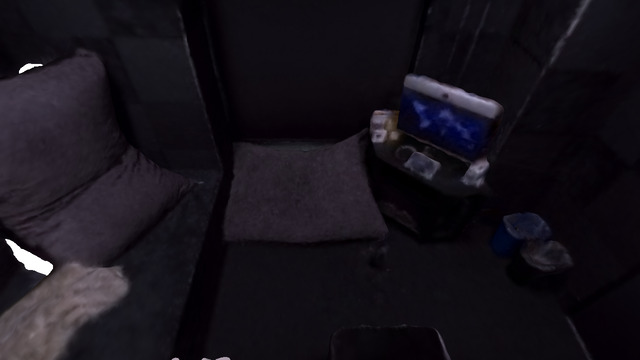}
        & \includegraphics[width=\linewidth]{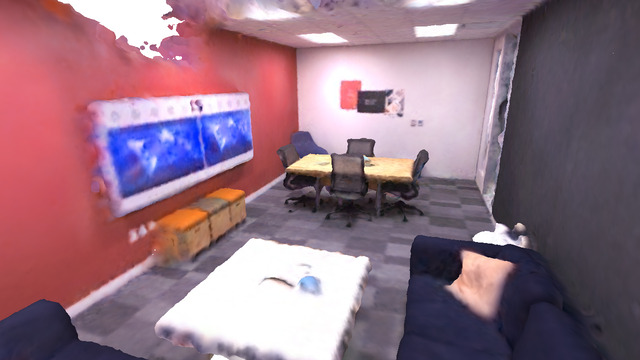}
        & \includegraphics[width=\linewidth]{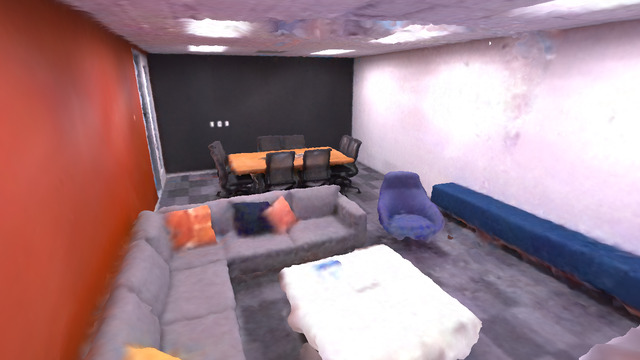} \\
        \rotatebox{90}{Ours}
        & \includegraphics[width=\linewidth]{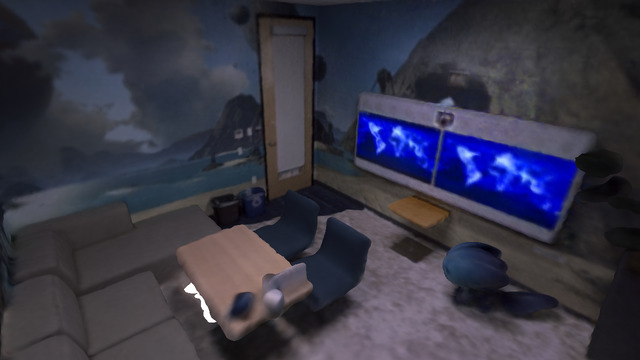}
        & \includegraphics[width=\linewidth]{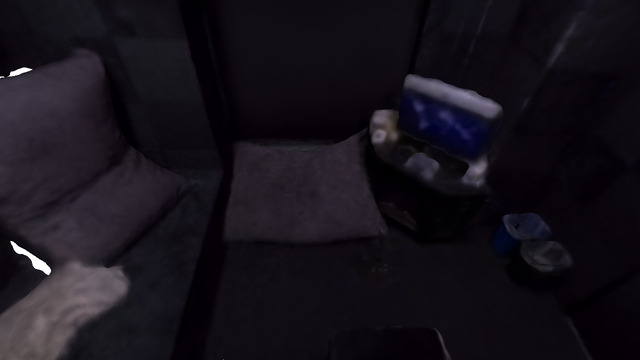}
        & \includegraphics[width=\linewidth]{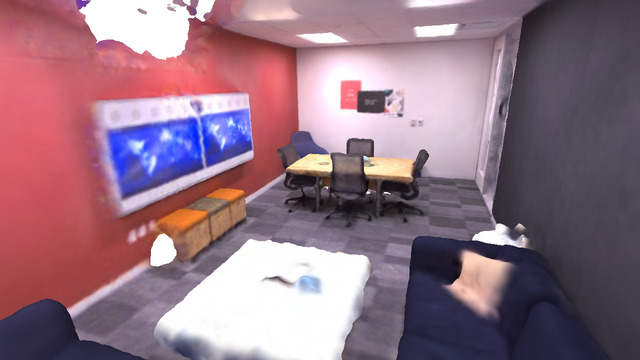}
        & \includegraphics[width=\linewidth]{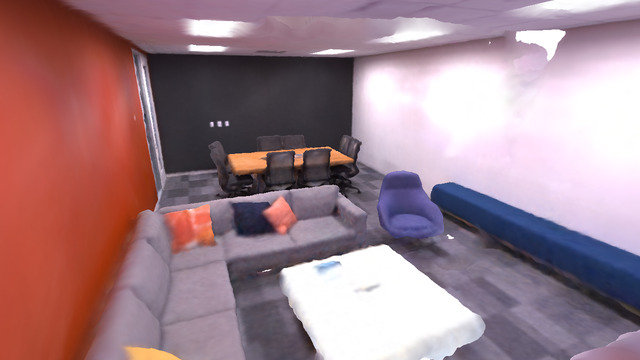}
    \end{tabular}
    \caption{Qualitative comparison of final meshes extracted by all methods on the Replica dataset \cite{replica19arxiv, sucar2021imap} (part 1).}
    \label{fig:qualitativereplica}
\end{figure*}

\begin{figure*}[p!]
    \centering
    \scriptsize
    \setlength{\fboxsep}{0pt}
    \setlength{\tabcolsep}{1pt}
    \renewcommand{\arraystretch}{0.8}
    \begin{tabular}{C{1em}C{0.24\linewidth}C{0.24\linewidth}C{0.24\linewidth}C{0.24\linewidth}C{0.24\linewidth}}
        & \texttt{office4} & \texttt{room0} & \texttt{room1} & \texttt{room2} \\
        \rotatebox{90}{Ground-truth} 
        & \includegraphics[width=\linewidth]{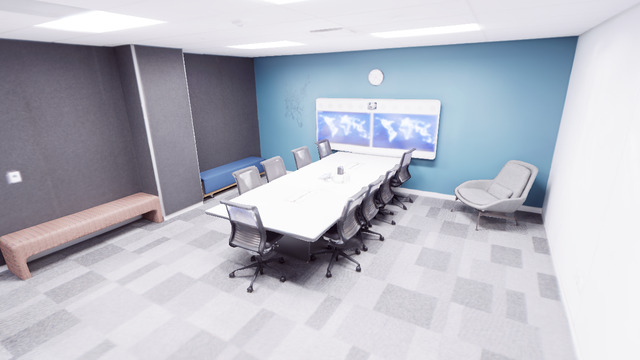} 
        & \includegraphics[width=\linewidth]{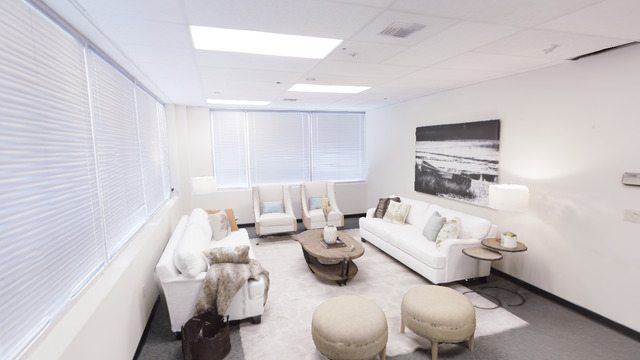}
        & \includegraphics[width=\linewidth]{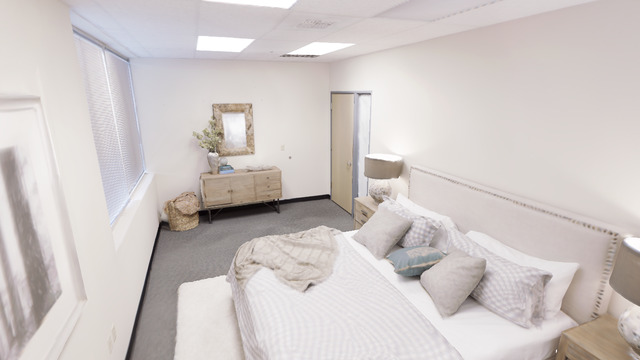}
        & \includegraphics[width=\linewidth]{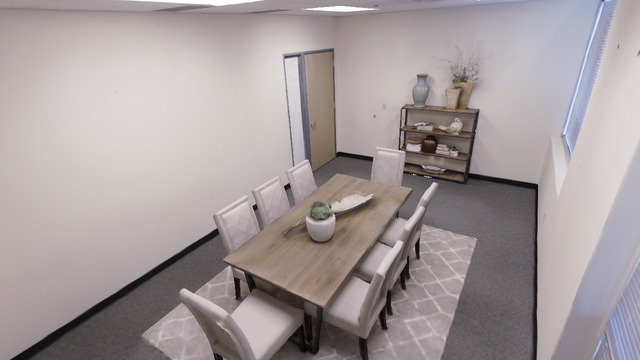} \\
        \rotatebox{90}{NICE-SLAM \cite{zhu2022nice}} 
        & \includegraphics[width=\linewidth]{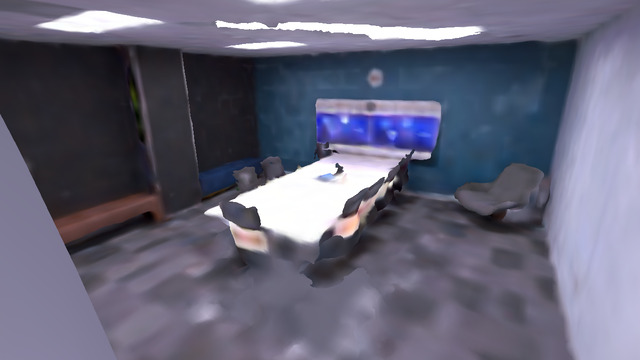}
        & \includegraphics[width=\linewidth]{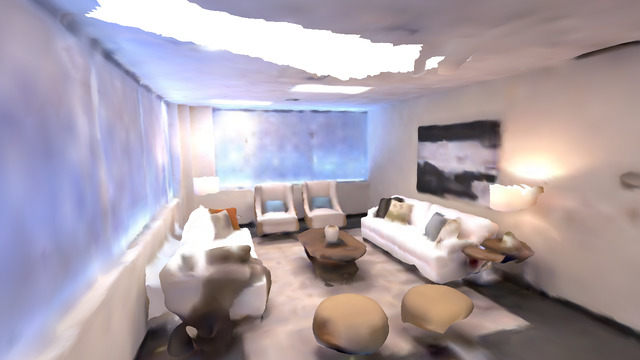}
        & \includegraphics[width=\linewidth]{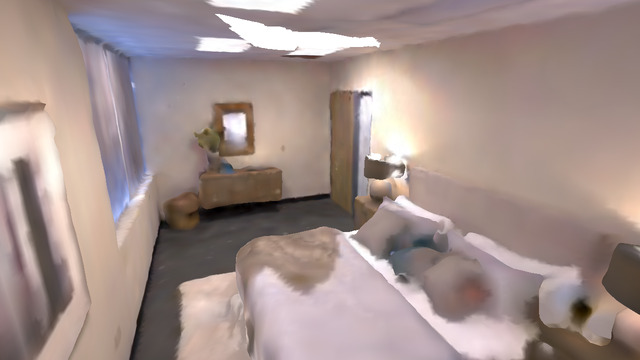}
        & \includegraphics[width=\linewidth]{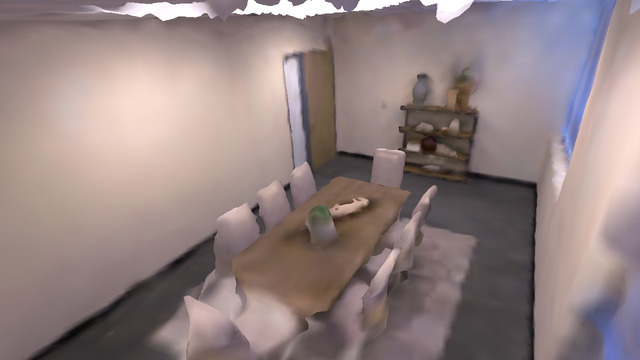} \\
        \rotatebox{90}{Co-SLAM \cite{wang2023co}}
        & \includegraphics[width=\linewidth]{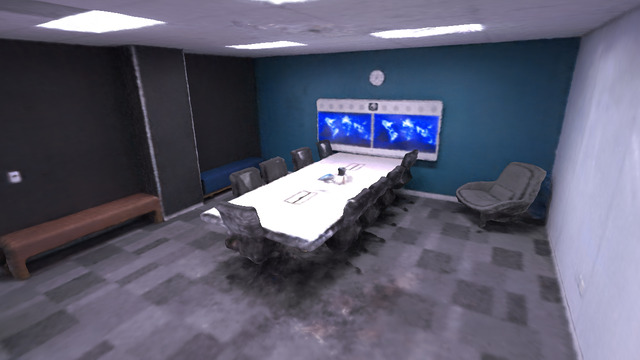}
        & \includegraphics[width=\linewidth]{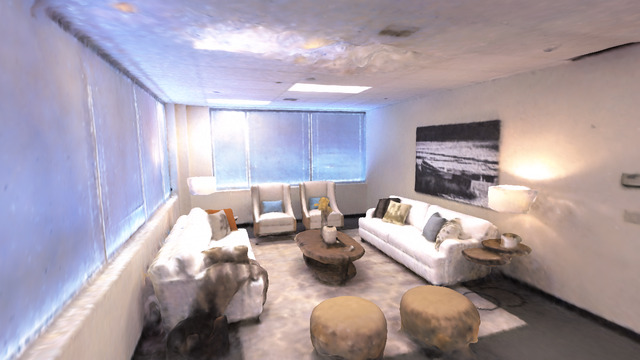}
        & \includegraphics[width=\linewidth]{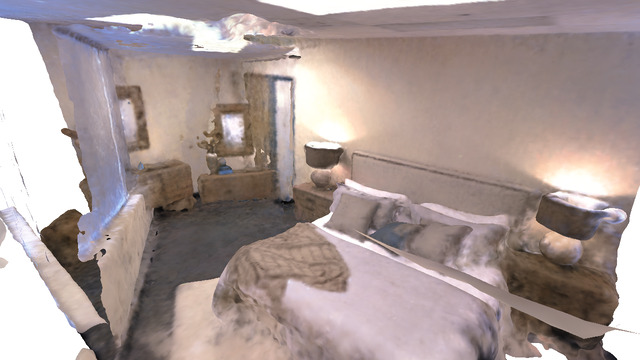}
        & \includegraphics[width=\linewidth]{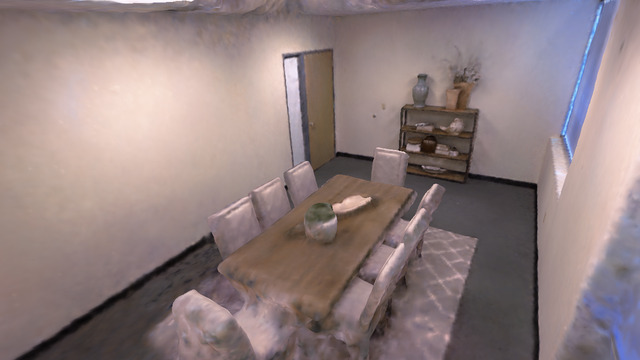} \\
        \rotatebox{90}{GO-SLAM \cite{zhang2023go}}
        & \includegraphics[width=\linewidth]{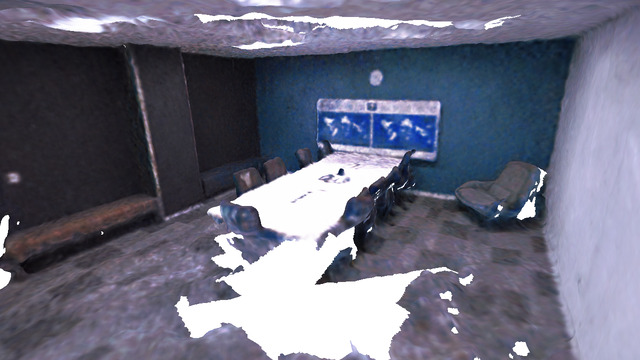}
        & \includegraphics[width=\linewidth]{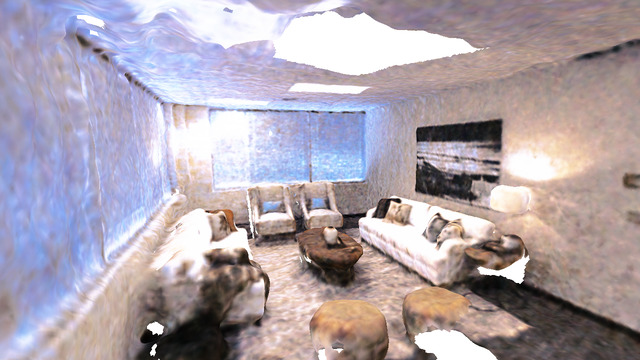}
        & \includegraphics[width=\linewidth]{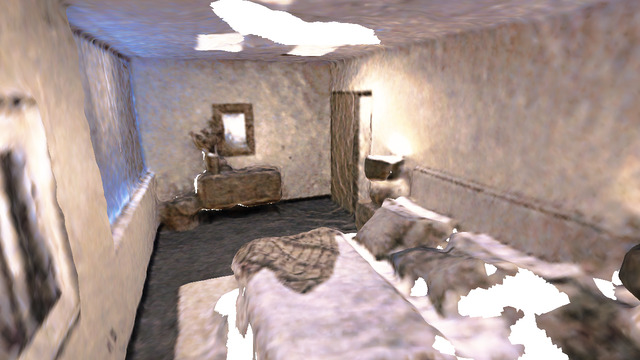}
        & \includegraphics[width=\linewidth]{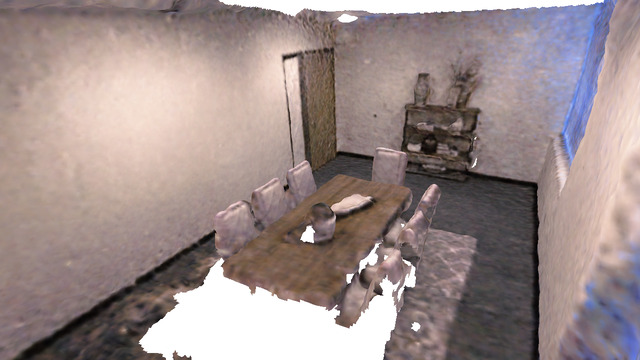} \\
        \rotatebox{90}{MIPS-Fusion \cite{tang2023mips}}
        & \includegraphics[width=\linewidth]{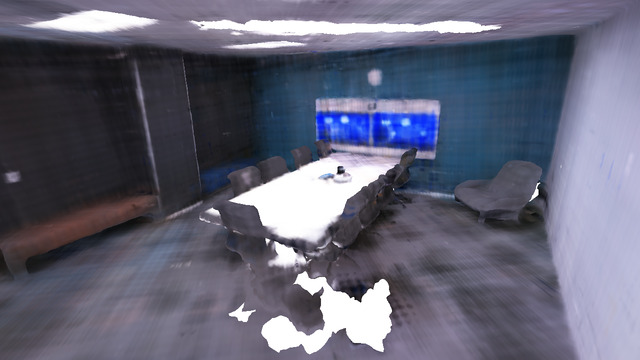}
        & \includegraphics[width=\linewidth]{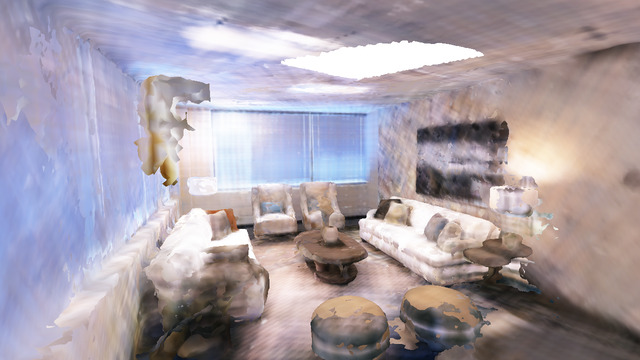}
        & \includegraphics[width=\linewidth]{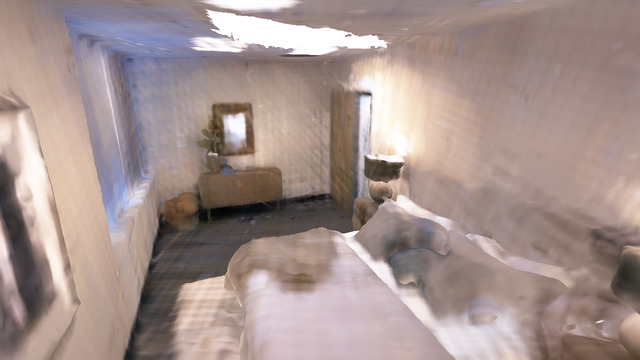}
        & \includegraphics[width=\linewidth]{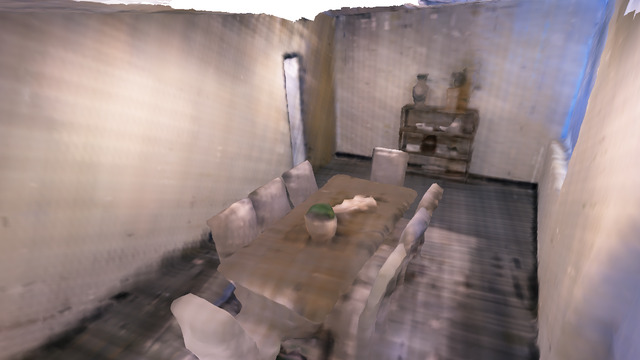} \\
        \rotatebox{90}{Loopy-SLAM \cite{liso2024loopy}}
        & \includegraphics[width=\linewidth]{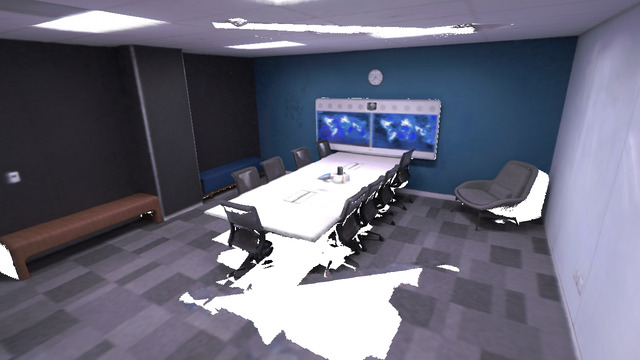}
        & \includegraphics[width=\linewidth]{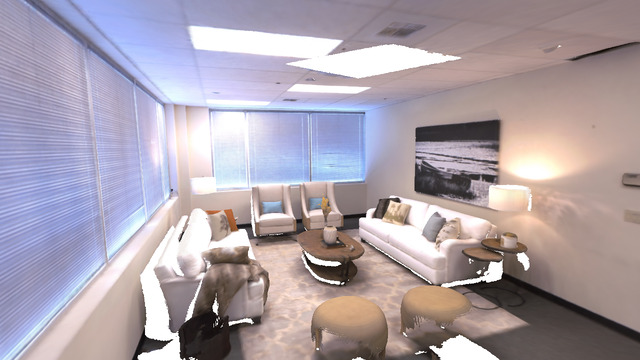}
        & \includegraphics[width=\linewidth]{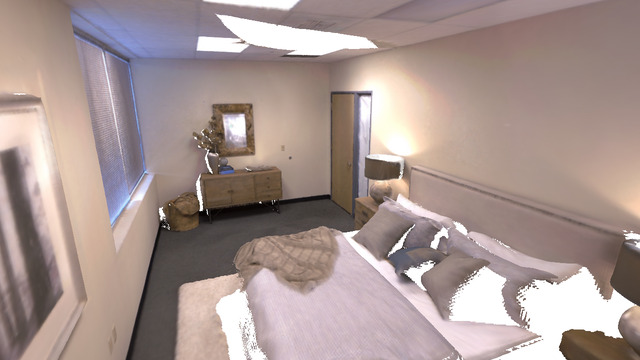}
        & \includegraphics[width=\linewidth]{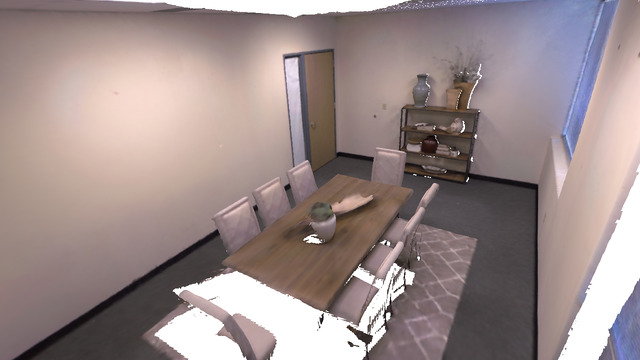} \\
        \rotatebox{90}{Ours-SF}
        & \includegraphics[width=\linewidth]{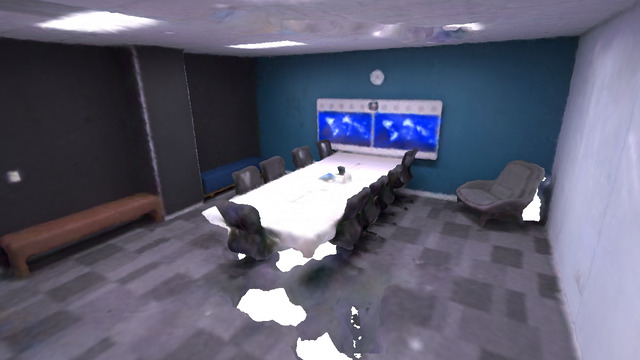}
        & \includegraphics[width=\linewidth]{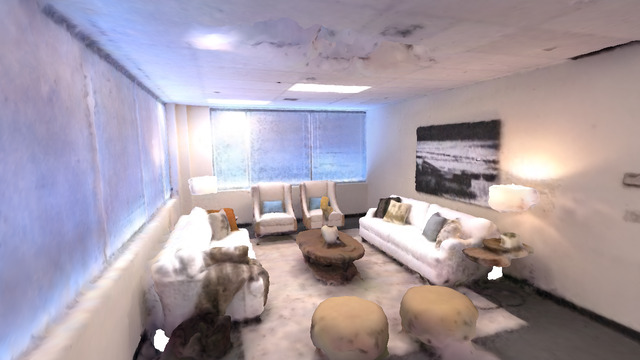}
        & \includegraphics[width=\linewidth]{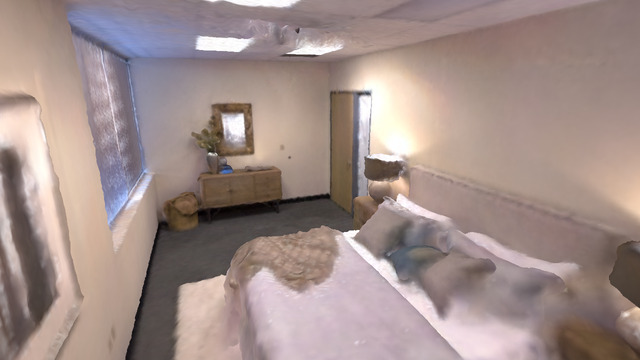}
        & \includegraphics[width=\linewidth]{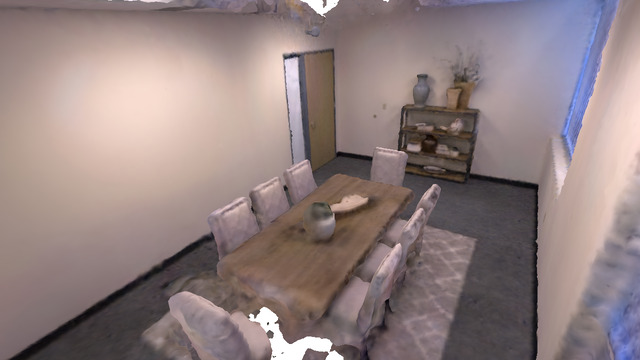} \\
        \rotatebox{90}{Ours}
        & \includegraphics[width=\linewidth]{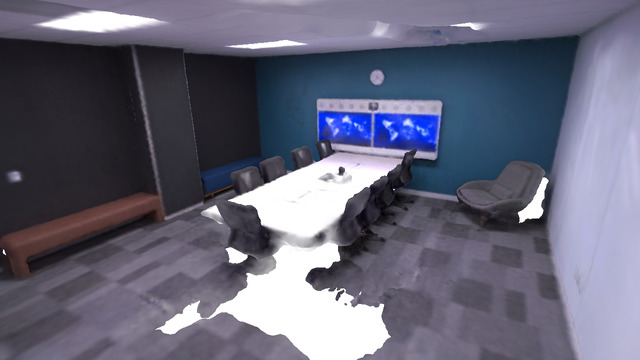}
        & \includegraphics[width=\linewidth]{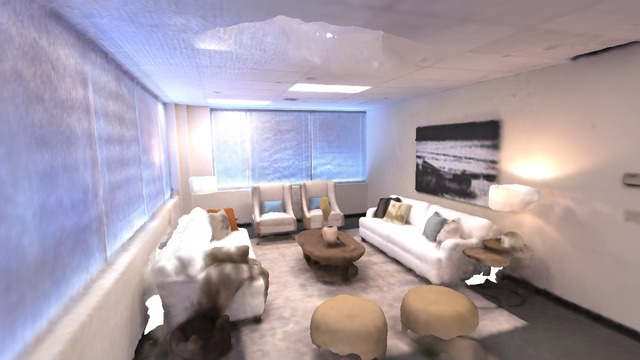}
        & \includegraphics[width=\linewidth]{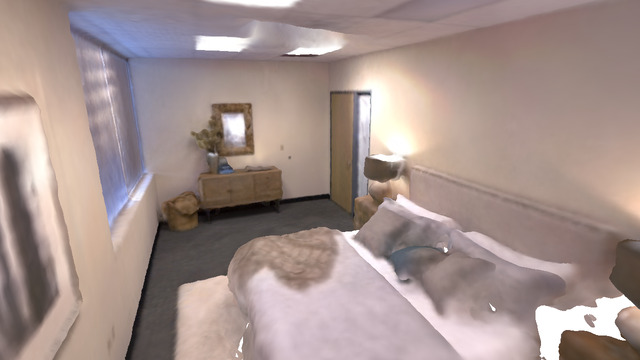}
        & \includegraphics[width=\linewidth]{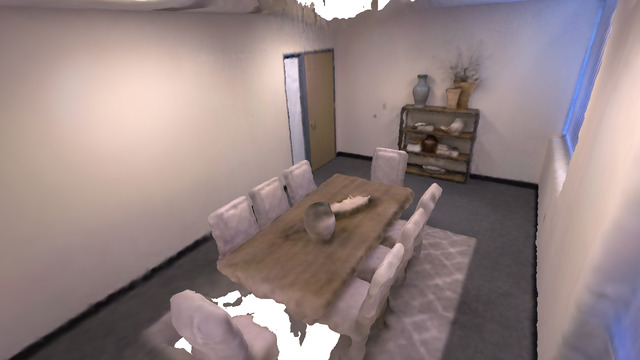}
    \end{tabular}
    \caption{Qualitative comparison of final meshes extracted by all methods on the Replica dataset \cite{replica19arxiv, sucar2021imap} (part 2).}
    \label{fig:qualitativereplicab}
\end{figure*}

\begin{figure*}[p!]
    \centering
    \scriptsize
    \setlength{\fboxsep}{0pt}
    \setlength{\tabcolsep}{1pt}
    \renewcommand{\arraystretch}{0.8}
    \begin{tabular}{C{1em}C{0.19\linewidth}C{0.19\linewidth}C{0.19\linewidth}C{0.19\linewidth}C{0.19\linewidth}}
        & \texttt{br} & \texttt{ck} & \texttt{gr} & \texttt{gwr} & \texttt{ki} \\
        \rotatebox{90}{Ground-truth} 
        & \includegraphics[width=\linewidth]{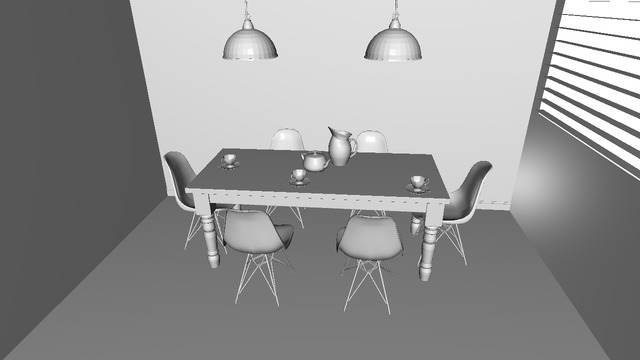} 
        & \includegraphics[width=\linewidth]{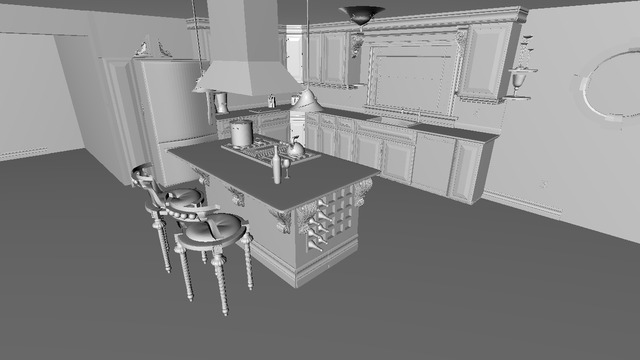}
        & \includegraphics[width=\linewidth]{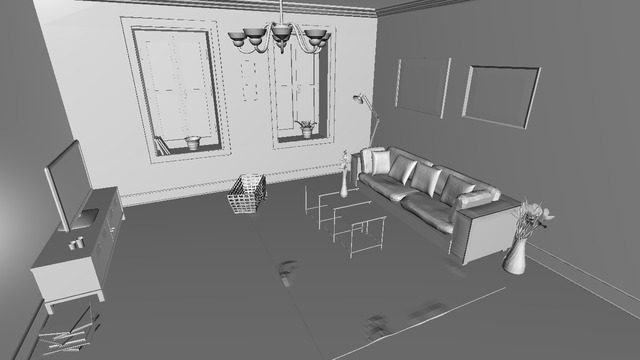}
        & \includegraphics[width=\linewidth]{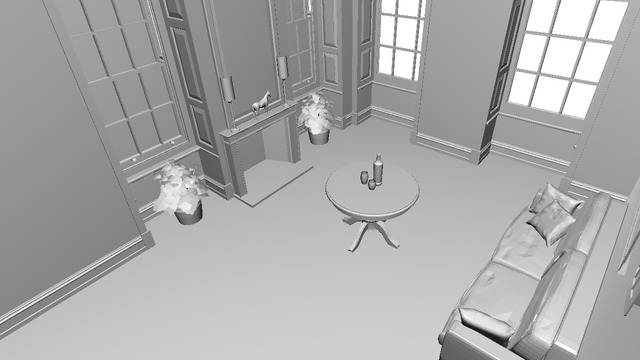}
        & \includegraphics[width=\linewidth]{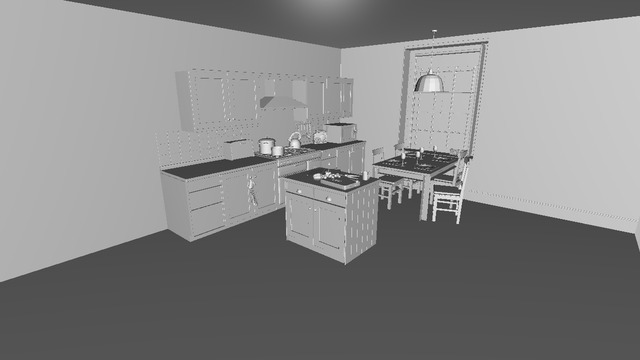} \\
        \rotatebox{90}{NICE-SLAM \cite{zhu2022nice}} 
        & \includegraphics[width=\linewidth]{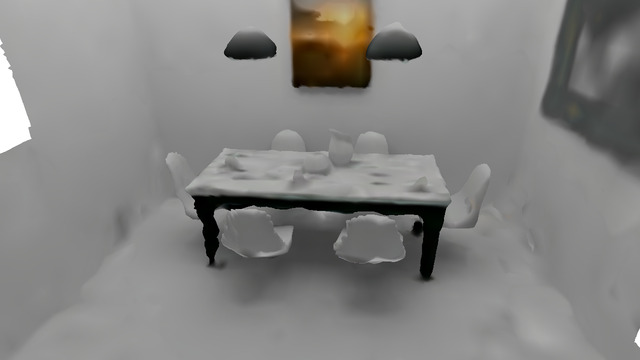}
        & \includegraphics[width=\linewidth]{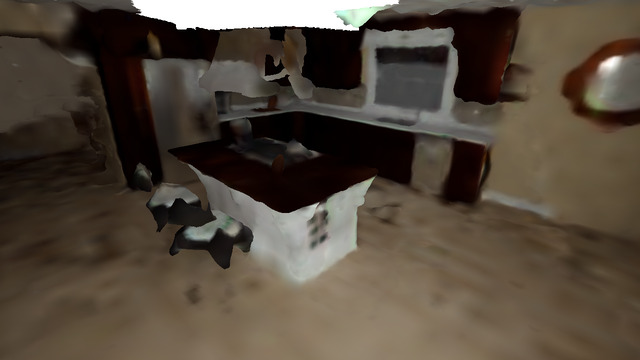}
        & \includegraphics[width=\linewidth]{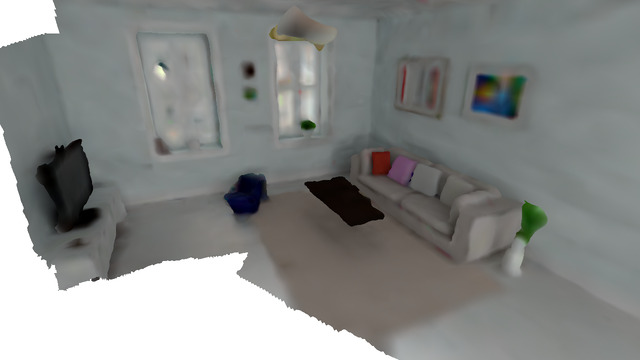}
        & \includegraphics[width=\linewidth]{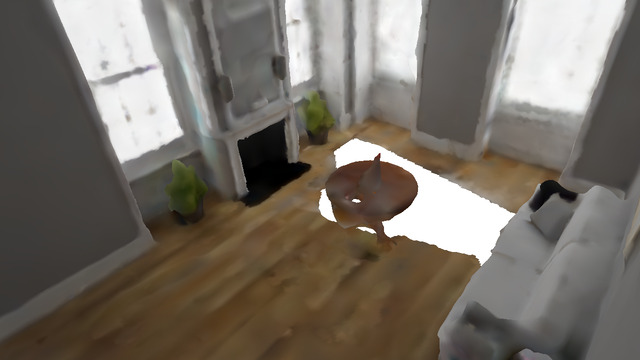}
        & \includegraphics[width=\linewidth]{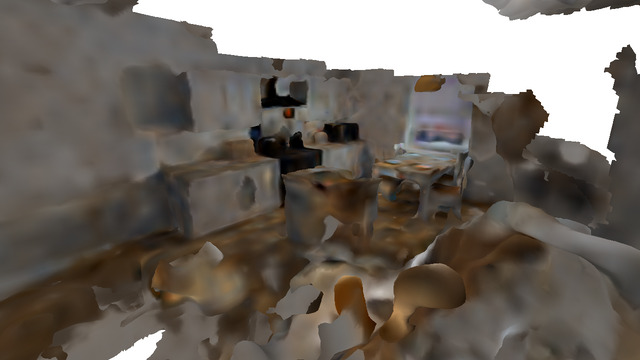} \\
        \rotatebox{90}{Co-SLAM \cite{wang2023co}}
        & \includegraphics[width=\linewidth]{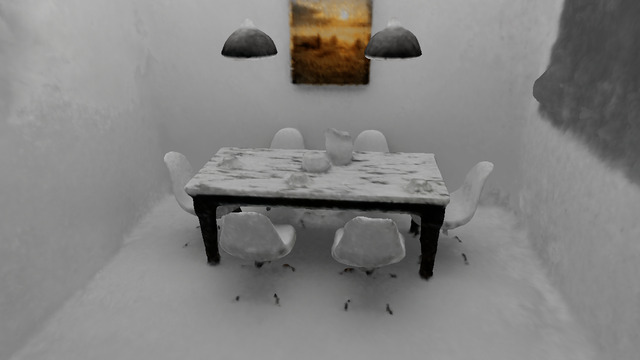}
        & \includegraphics[width=\linewidth]{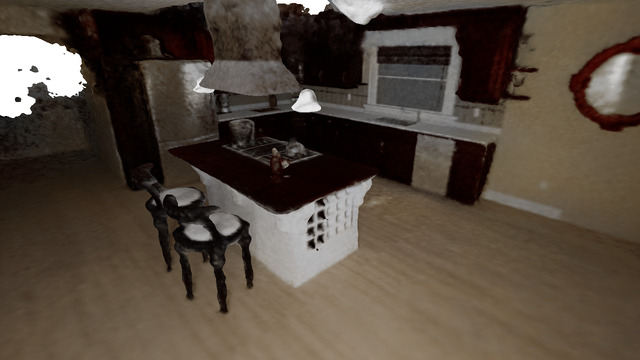}
        & \includegraphics[width=\linewidth]{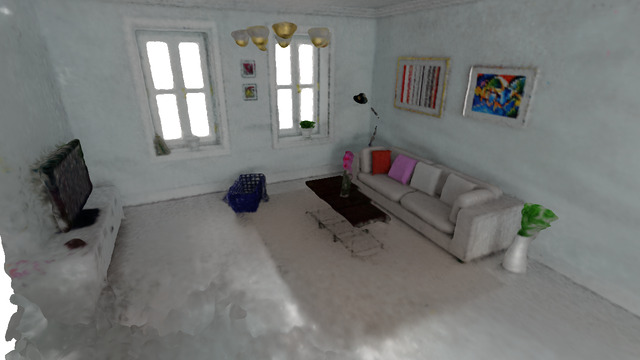}
        & \includegraphics[width=\linewidth]{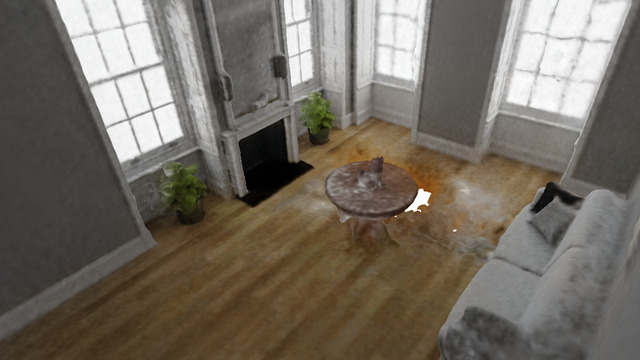}
        & \includegraphics[width=\linewidth]{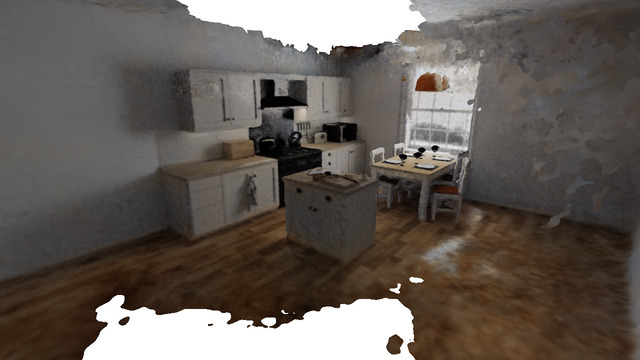} \\
        \rotatebox{90}{GO-SLAM \cite{zhang2023go}}
        & \includegraphics[width=\linewidth]{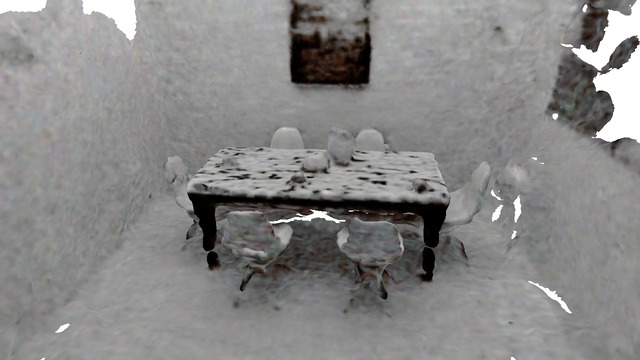}
        & \includegraphics[width=\linewidth]{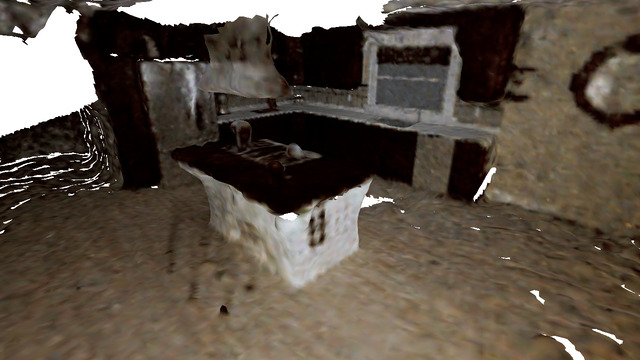}
        & \includegraphics[width=\linewidth]{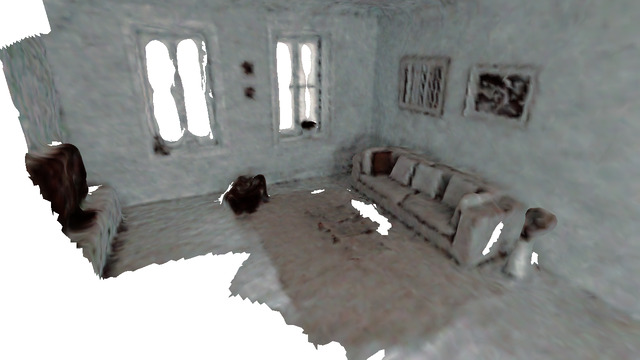}
        & \includegraphics[width=\linewidth]{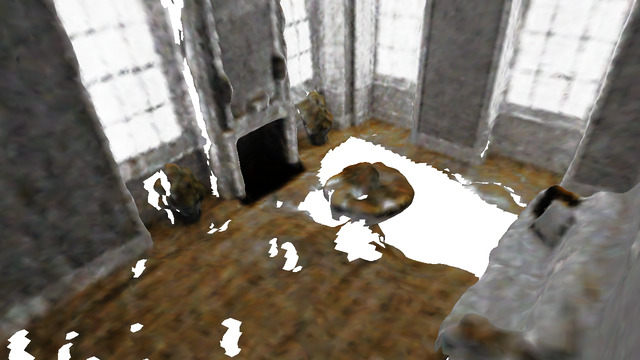}
        & \includegraphics[width=\linewidth]{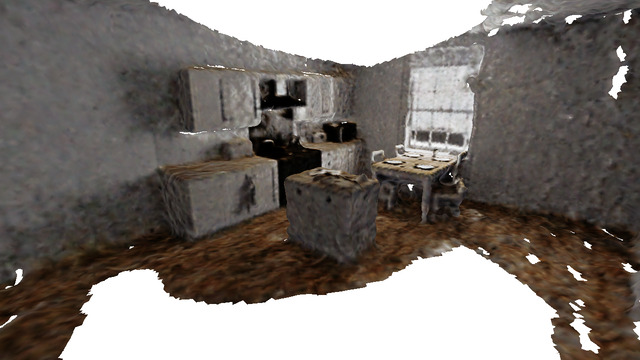} \\
        \rotatebox{90}{MIPS-Fusion \cite{tang2023mips}}
        & \includegraphics[width=\linewidth]{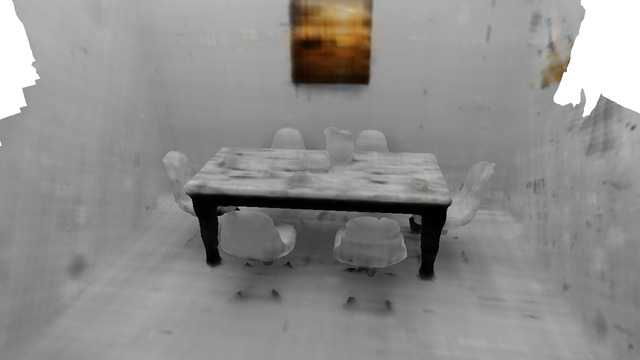}
        & \includegraphics[width=\linewidth]{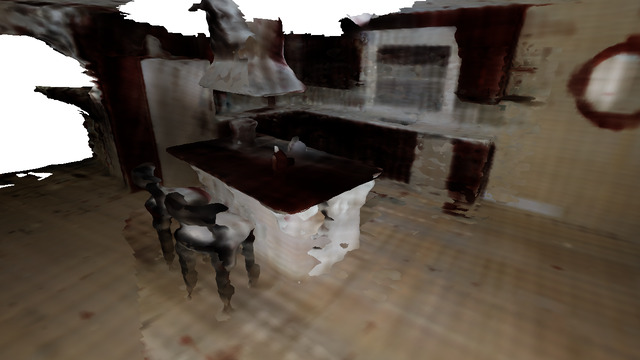}
        & \includegraphics[width=\linewidth]{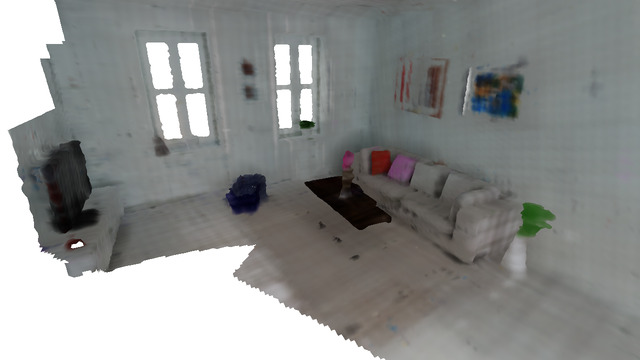}
        & \includegraphics[width=\linewidth]{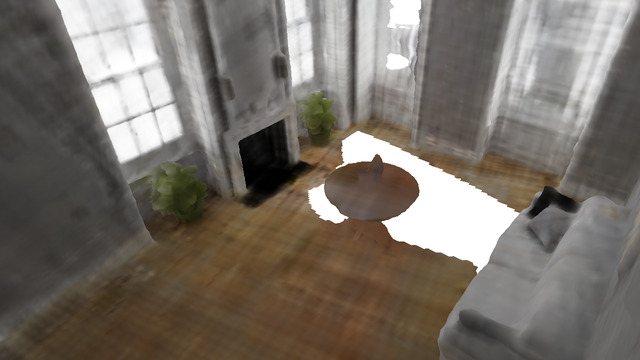}
        & \includegraphics[width=\linewidth]{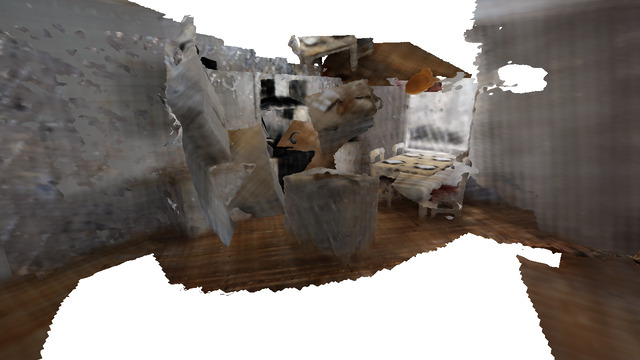} \\
        \rotatebox{90}{Loopy-SLAM \cite{liso2024loopy}}
        & \includegraphics[width=\linewidth]{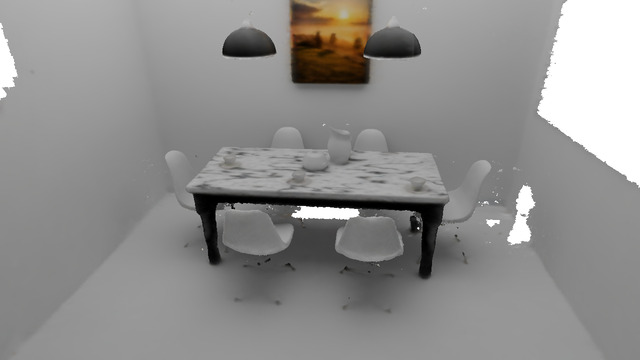}
        & \includegraphics[width=\linewidth]{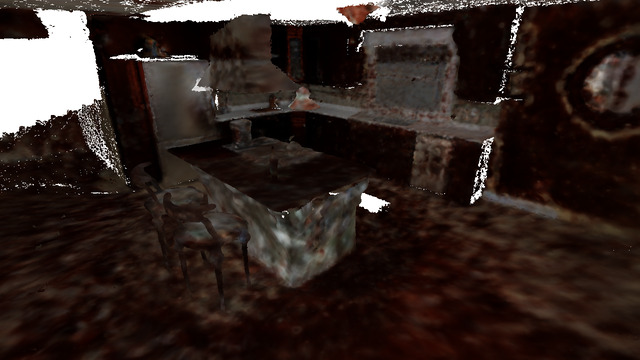}
        & \includegraphics[width=\linewidth]{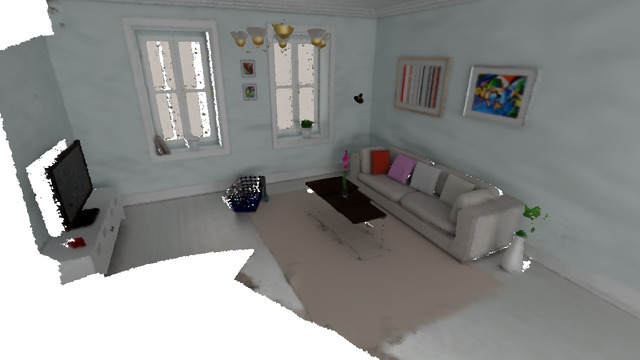}
        & \includegraphics[width=\linewidth]{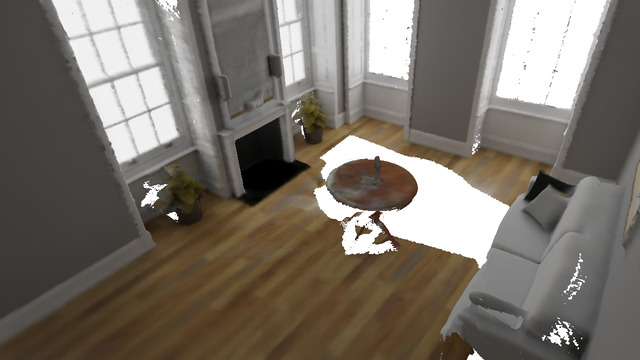}
        & \includegraphics[width=\linewidth]{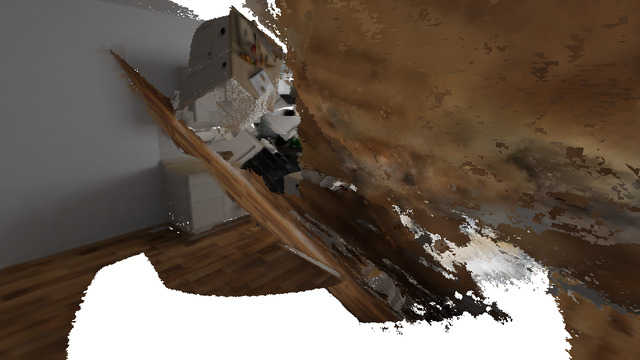} \\
        \rotatebox{90}{Ours-SF}
        & \includegraphics[width=\linewidth]{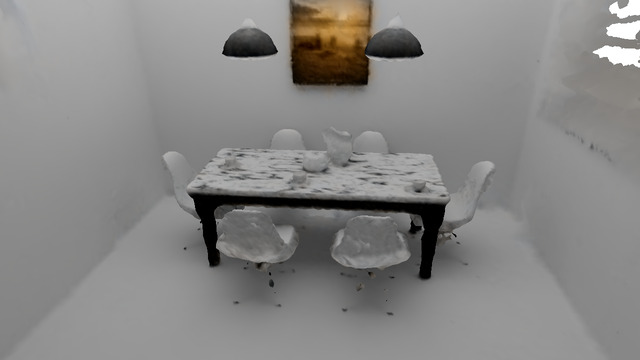}
        & \includegraphics[width=\linewidth]{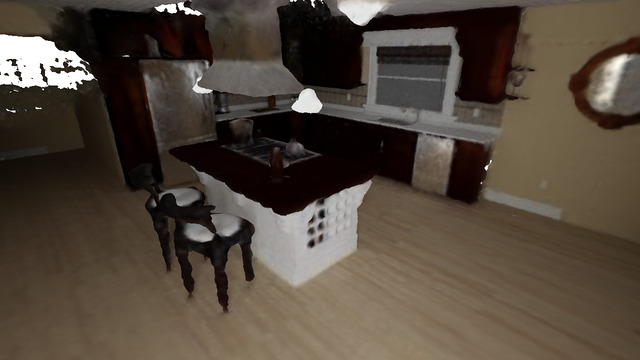}
        & \includegraphics[width=\linewidth]{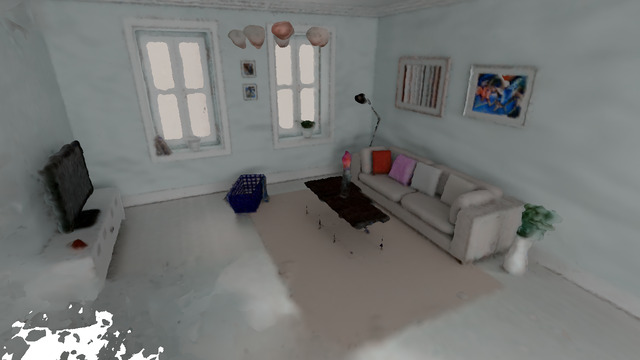}
        & \includegraphics[width=\linewidth]{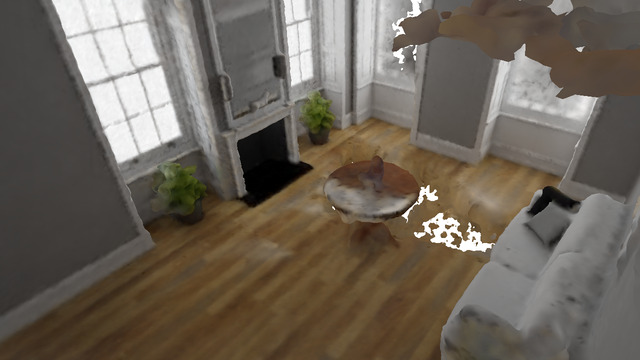}
        & \includegraphics[width=\linewidth]{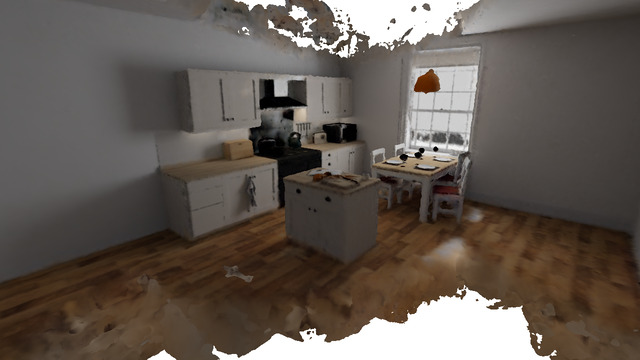} \\
        \rotatebox{90}{Ours}
        & \includegraphics[width=\linewidth]{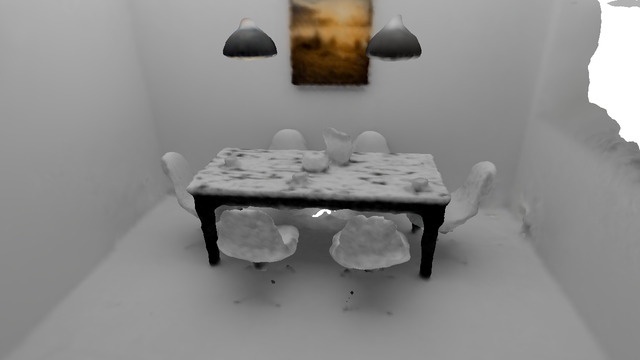}
        & \includegraphics[width=\linewidth]{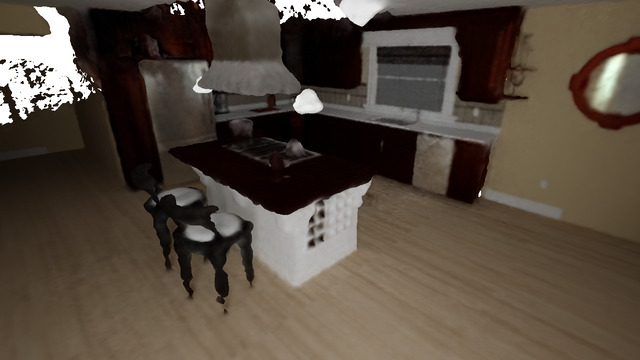}
        & \includegraphics[width=\linewidth]{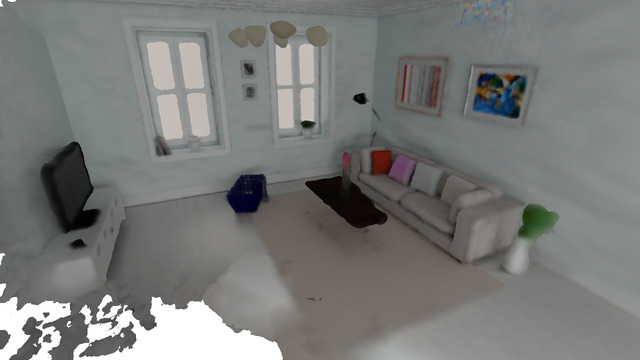}
        & \includegraphics[width=\linewidth]{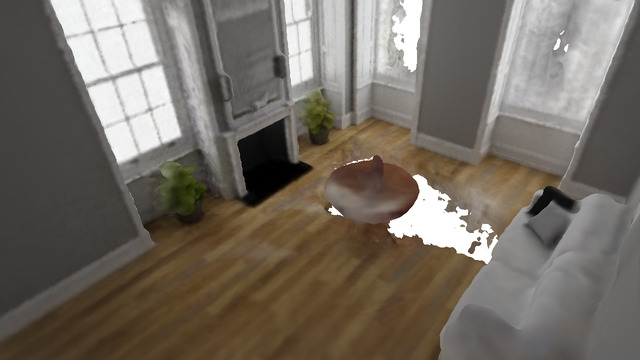}
        & \includegraphics[width=\linewidth]{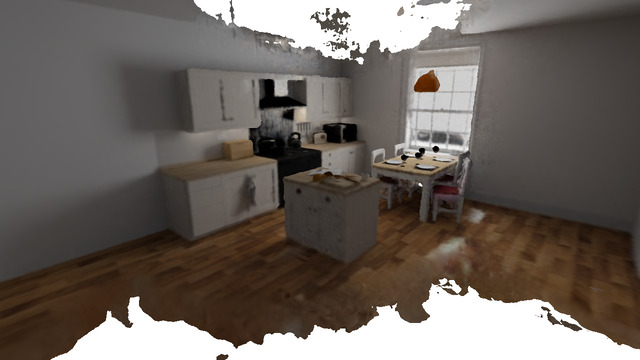}
    \end{tabular}
    \caption{Qualitative comparison of final meshes extracted by all evaluated methods on the NRGBD dataset \cite{azinovic2022neural} (part 1).}
    \label{fig:qualitativenrgbd}
\end{figure*}

\begin{figure*}[p!]
    \centering
    \scriptsize
    \setlength{\fboxsep}{0pt}
    \setlength{\tabcolsep}{1pt}
    \renewcommand{\arraystretch}{0.8}
    \begin{tabular}{C{1em}C{0.24\linewidth}C{0.24\linewidth}C{0.24\linewidth}C{0.24\linewidth}}
        & \texttt{ma} & \texttt{sc} & \texttt{tg} & \texttt{wr} \\
        \rotatebox{90}{Ground-truth} 
        & \includegraphics[width=\linewidth]{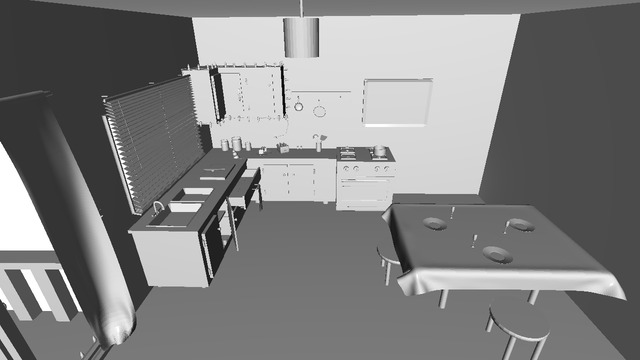} 
        & \includegraphics[width=\linewidth]{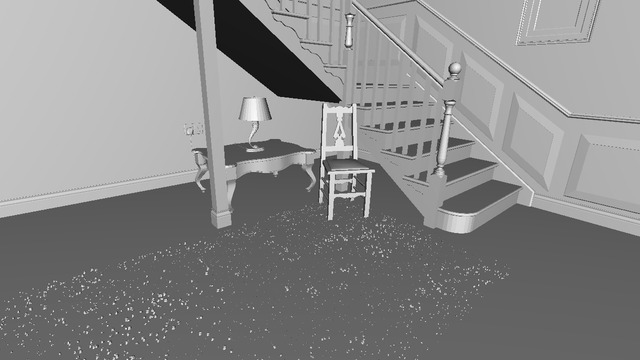}
        & \includegraphics[width=\linewidth]{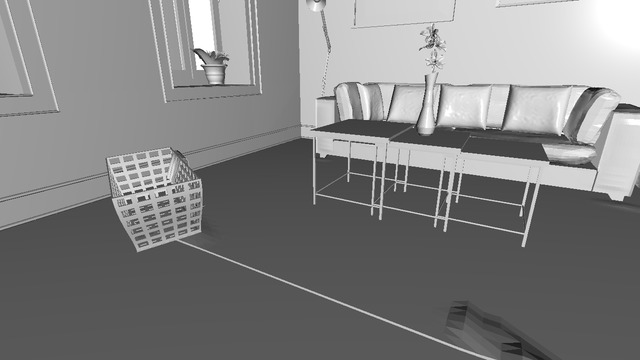}
        & \includegraphics[width=\linewidth]{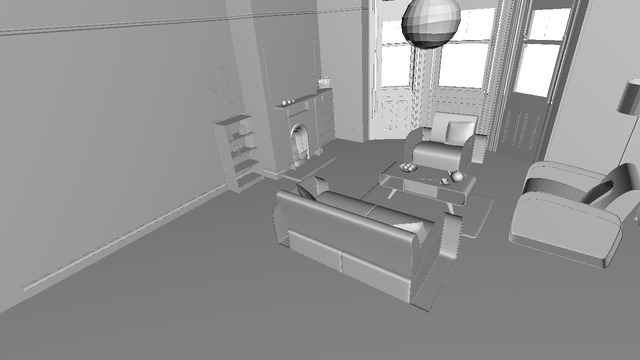} \\
        \rotatebox{90}{NICE-SLAM \cite{zhu2022nice}} 
        & \includegraphics[width=\linewidth]{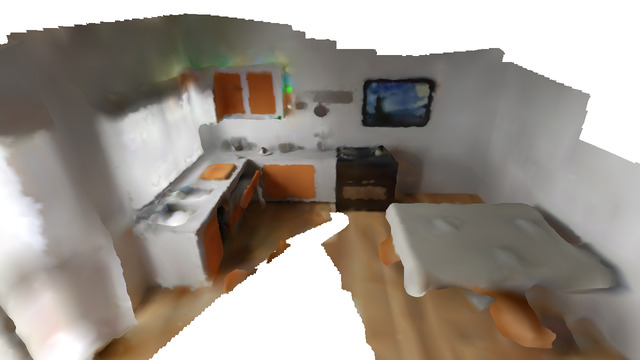}
        & \includegraphics[width=\linewidth]{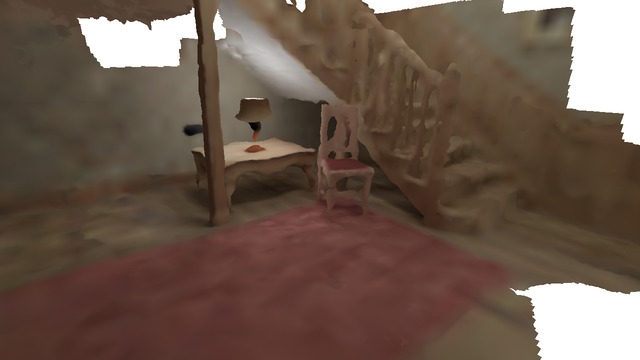}
        & \includegraphics[width=\linewidth]{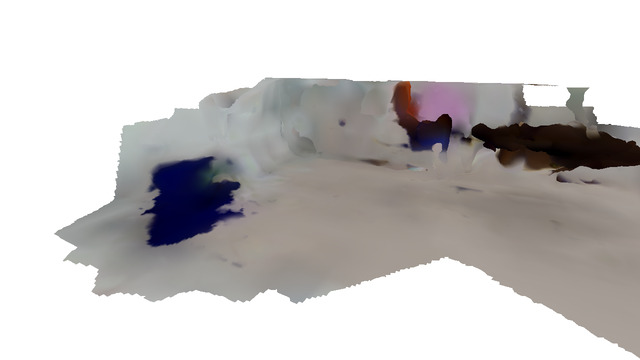}
        & \includegraphics[width=\linewidth]{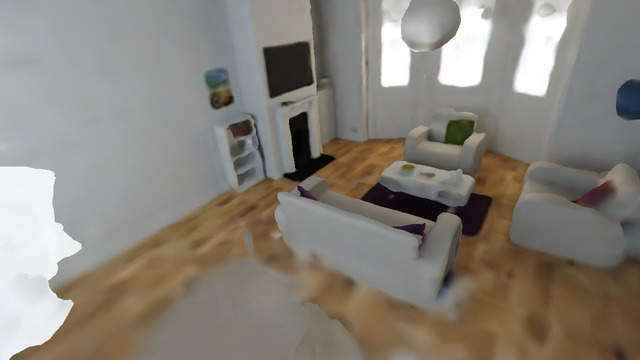} \\
        \rotatebox{90}{Co-SLAM \cite{wang2023co}}
        & \includegraphics[width=\linewidth]{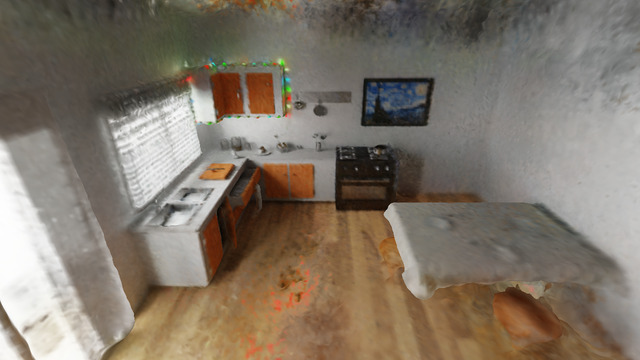}
        & \includegraphics[width=\linewidth]{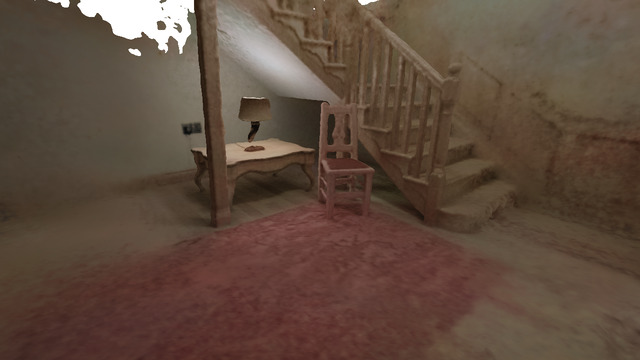}
        & \includegraphics[width=\linewidth]{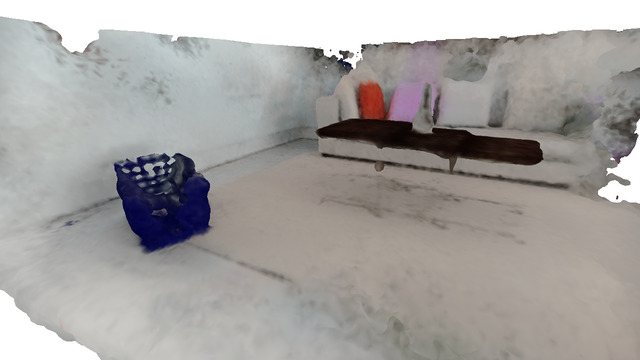}
        & \includegraphics[width=\linewidth]{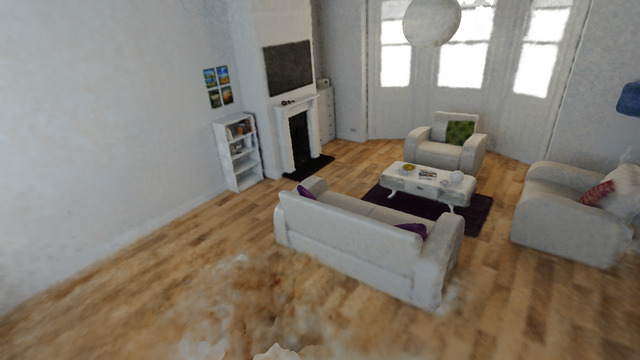} \\
        \rotatebox{90}{GO-SLAM \cite{zhang2023go}}
        & \includegraphics[width=\linewidth]{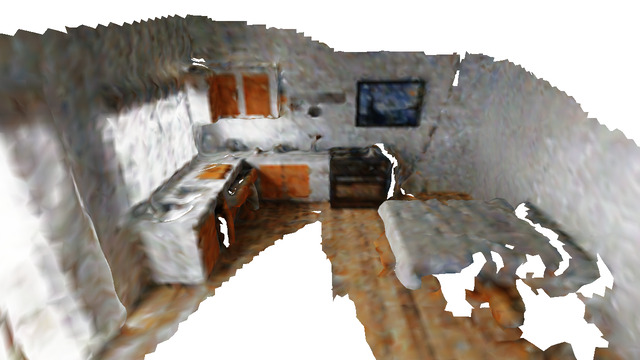}
        & \includegraphics[width=\linewidth]{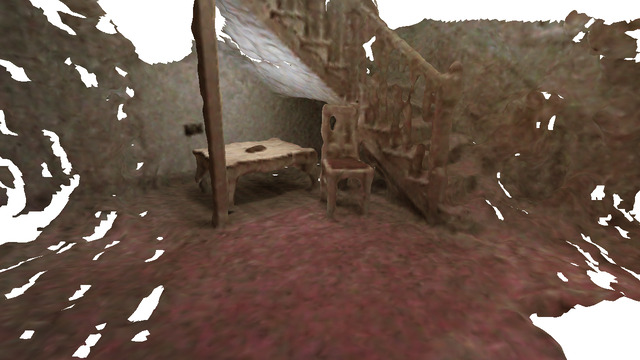}
        & \includegraphics[width=\linewidth]{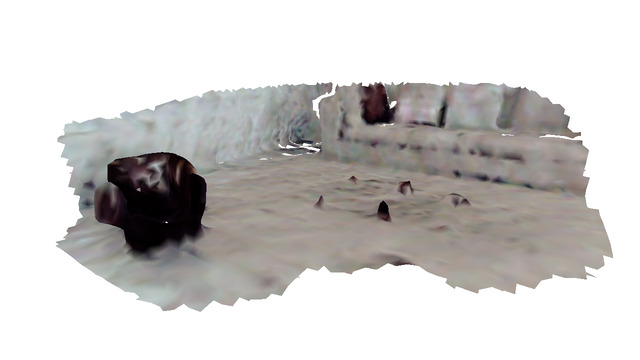}
        & \includegraphics[width=\linewidth]{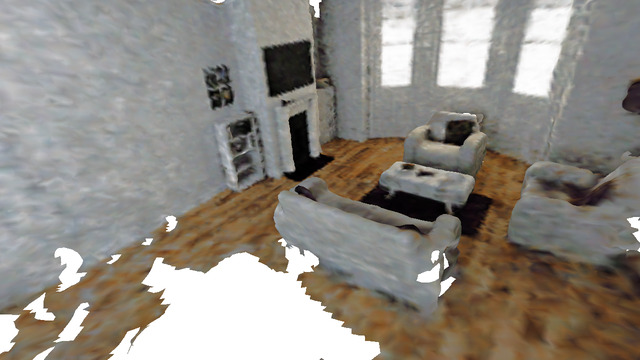} \\
        \rotatebox{90}{MIPS-Fusion \cite{tang2023mips}}
        & \includegraphics[width=\linewidth]{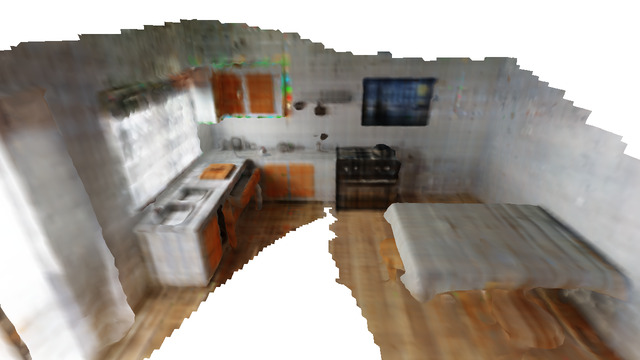}
        & \includegraphics[width=\linewidth]{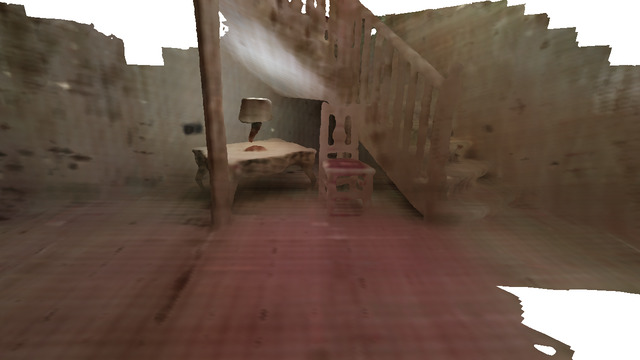}
        & \includegraphics[width=\linewidth]{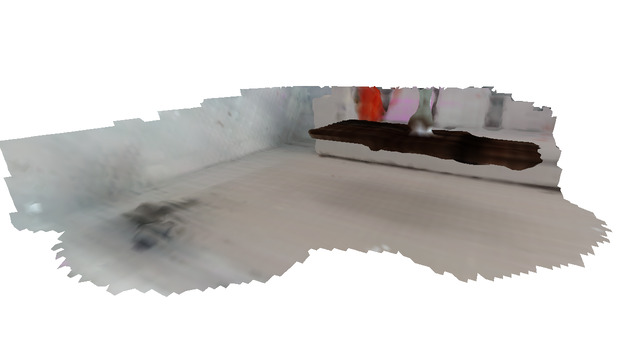}
        & \includegraphics[width=\linewidth]{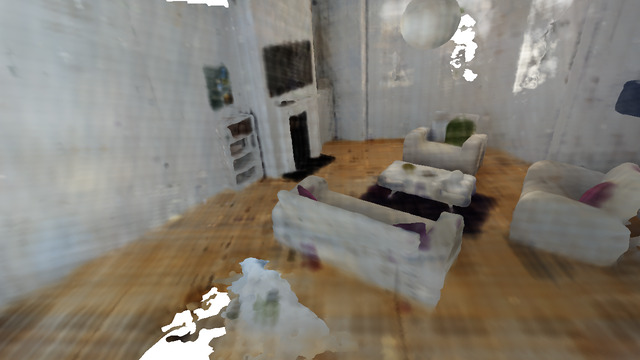} \\
        \rotatebox{90}{Loopy-SLAM \cite{liso2024loopy}}
        & \includegraphics[width=\linewidth]{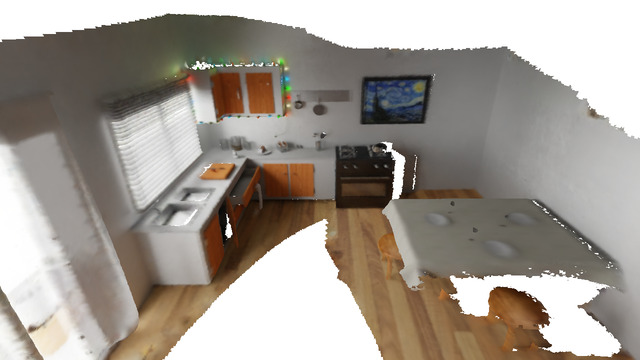}
        & \includegraphics[width=\linewidth]{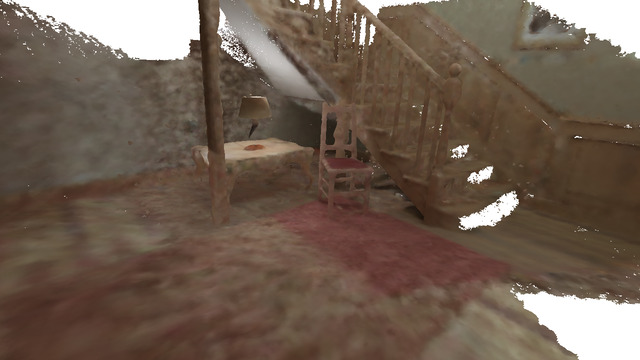}
        & \includegraphics[width=\linewidth]{figures/comp_extended/sc/loopyslam.jpg}
        & \includegraphics[width=\linewidth]{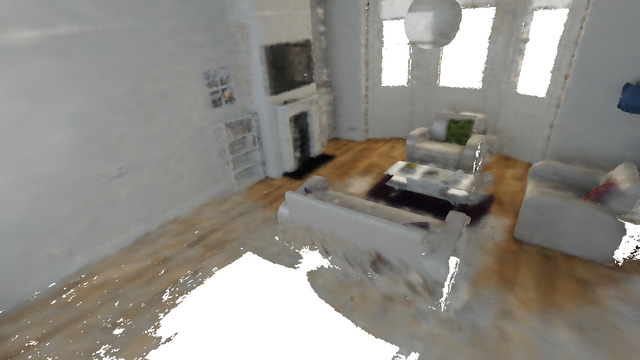} \\
        \rotatebox{90}{Ours-SF}
        & \includegraphics[width=\linewidth]{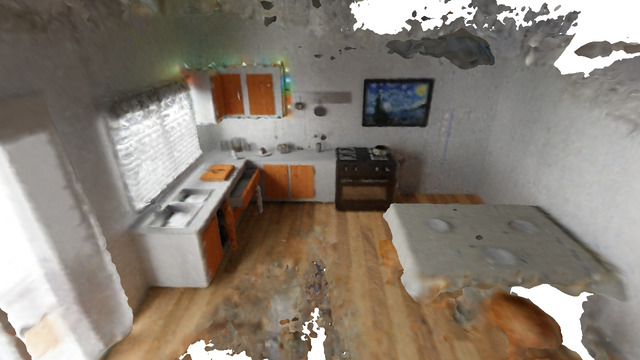}
        & \includegraphics[width=\linewidth]{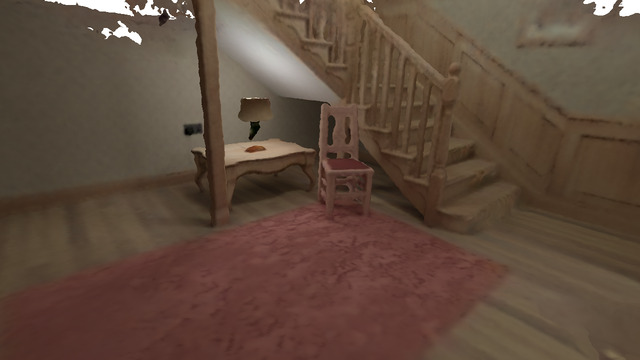}
        & \includegraphics[width=\linewidth]{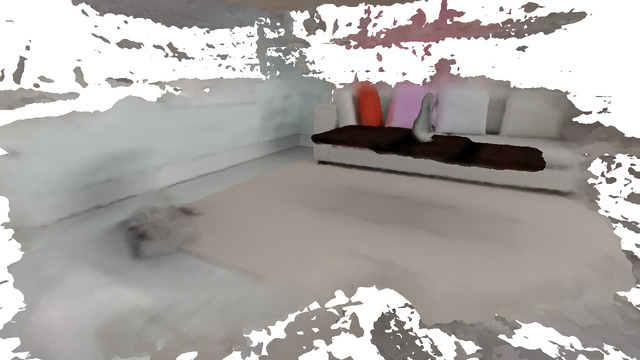}
        & \includegraphics[width=\linewidth]{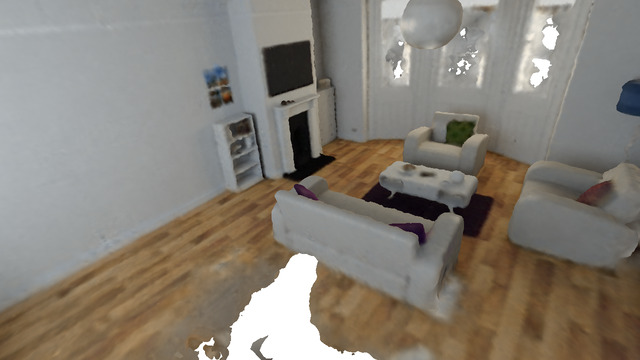} \\
        \rotatebox{90}{Ours}
        & \includegraphics[width=\linewidth]{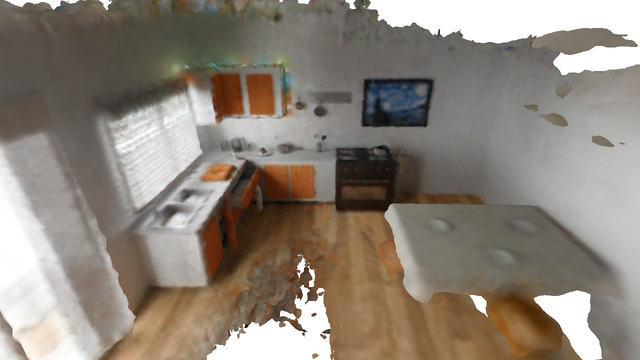}
        & \includegraphics[width=\linewidth]{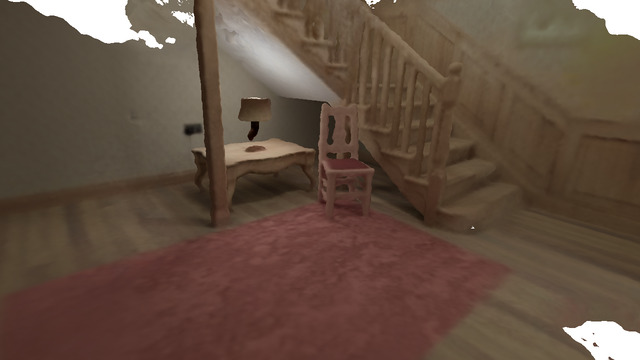}
        & \includegraphics[width=\linewidth]{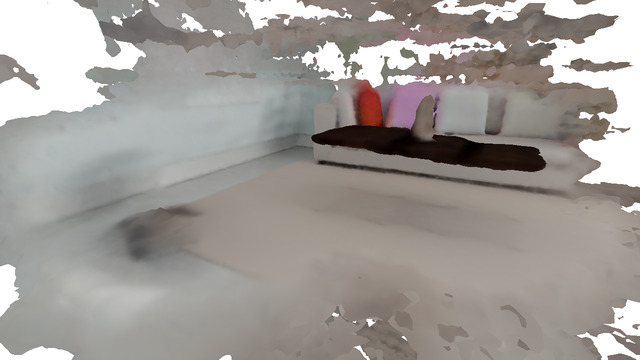}
        & \includegraphics[width=\linewidth]{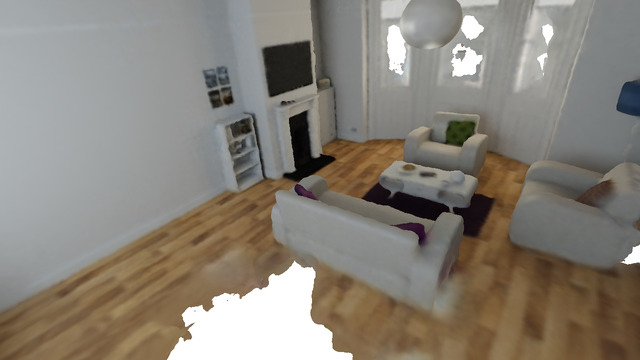} \\
    \end{tabular}
    \caption{Qualitative comparison of final meshes extracted by all evaluated methods on the NRGBD dataset \cite{azinovic2022neural} (part 2).}
    \label{fig:qualitativenrgbdb}
\end{figure*}

\begin{figure*}[p!]
    \centering
    \scriptsize
    \setlength{\fboxsep}{0pt}
    \setlength{\tabcolsep}{1pt}
    \renewcommand{\arraystretch}{0.8}
    \begin{tabular}{C{1em}C{0.25\linewidth}C{0.25\linewidth}C{0.25\linewidth}}
        & \texttt{scene0059\_00} & \texttt{scene0181\_00} & \texttt{scene0207\_00} \\
        \rotatebox{90}{Ground-truth} 
        & \includegraphics[width=\linewidth]{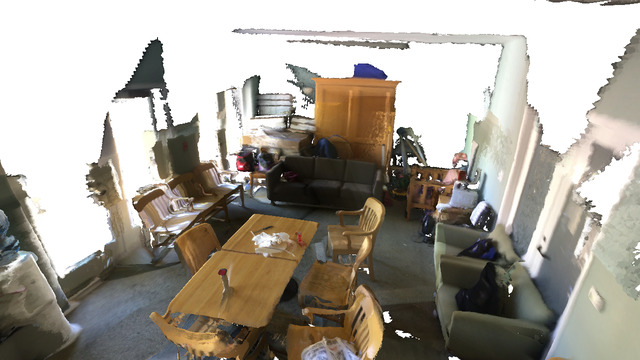} 
        & \includegraphics[width=\linewidth]{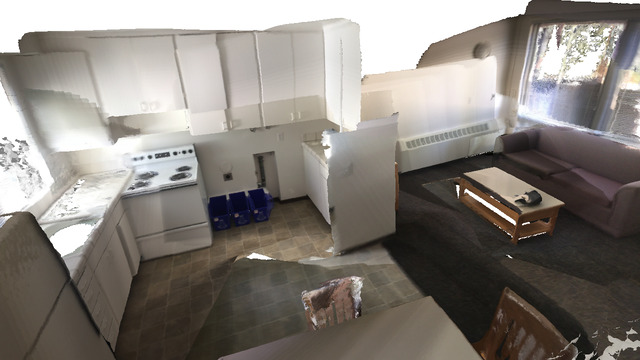}
        & \includegraphics[width=\linewidth]{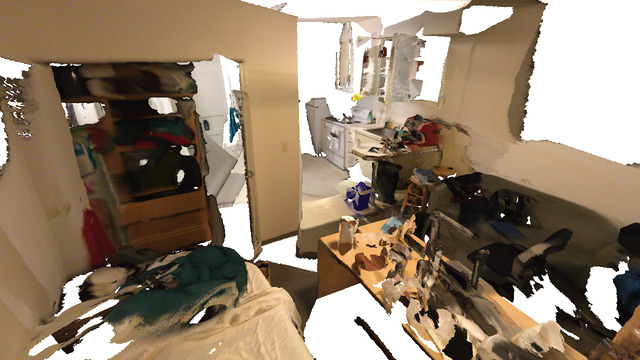} \\
        \rotatebox{90}{NICE-SLAM \cite{zhu2022nice}} 
        & \includegraphics[width=\linewidth]{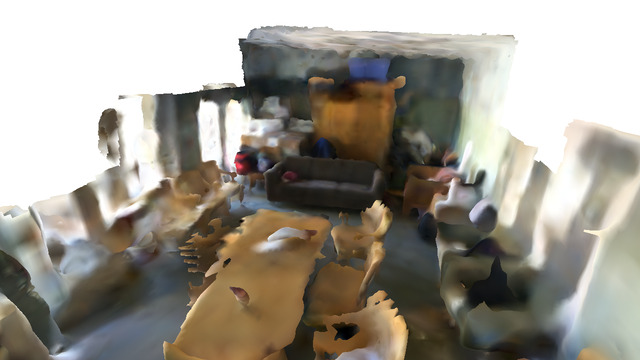}
        & \includegraphics[width=\linewidth]{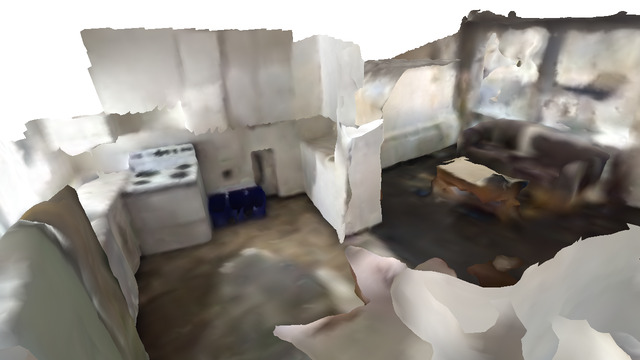}
        & \includegraphics[width=\linewidth]{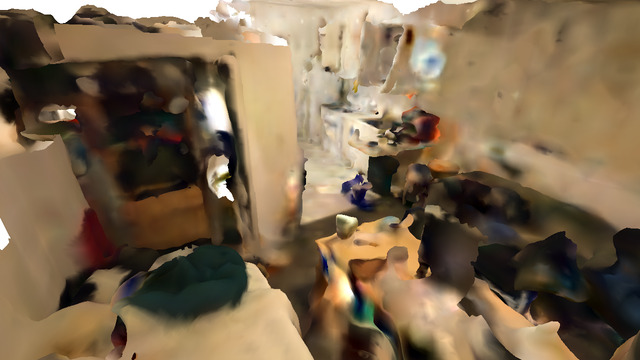} \\
        \rotatebox{90}{Co-SLAM \cite{wang2023co}}
        & \includegraphics[width=\linewidth]{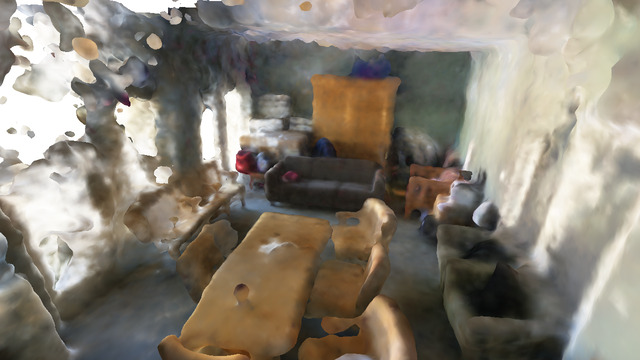}
        & \includegraphics[width=\linewidth]{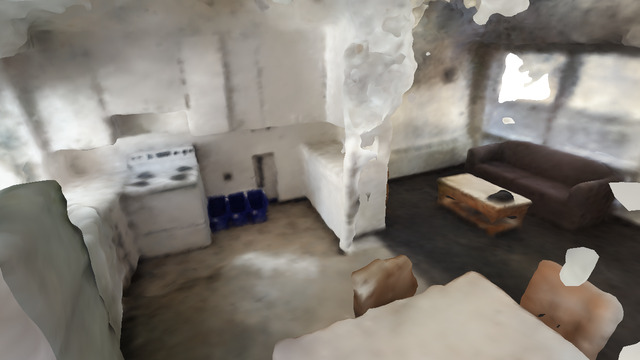}
        & \includegraphics[width=\linewidth]{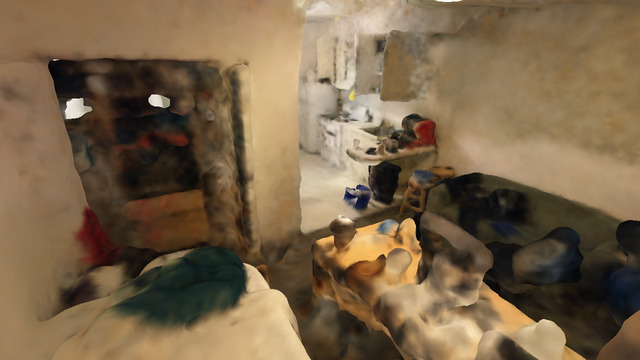} \\
        \rotatebox{90}{GO-SLAM \cite{zhang2023go}}
        & \includegraphics[width=\linewidth]{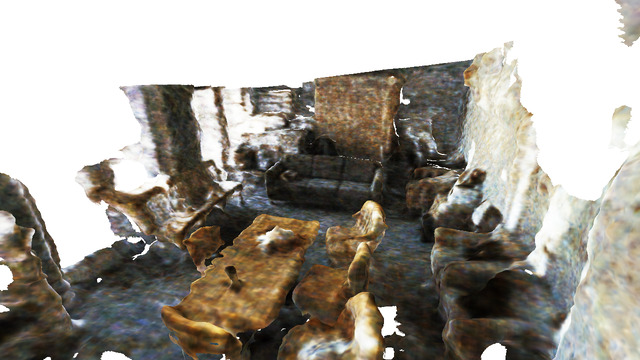}
        & \includegraphics[width=\linewidth]{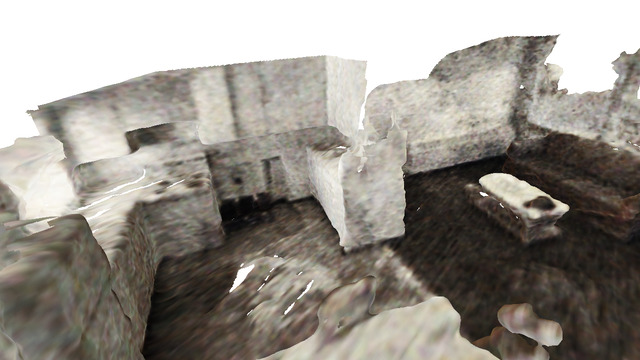}
        & \includegraphics[width=\linewidth]{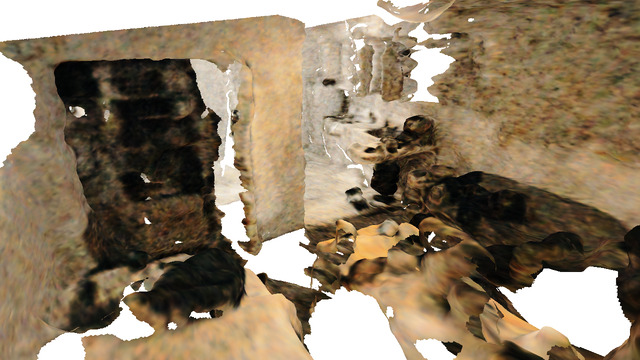} \\
        \rotatebox{90}{MIPS-Fusion \cite{tang2023mips}}
        & \includegraphics[width=\linewidth]{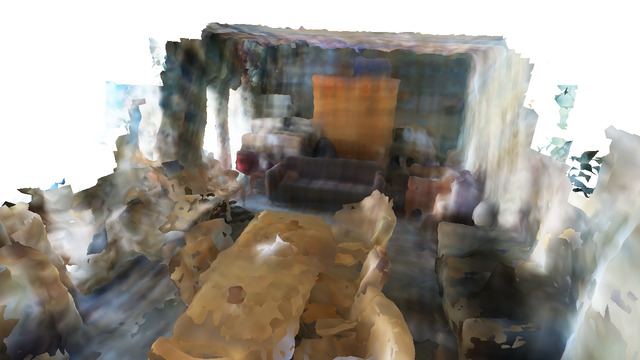}
        & \includegraphics[width=\linewidth]{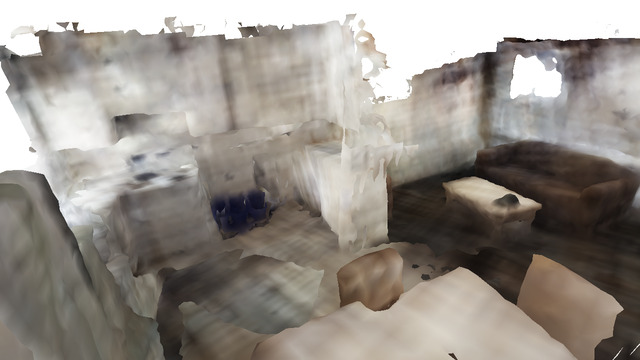}
        & \includegraphics[width=\linewidth]{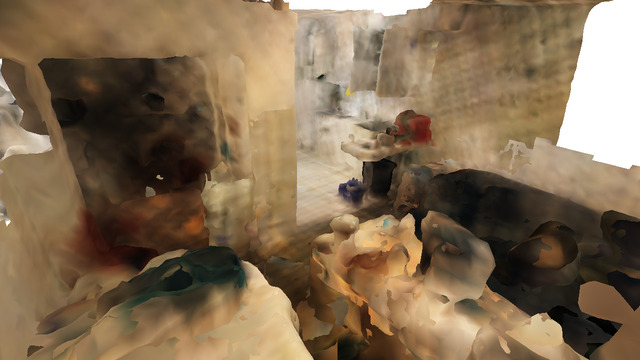} \\
        \rotatebox{90}{Loopy-SLAM \cite{liso2024loopy}}
        & \includegraphics[width=\linewidth]{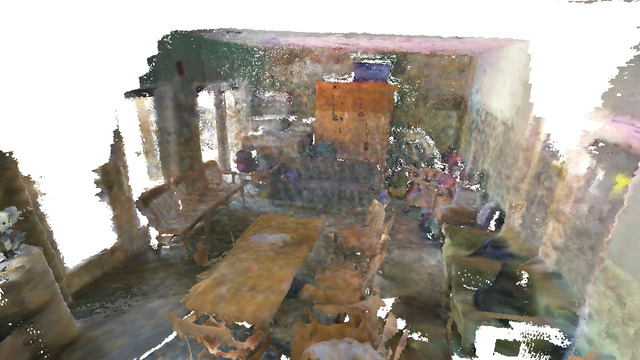}
        & \includegraphics[width=\linewidth]{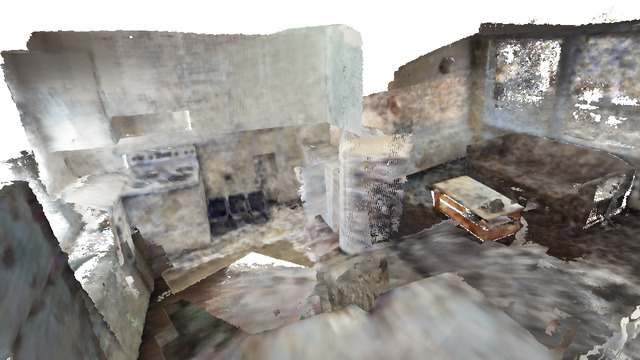}
        & \includegraphics[width=\linewidth]{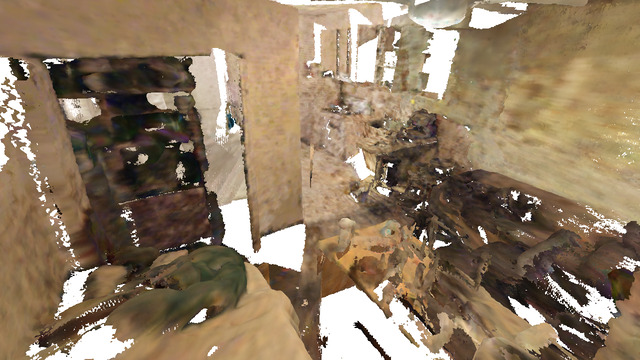} \\
        \rotatebox{90}{Ours-SF}
        & \includegraphics[width=\linewidth]{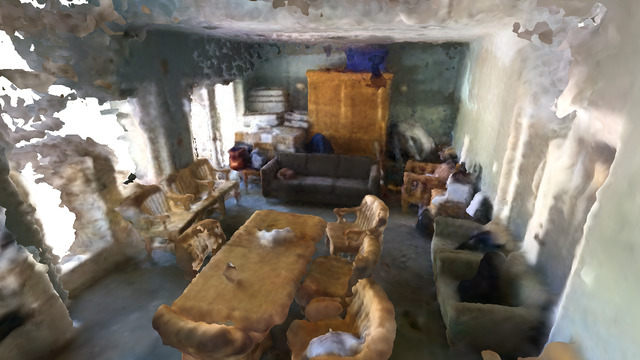}
        & \includegraphics[width=\linewidth]{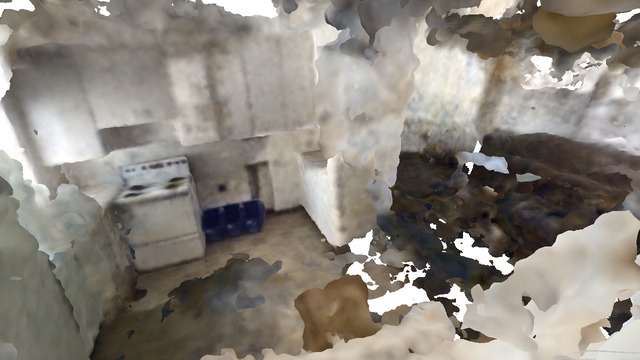}
        & \includegraphics[width=\linewidth]{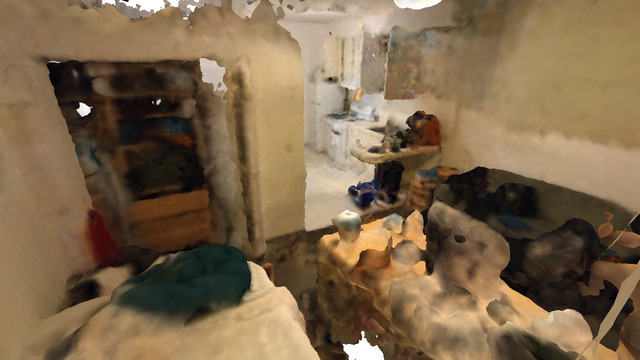} \\
        \rotatebox{90}{Ours}
        & \includegraphics[width=\linewidth]{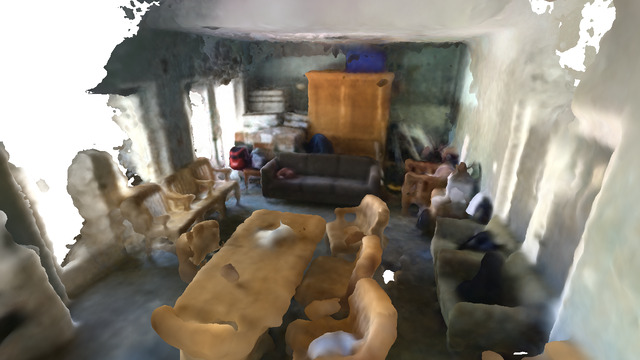}
        & \includegraphics[width=\linewidth]{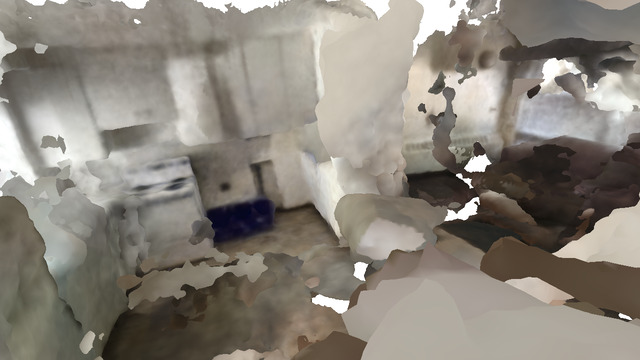}
        & \includegraphics[width=\linewidth]{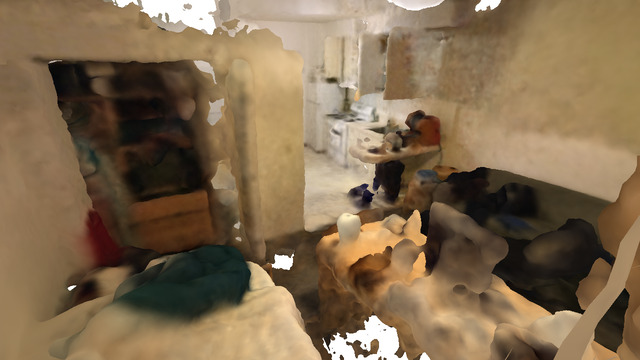}
    \end{tabular}
    \caption{Qualitative comparison of final meshes extracted by all methods on the ScanNet dataset \cite{dai2017scannet}. Our method fails on \texttt{scene0181\_00} due to tracking issues in the underlying SLAM system in a feature-less region.}
    \label{fig:qualitativescannet}
\end{figure*}

%% file: figures/knn.tikz
\begin{tikzpicture}
\scriptsize

\node[inner sep=0pt, outer sep=0pt] (knn1) at (0,0) {\includegraphics[width=0.32\linewidth]{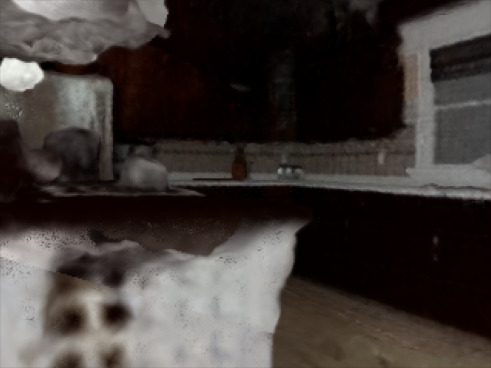}};
\node[inner sep=3pt, outer sep=3pt, anchor=south east, fill=white, fill opacity=0.5, text opacity=1.0] (knn1label) at (knn1.south east) {$k=1$};

\node[inner sep=0pt, outer sep=0pt, anchor=north west] (knn2) at ($(knn1.north east) + (0.01\linewidth,0.0)$ ) {\includegraphics[width=0.32\linewidth]{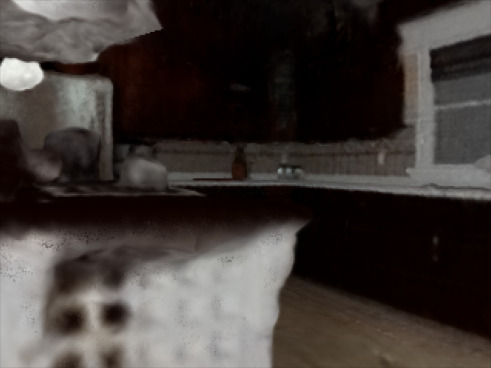}};
\node[inner sep=3pt, outer sep=3pt, anchor=south east, fill=white, fill opacity=0.5, text opacity=1.0] (knn2label) at (knn2.south east) {$k=2$};

\node[inner sep=0pt, outer sep=0pt, anchor=north west] (knn4) at ($(knn2.north east) + (0.01\linewidth,0.0)$) {\includegraphics[width=0.32\linewidth]{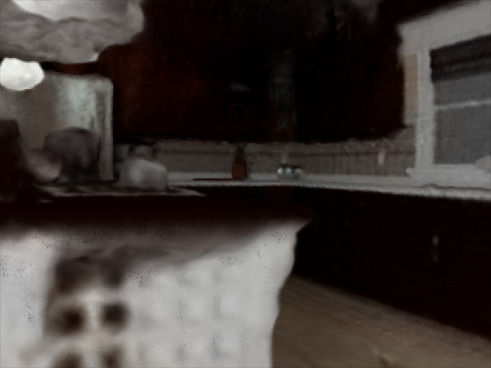}};
\node[inner sep=3pt, outer sep=3pt, anchor=south east, fill=white, fill opacity=0.5, text opacity=1.0] (knn4label) at (knn4.south east) {$k=4$};

\draw[thick, draw=orange, rounded corners=5pt] ($(knn1) + (-0.155\linewidth, -0.08\linewidth)$) rectangle ($(knn1) + (0.0\linewidth, 0.0\linewidth)$);
\draw[thick, draw=orange, rounded corners=5pt] ($(knn1) + (-0.11\linewidth, 0.06\linewidth)$) rectangle ($(knn1) + (-0.03\linewidth, 0.12\linewidth)$);

\draw[thick, draw=orange, rounded corners=5pt] ($(knn2) + (-0.155\linewidth, -0.08\linewidth)$) rectangle ($(knn2) + (0.0\linewidth, 0.0\linewidth)$);
\draw[thick, draw=orange, rounded corners=5pt] ($(knn2) + (-0.11\linewidth, 0.06\linewidth)$) rectangle ($(knn2) + (-0.03\linewidth, 0.12\linewidth)$);

\draw[thick, draw=orange, rounded corners=5pt] ($(knn4) + (-0.155\linewidth, -0.08\linewidth)$) rectangle ($(knn4) + (0.0\linewidth, 0.0\linewidth)$);
\draw[thick, draw=orange, rounded corners=5pt] ($(knn4) + (-0.11\linewidth, 0.06\linewidth)$) rectangle ($(knn4) + (-0.03\linewidth, 0.12\linewidth)$);

\end{tikzpicture}%